%% file: manuscript.tex
\newif\iffinal
\finalfalse
\documentclass[default,iicol]{sn-jnl}

\usepackage{amsmath}
\usepackage{amssymb}
\usepackage{boxedminipage}
\usepackage{enumerate}
\usepackage{multirow}
\usepackage{graphics}
\usepackage{url}
\usepackage{times}
\usepackage{version}
\usepackage{epsfig}
\input lstset.tex
\input macros.tex

\jyear{2023}%

\theoremstyle{thmstyleone}

\theoremstyle{thmstyletwo}%

\theoremstyle{thmstylethree}%

\newcommand{\xst}{\mathop{\rm subject~to}}
\newcommand{\xoptop}{\mathop{\rm\odot}}
\newcommand{\xdf}[1]{Definition~\ref{#1}}

\begin{document}

\title[Space Net Optimization]{Space Net Optimization}
\author*{\fnm{Chun-Wei} \sur{Tsai}}

\author{\fnm{Yi-Cheng} \sur{Yang}}
\author{\fnm{Tzu-Chieh} \sur{Tang}}
\author{\fnm{Che-Wei} \sur{Hsu}}

\affil{\orgdiv{Computer Science and Engineering}, \orgname{National Sun Yat-sen University},
       \orgaddress{\city{Kaohsiung},  \country{Taiwan}}}

     \abstract{Most metaheuristic algorithms rely on a few searched
       solutions to guide later searches during the convergence
       process for a simple reason: the limited computing resource of
       a computer makes it impossible to retain all the searched
       solutions.  This also reveals that each search of most
       {\xmetaha}s is just like a ballpark guess.  To help address
       this issue, we present a novel {\xmetaha} called {\xsnoa}. It
       is equipped with a new mechanism called space net; thus, making
       it possible for a {\xmetaha} to use most information provided
       by all searched solutions to depict the landscape of the
       solution space.  With the space net, a {\xmetaha} is kind of
       like having a ``vision'' on the solution space.  Simulation
       results show that {\xsnos} outperforms all the other
       {\xmetaha}s compared in this study for a set of well-known
       single objective bound constrained problems in most cases.  }

\maketitle

\section{Introduction}\label{sec1}
{\xMetaha}s provide an efficient way to find approximate solution(s)
for complex optimization problems in a reasonable time.  They can be
regarded as an important branch of unsupervised learning algorithms.
The development of {\xmetaha}s \cite{Blum_2003, Glover_2003} dated
back to the 1960s or even earlier. Different from most exhaustive
search algorithms that will check all possible feasible solutions to
find the global optimum solution and thus may be inefficient for
solving large and complex optimization problems, most {\xmetaha}s are
built on the idea of ``transit'' solutions strategically to generate
new candidate solution(s) and then ``evaluate'' these newly generated
solutions to ``determine'' later search directions or regions during
the convergence process so as to find approximate solution(s) for
complex optimization problems \cite{Tsai_2023}.
{\xMetaha}s can not only solve various complex optimization problems,
they can also be used to enhance the performance of supervised
learning algorithms for different research issues, such as finding a
set of proper hyperparameters \cite{Yang-2020} or a suitable neural
architecture \cite{Liu-2021} to enhance the performance of a deep
neural network (DNN).
Moreover, some recent studies \cite{Huang-2020, Liu-2023, Li-2023}
also showed that the performance of {\xmetaha}s for solving complex
optimization problems can be enhanced by using machine learning
algorithms.
\begin{figure*}[tbh]
  \centering
  \footnotesize
    \resizebox{.85\textwidth}{!}{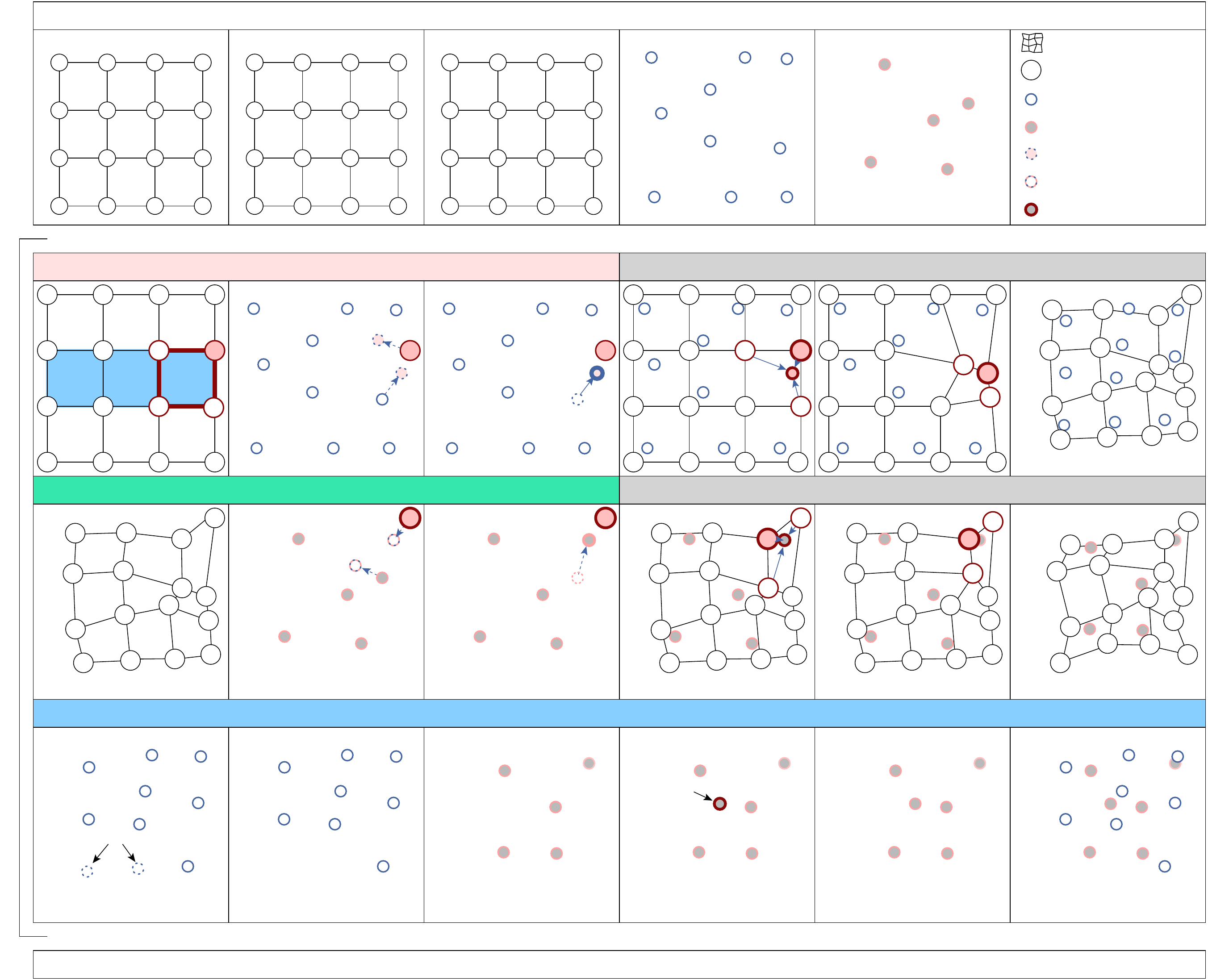}
    \caption{An overview of the {\xsno} that contains the operators
      for region search (RS), point search (PS), space net adjustment
      (SNA), and population adjustment (PA).}
  \label{fig:overall-1}
  \setlength\tabcolsep{1pt}
  \begin{tabular}{cccc}
      \includegraphics[width=0.25\textwidth]{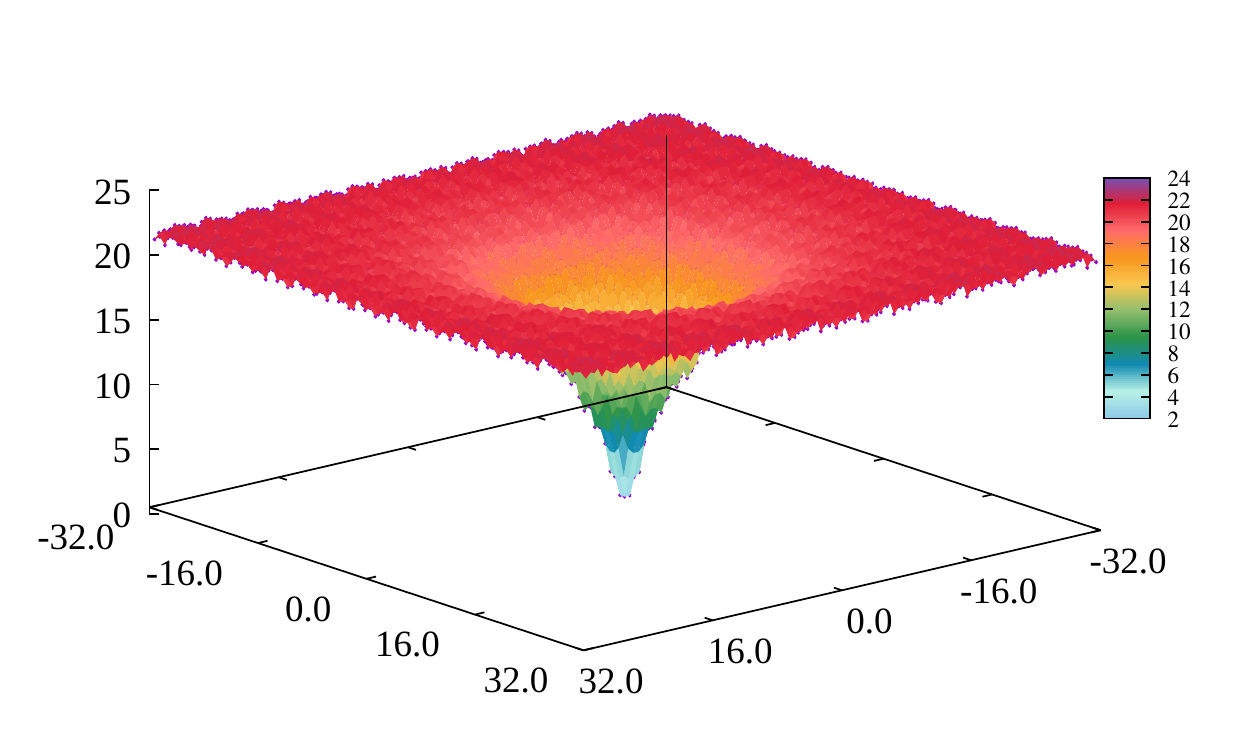} &
      \includegraphics[width=0.25\textwidth]{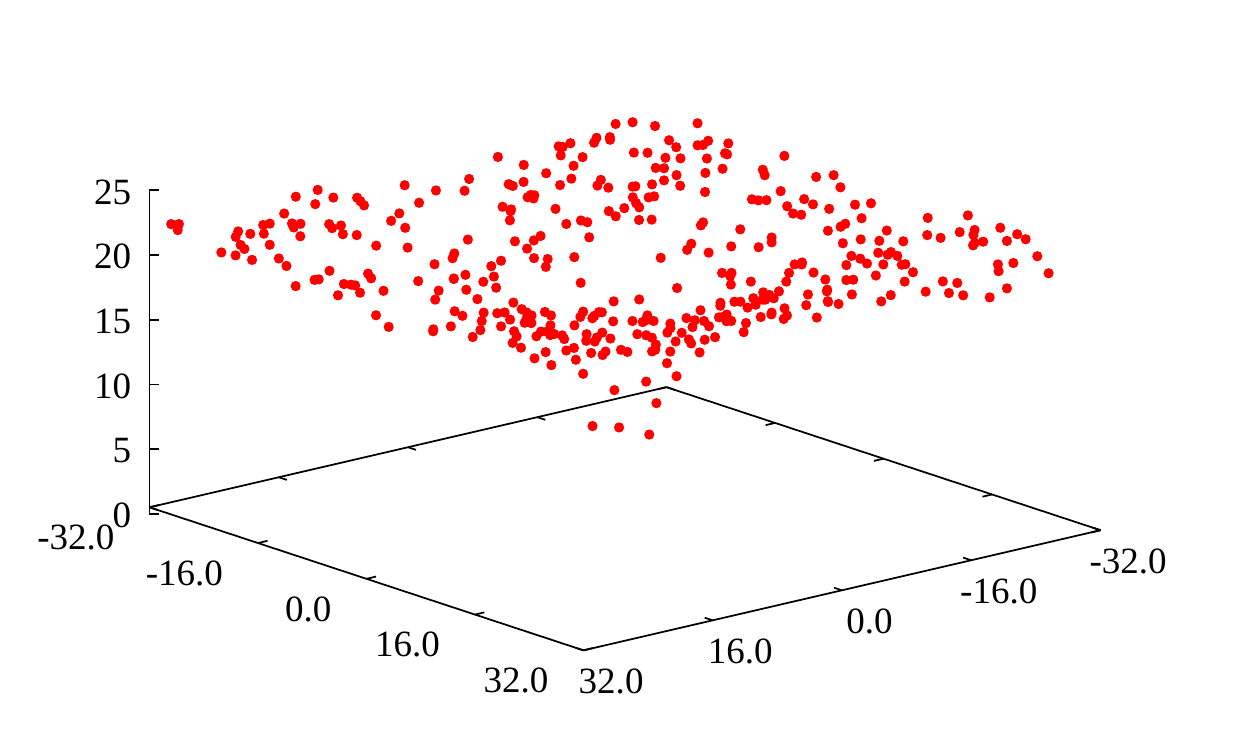} &
      \includegraphics[width=0.25\textwidth]{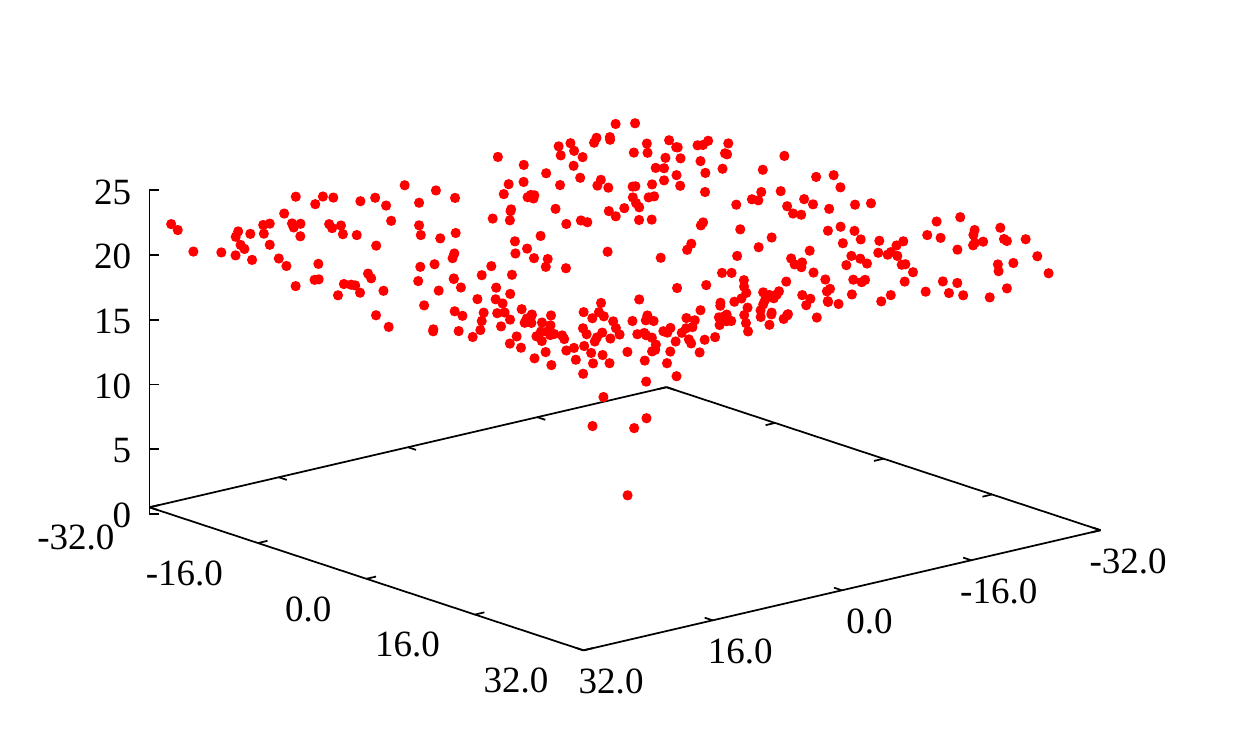} &
      \includegraphics[width=0.25\textwidth]{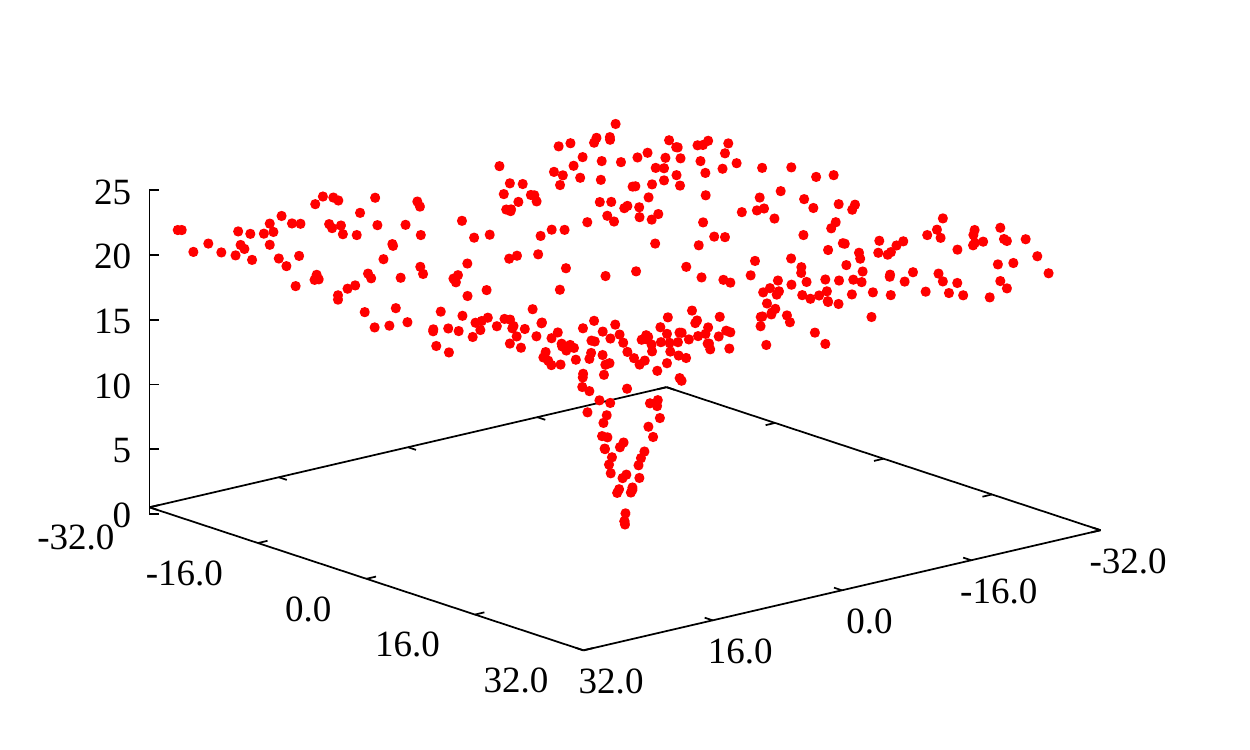} \\
      (a) & (b) & (c) & (d) \\
    \end{tabular}
    \caption{A simple example that applies {\xsno} to the 2-D Ackley
      function to explain its convergence process. (a) The landscape
      of the 2-D Ackley function.  (b) The scattering situation of
      elastic points on the space net after 400 evaluations.  (c) The
      scattering situation of elastic points on the space net after
      800 evaluations.  (d) The scattering situation of elastic points
      on the space net after 4,000 evaluations.}
       \label{fig:overall-2}     
\end{figure*}
Until now, only a few {\xmetaha}s focus on ``extracting'' the
characteristics from a certain number of recent good searched
solutions to guide the searches to move to particular regions (or
directions).
They are tabu list of tabu search (TS) \cite{Glover_1989}, personal
and global best solutions of particle swarm optimization (PSO)
\cite{Kennedy_1995}, and pheromone table of ant colony optimization
(ACO) \cite{Dorigo-1991-acoe1, Dorigo-2006}.
A similar idea can also be found in a more recent study \cite{Tanabe-2013}
that used a ``historical memory'' during the convergence process to
record successful control parameter settings of differential evolution
(DE) to dynamically adjust the transition operations, called
success-history based adaptive differential evolution (SHADE).
From these studies, it can be easily seen that (1) keeping a certain
number of good searched solutions (e.g., PSO \cite{Kennedy_1995} and
TS \cite{Glover_1989}), (2) using a new data structure to extract the
feature of subsolutions (e.g., ACO \cite{Dorigo-1991-acoe1}) from good
searched solutions, (3) keeping a set of good parameter settings
(e.g., SHADE \cite{Tanabe-2013}) are three useful ways to provide
information for improving the search accuracy.
Although all these {\xmetaha}s embed mechanisms to keep the
information of a certain number of searched solutions, searches are
still like looking for a needle in a haystack because they all lack
the knowledge of the exact landscape of the solution space or even a
rough outline.

\section{A Metaheuristic Algorithm with Vision of Solution Space}
To explain the basic idea of the proposed algorithm, named {\xsnoa}, a
simple example is given in \xfig{fig:overall-1} to show the details of
the major operators of {\xsnos}.  The basic idea of the proposed
algorithm is to use a mechanism to depict the landscape of the
solution space based on the searched solutions to enhance the search
performance of a {\xmetaha} for solving an optimization problem, just
like making a map to increase the chance to find the treasure in a new
place.  In the initialization operator, the space net that consists of
a set of elastic points $p$ will first be randomly generated in the
solution space, and its role is to fit the surface of the solution
space during the later convergence process.  By using the space net,
the elastic points can be used to divide the solution space into a set
of regions $r$, each of which is associated with an expected value to
describe the potential of finding a better solution in this region
compared with other regions so that a set of expected values $e$ will
also be generated.  Two populations $s$ and $x$ will also be randomly
generated in the solution space to play the roles of global search and
local search, respectively.  The proposed algorithm will then perform
the following operators---region search (RS), point search (PS), space
net adjustment (SNA), and population adjustment (PA)---repeatedly for
a certain number of iterations to find the solution.

\xfig{fig:overall-1} also shows that the \xe{\textbf{RS}} operator
will first select a set of regions (e.g., regions $r_4$, $r_5$, and
$r_6$) before the tournament selection kicks in to pick one of them
(e.g., region $r_6$) to be searched later.
In this example, because the objective value of the 8-th elastic point
$p_8$ is better than those of the other three elastic points $p_7$,
$p_{11}$, and $p_{12}$, the possible solutions will be generated
around the reference point, that is, elastic point $p_8$.  The
possible solutions can be generated by either moving a solution in the
population $s$ toward $p_8$ or moving $p_8$ toward a solution in the
population $s$.
Once a new solution $u_i$ is generated from $s_i$ to replace the
current solution $s_i$, the \xe{\textbf{SNA}} operator will then be
used to move the top-3 elastic points of region $r_6$ that is close to
$u_i$ toward this new solution. The space net will then be adjusted to
fit the population $s$.

The \xe{\textbf{PS}} operator is for the population $x$. The way a new
solution $v_i$ is generated is similar to that of RS, but the
reference point is now one of the top-$\rho$ elastic points (i.e., an
elastic point with its objective value among the best $\rho$ elastic
points).  The PS operator will work on the space net that has already
been adjusted by the RS operator to select a good elastic point (e.g.,
$p_4$) as the reference point. The possible solution can be either
based on a solution in the population $x$ moving toward the selected
elastic point or on the selected elastic point moving toward a
solution in the population $x$. The following process is similar to
but not exactly the same as the SNA after RS; that is, the
\xe{\textbf{SNA}} operator will then move the three elastic points
that are closest to the new solution $v_i$ even closer to this new
solution.

Unlike the RS and PS operators that are responsible for generating the
new solutions and adjusting the space net, the \xe{\textbf{PA}}
operator is used to reduce and expand the population sizes of $s$ and
$x$, as shown in the PA of \xfig{fig:overall-1}. Because the basic
idea of {\xsnos} is to use the population $s$ for global search and
the population $x$ for local search, the initial population size of
$s$ is set larger than the population size of $x$ to make it possible
for the searches of {\xsnos} to trend to global search in the early
stage of the convergence process. To make the searches of {\xsnos}
gradually trend to 50\% global search and 50\% local search and then
trend to local search in the later stage of the convergence process,
PA will delete a set of solution(s) in the population $s$ and also
expand the population $x$ by adding new solution(s), to adjust the
ratio of the population sizes of these two populations. That is why a
number of solution(s) $s_w$ that have the worst objective values in
the population $s$ will be randomly removed first, and then new
solution(s) $v_p$ generated around a good elastic point of the space
net will be added to the population $x$.

To further show the convergence process of {\xsnos}, the 2-D Ackley
function is used as the test optimization problem.
\xfig{fig:overall-2}(a) shows the landscape of the 2-D Ackley
function.  \xfiga{fig:overall-2}(b)--(d) show the elastic points $p$
of {\xsnos} after 400, 800, and 4,000 evaluations. Note that the
number of elastic points $p$ is set equal to 400 in this example.  As
shown in \xfig{fig:overall-2}(b), it can be easily seen that the
elastic points are randomly distributed in the solution space during
the early stage of the convergence process.  As shown in
\xfig{fig:overall-2}(c), after another 400 evaluations, the
distribution of the elastic points looks now a little bit like the
landscape of the 2-D Ackley function.  As shown in
\xfig{fig:overall-2}(d), after 4,000 evaluations, the distribution of
elastic points is now much more like the landscape of the 2-D Ackley
function. It is not hard to imagine that {\xsnos} now knows roughly
the whole solution space, and where good solutions are likely to be
found and where they are not.  This also means that {\xsnos} has a
full vision of the entire landscape of the solution space that the
other {\xmetaha}s do not have and thus can only base the searches on
feeble vision.

\section{Results}
\label{sec2}
The empirical analysis is conducted on a PC with two Intel Xeon
E5-2620v4 CPUs (2.10GHz, 20 MB cache, and 8 cores)  and 16 GB of memory
running Linux Ubuntu 18.04 x86\_64, and the
programs are written in C++ and compiled using g++.
To evaluate the performance of the proposed algorithm (\xsnos), we
compare it with L-SHADE \cite{Tanabe-2014}, jSO \cite{Brest-2017},
j2020 \cite{Brest-2020}, NL-SHADE-RSP \cite{StanovovAS21},
NL-SHADE-LBC \cite{Stanovov-2022}, and S-LSHADE-DP \cite{Van-2022} for
two well-known single objective bound constrained problem (SOP)
benchmarks, CEC2021 \cite{cec2021} and CEC2022 \cite{cec2022}.  The
parameter settings of L-SHADE, jSO, j2020, NL-SHADE-RSP, NL-SHADE-LBC,
and S-LSHADE-DP are based on \cite{Tanabe-2014, Brest-2017,
  Brest-2020, StanovovAS21, Stanovov-2022, Van-2022}, respectively.
The parameter settings of {\xsnos} are based on the search results of
tree-structured Parzen estimator (TPE) \cite{bergstra2011,
  akiba2019optuna}. For more details of {\xsnos}, please see
\xsec{ssec:algorithm}.

\subsection{Comparisons of {\xsnos} and Other Search Algorithms}
\label{ssec:results}

\begin{figure}[tbh] \centering \footnotesize \setlength\tabcolsep{1pt}
  \includegraphics[width=0.47\textwidth]{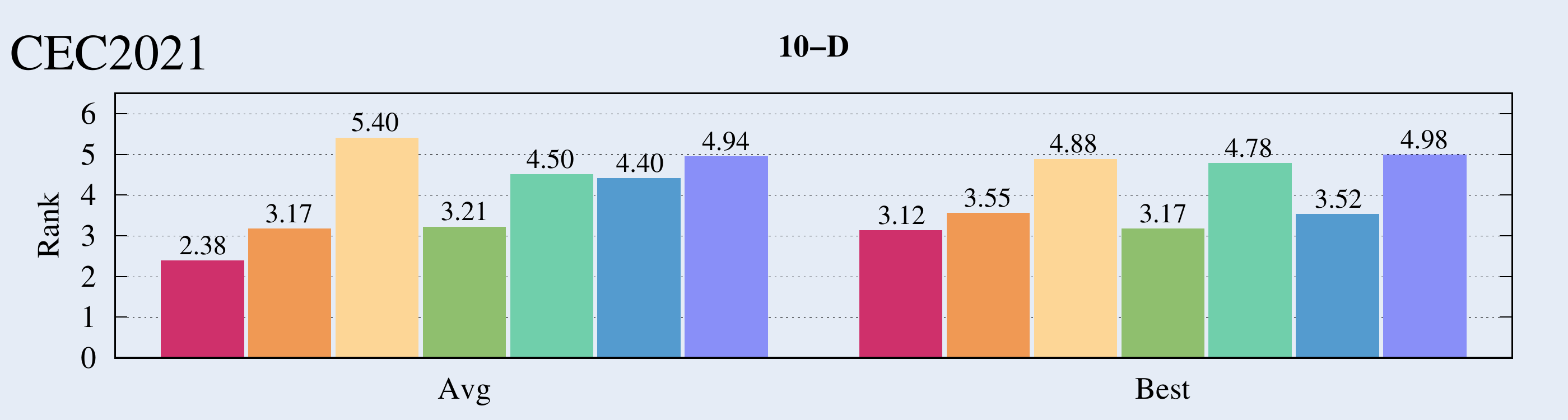}\\
  \vspace{-0.15\baselineskip}
  \includegraphics[width=0.47\textwidth]{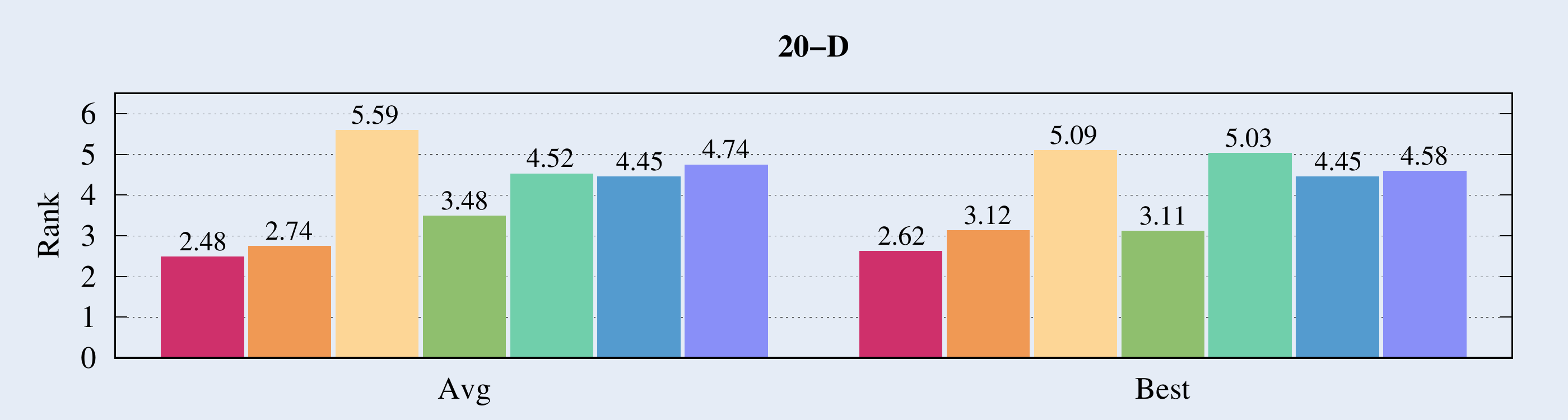}\\
  \includegraphics[width=0.47\textwidth]{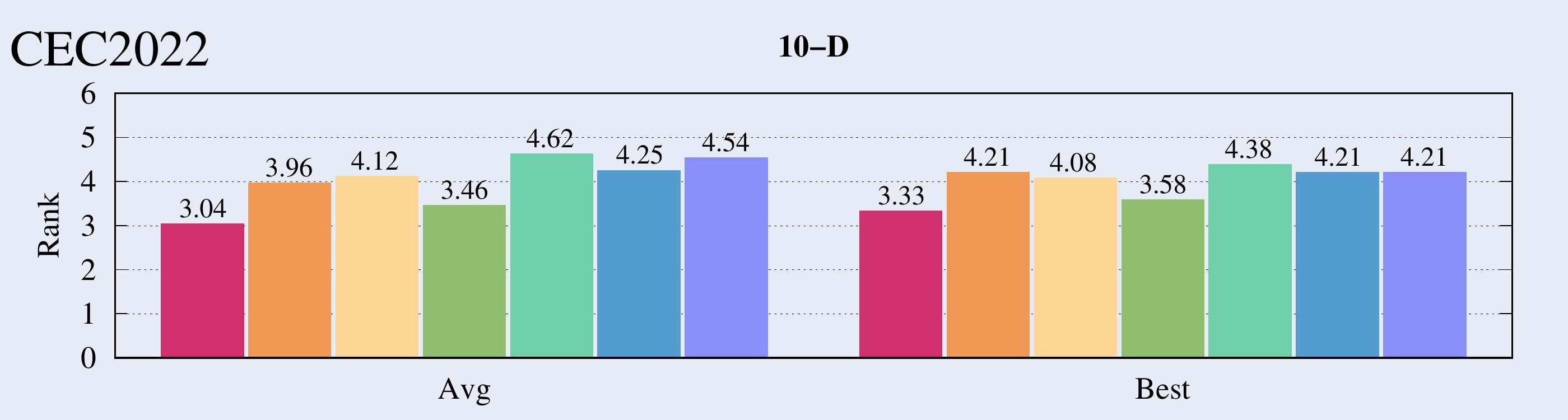}\\
  \vspace{-0.15\baselineskip}
  \includegraphics[width=0.47\textwidth]{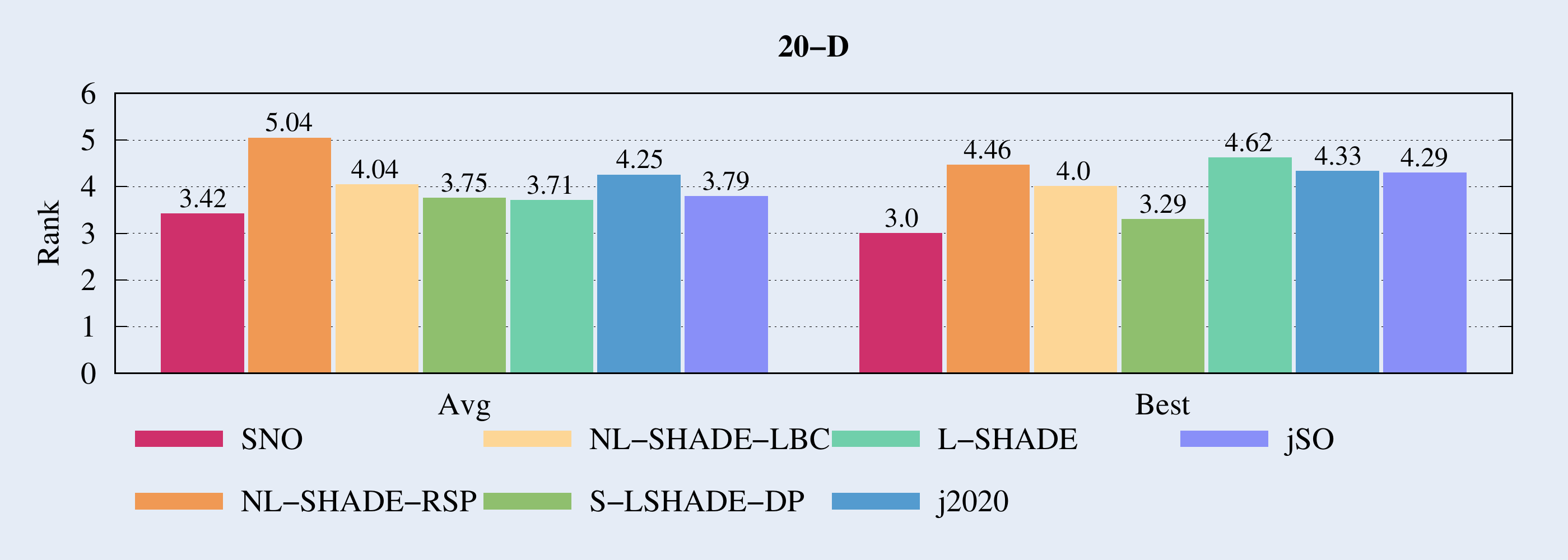}\\
  \caption{The average rank of each algorithm for CEC2021 and
CEC2022.}
\label{fig:rank22}
\end{figure}

\xfig{fig:rank22} gives the comparisons between the proposed algorithm
and other state-of-the-art search algorithms for solving CEC2021
\cite{cec2021} and CEC2022 \cite{cec2022}.
They are the average rank of a search algorithm in terms of (1) the
average rank of an algorithm among all algorithms for all given test
functions in all trials, i.e., CEC2021 (Avg) and CEC2022 (Avg), and
(2) the average rank of an algorithm among all algorithms for all
given test functions in the best trial, i.e., CEC2021 (Best) and
CEC2022 (Best), which are also usually used to evaluate the
performance of a search algorithm for solving single objective bound
constrained problems.  Note that 10-D and 20-D represent 10 dimensions
and 20 dimensions, respectively.  The results show that (\xsnos) can
get the best results for CEC2021 and CEC2022, which also imply that
(\xsnos) can find the best results compared to all other search
algorithms evaluated in this study for most test functions using the
same maximum number of function evaluations (MaxFES).

\begin{figure}[tbh] \centering \footnotesize \setlength\tabcolsep{1pt}
\includegraphics[width=0.47\textwidth]{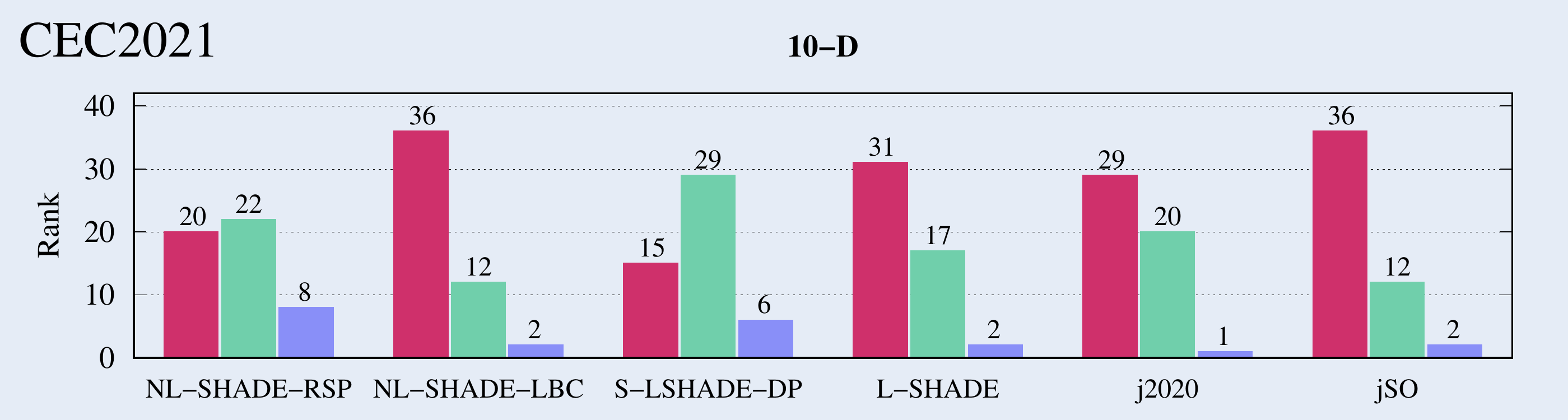}\\
\vspace{-0.5\baselineskip}
\includegraphics[width=0.47\textwidth]{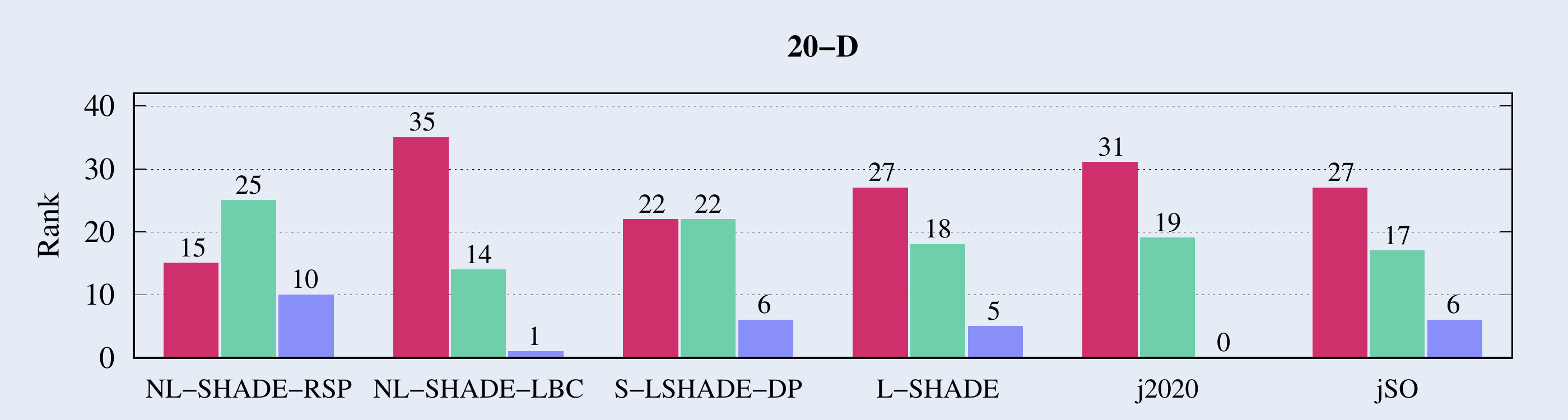}\\
  \label{fig:rank21}
\includegraphics[width=0.47\textwidth]{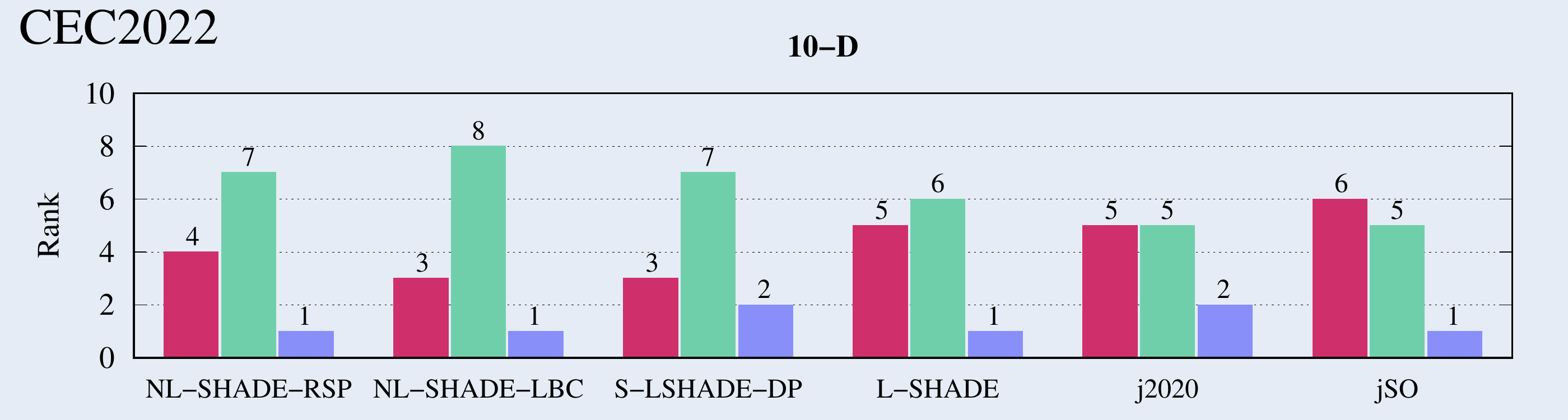}\\
\vspace{-0.5\baselineskip}
\includegraphics[width=0.47\textwidth]{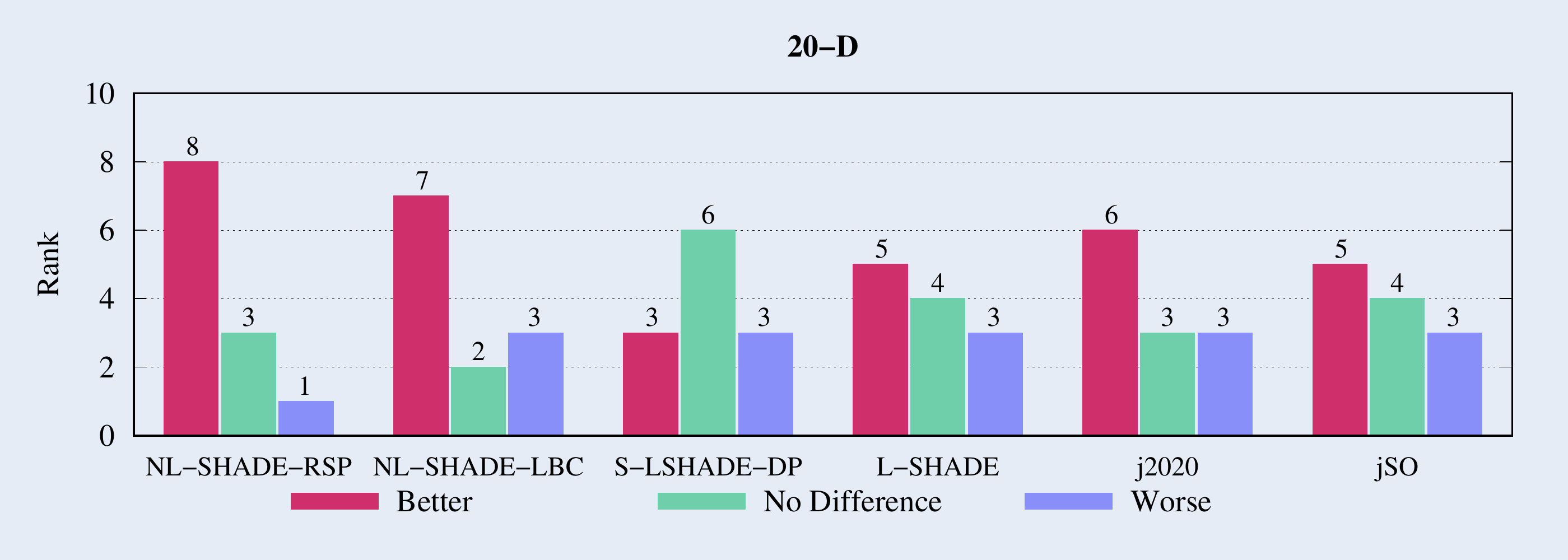}\\
  \caption{The comparisons between {\xsnos} and the other search algorithms for CEC2021 and CEC2022 in terms of the Wilcoxon test.}
  \label{fig:cec2122_Wilcoxon}
\end{figure}

\begin{figure*}[tbh]
  \centering
  \footnotesize
  \setlength\tabcolsep{1pt}
  \begin{tabular}{ccccc}
    \includegraphics[width=0.2\textwidth]{gnuplot/Ackley.pdf}&
    \includegraphics[width=0.2\textwidth]{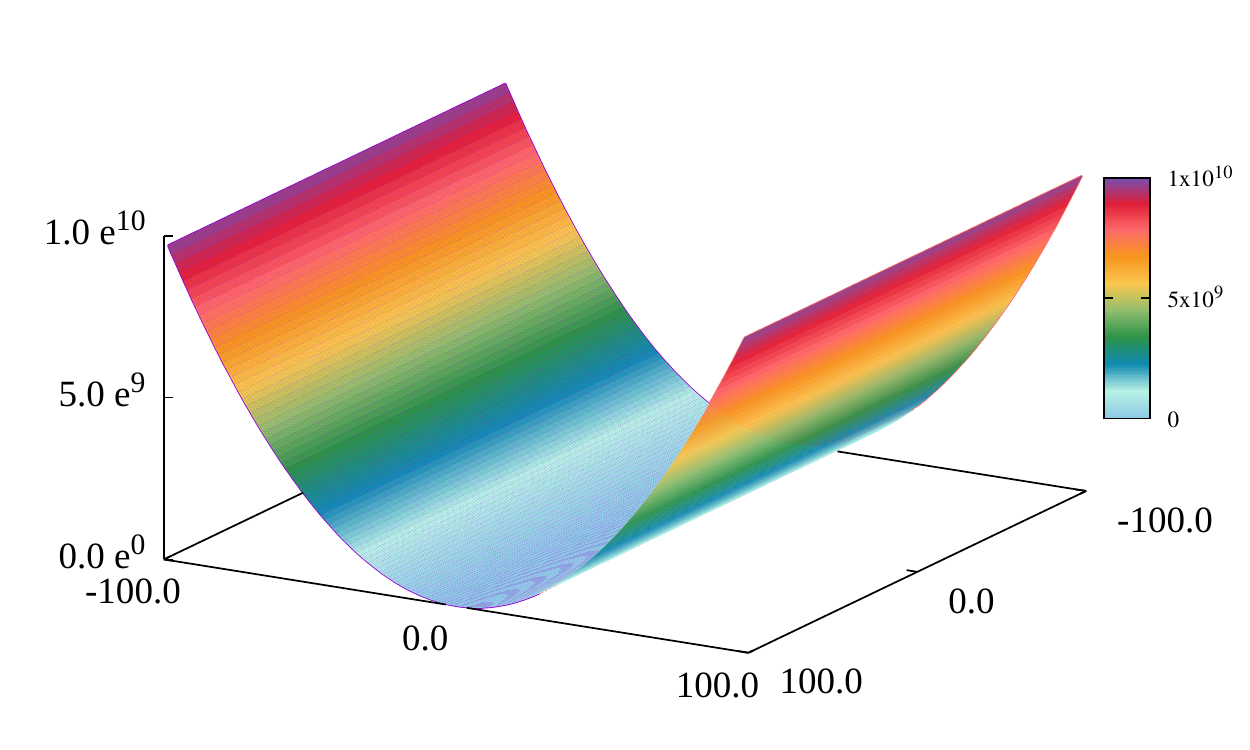} &
    \includegraphics[width=0.2\textwidth]{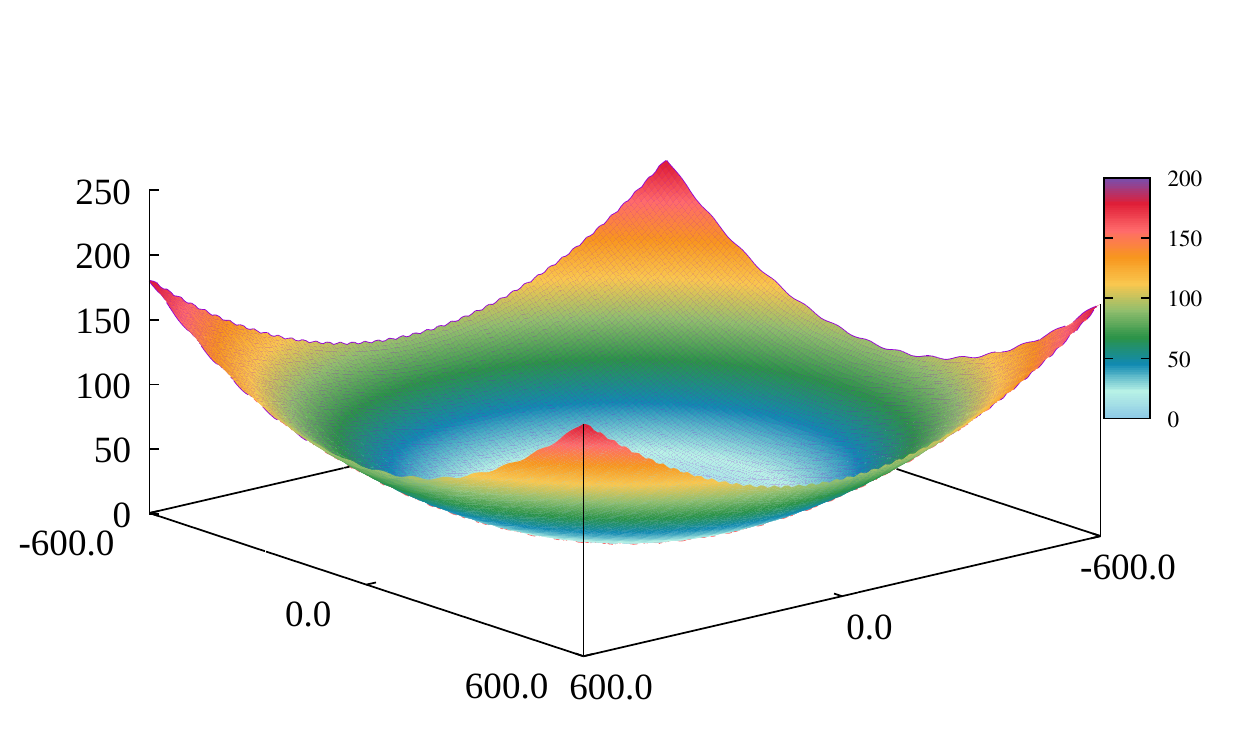} &
    \includegraphics[width=0.2\textwidth]{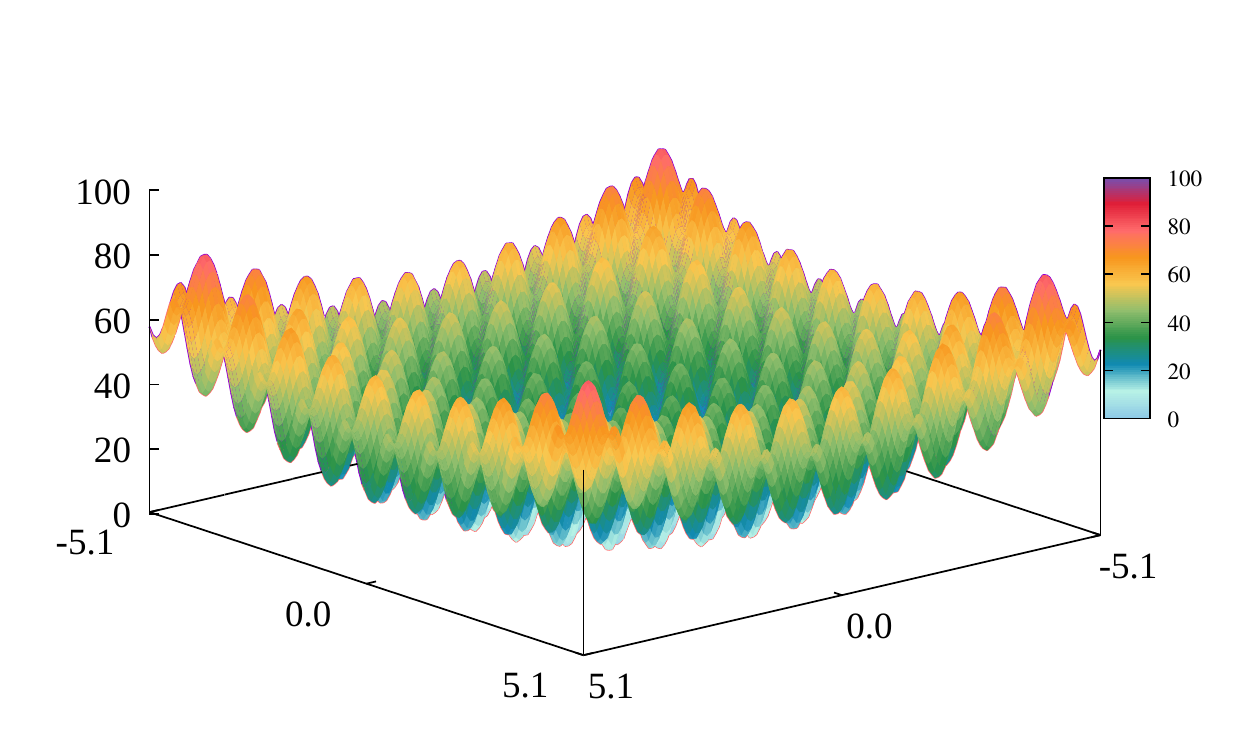} &
    \includegraphics[width=0.2\textwidth]{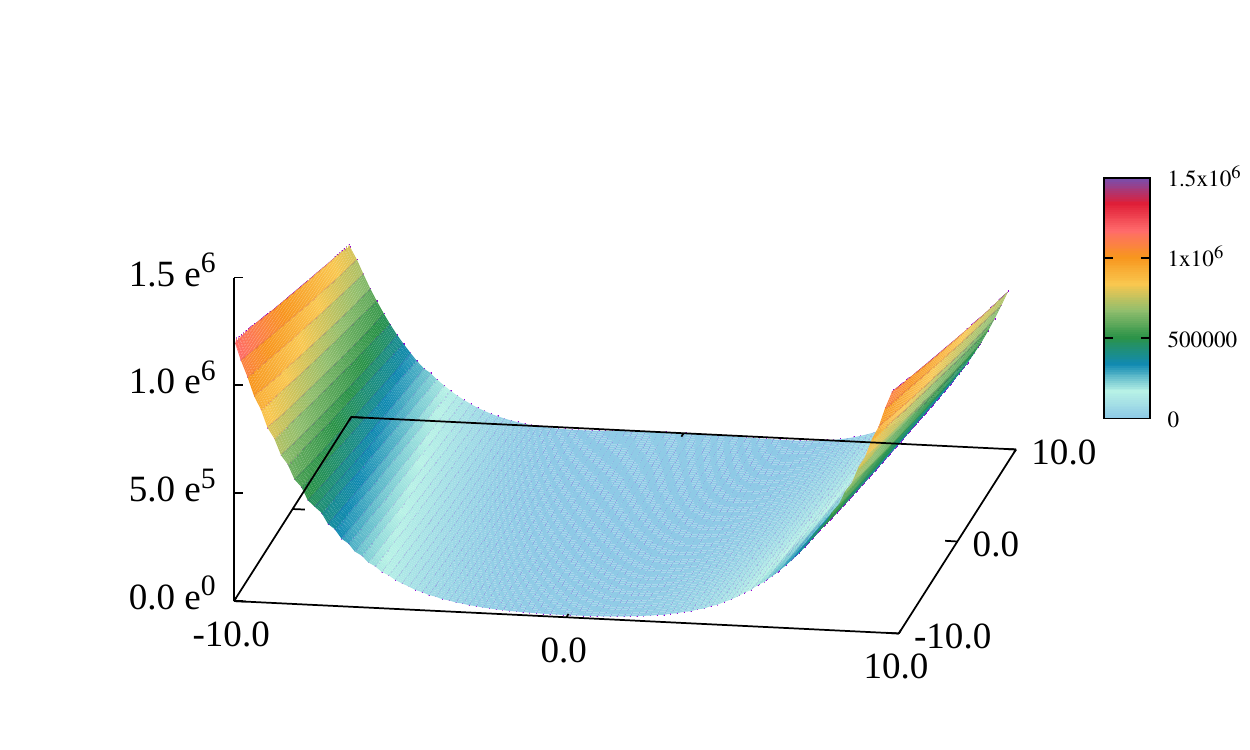} \\                        
    \includegraphics[width=0.2\textwidth]{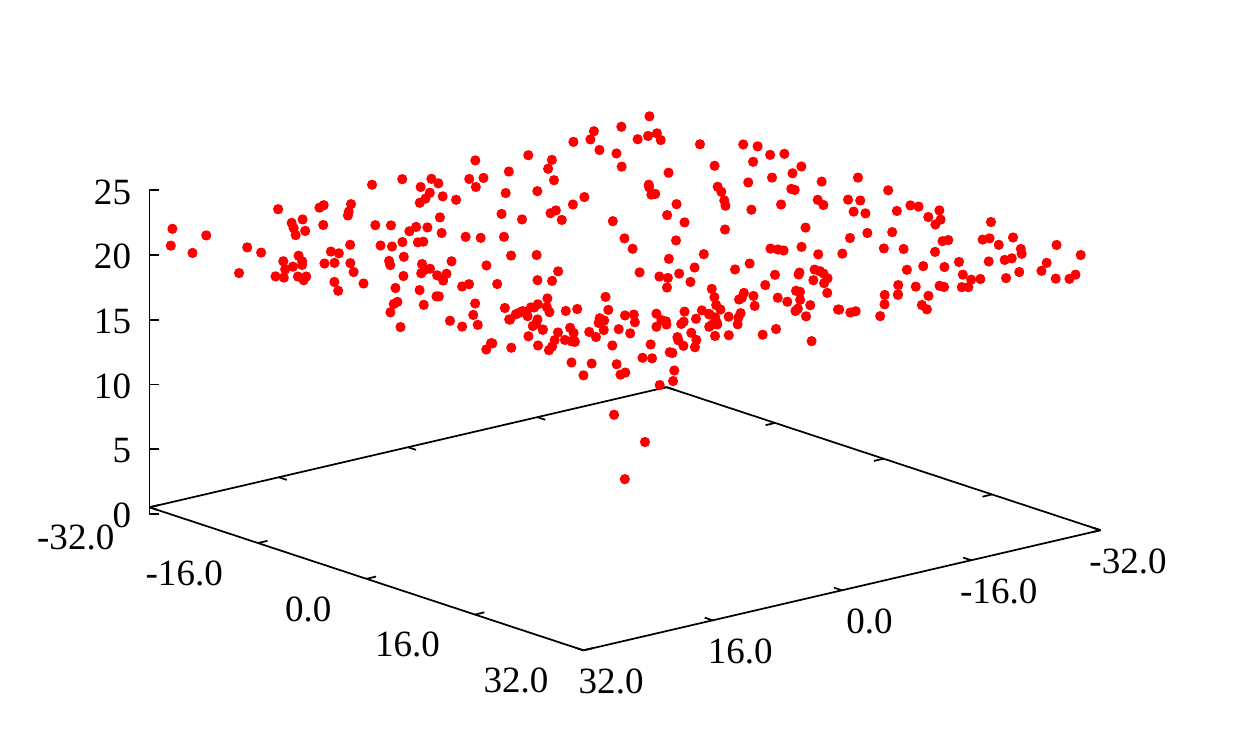} &
    \includegraphics[width=0.2\textwidth]{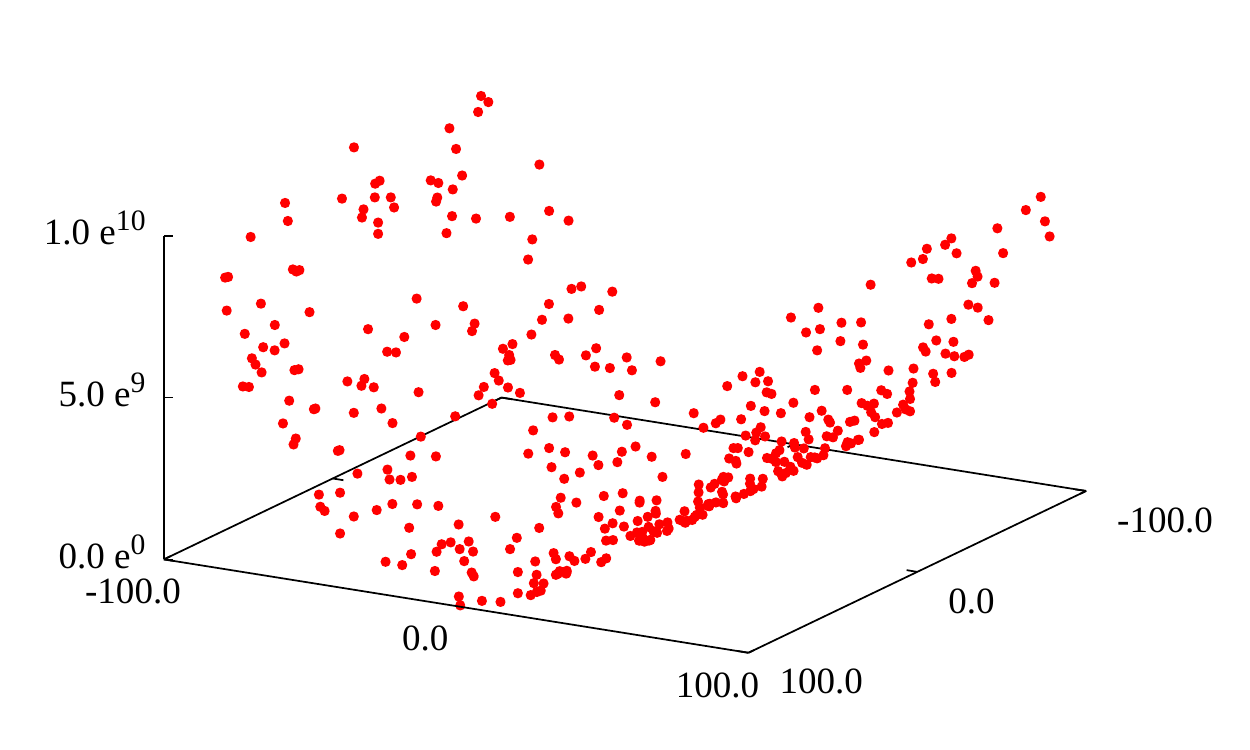} &
    \includegraphics[width=0.2\textwidth]{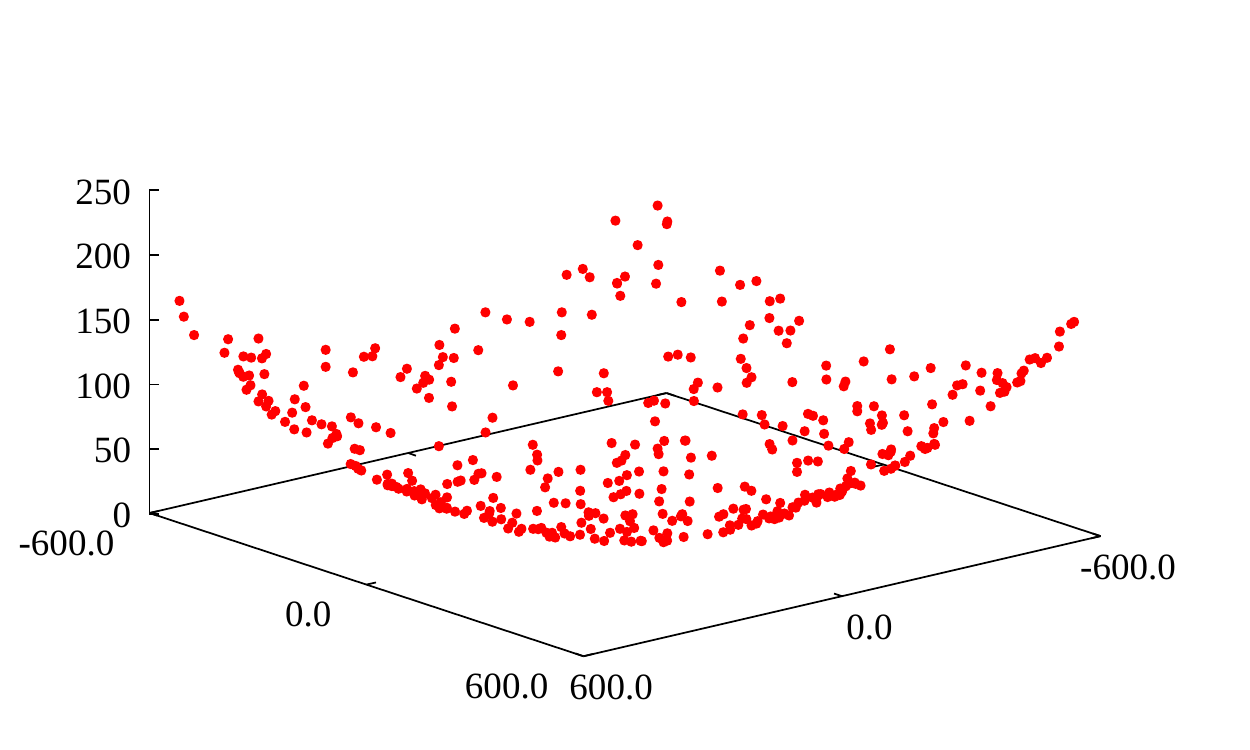} &
    \includegraphics[width=0.2\textwidth]{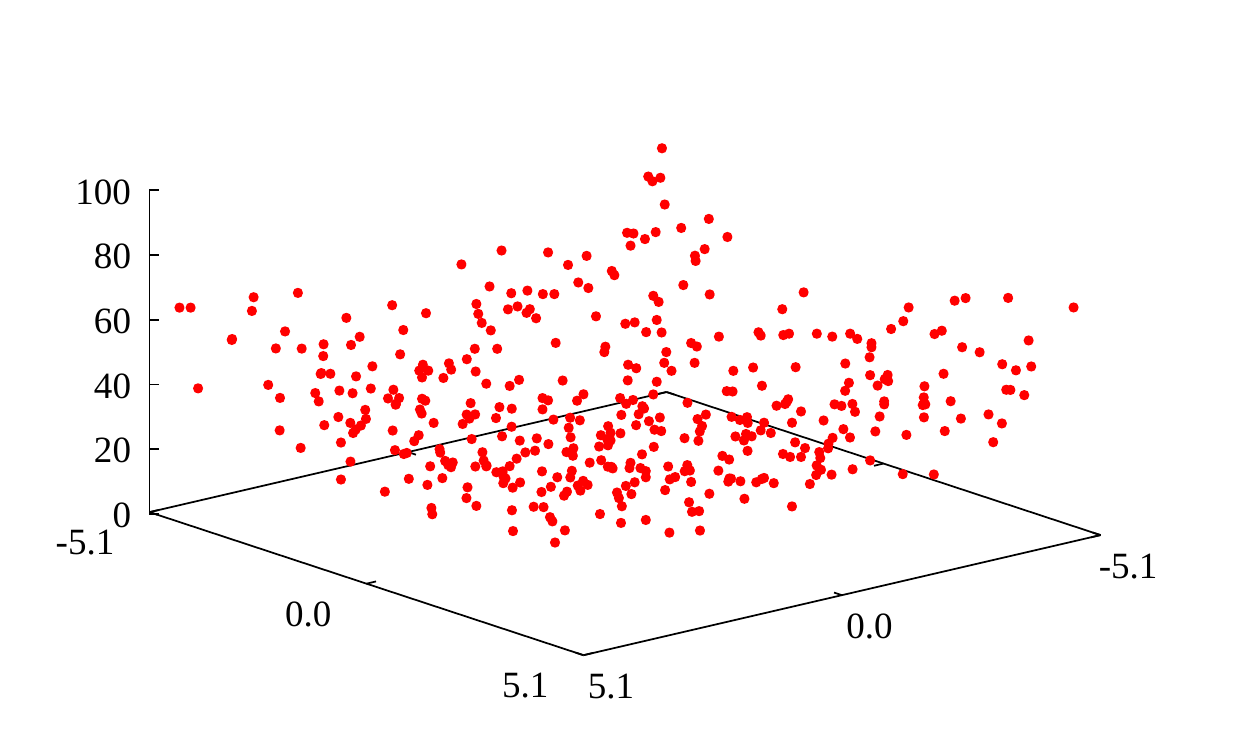} &
    \includegraphics[width=0.2\textwidth]{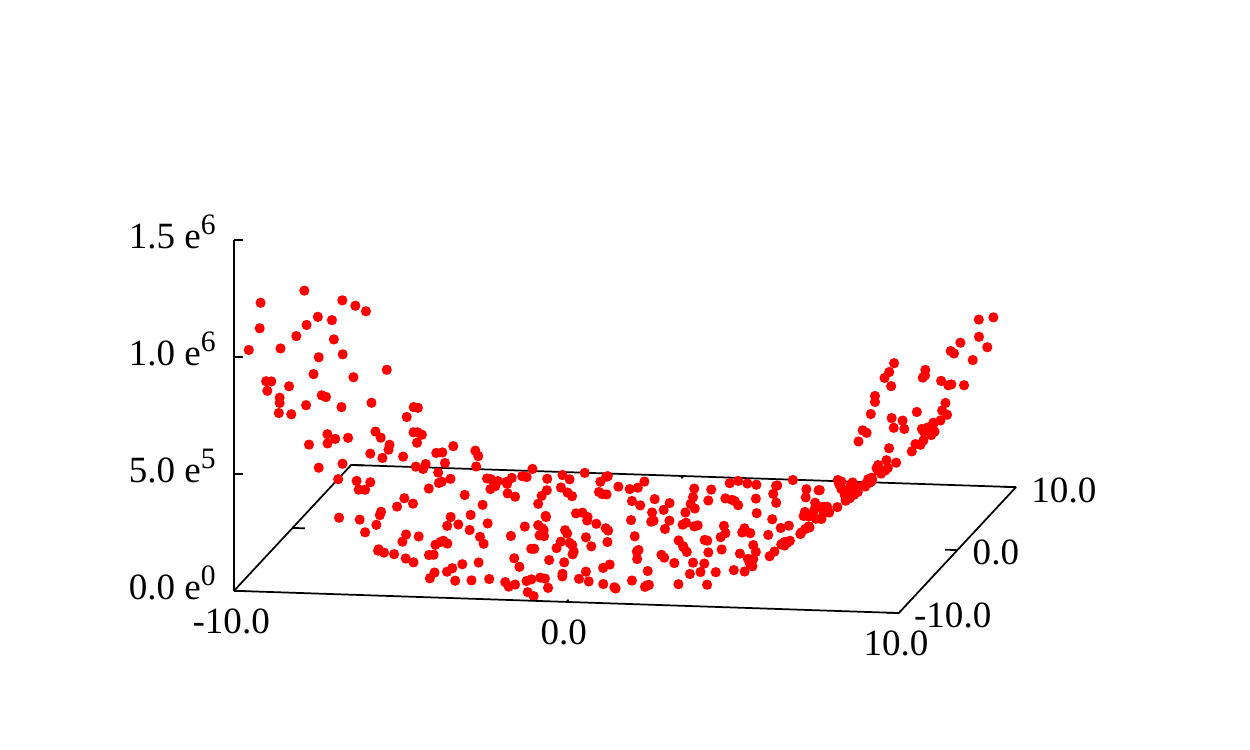} \\
    (a) & (b) & (c) & (d) & (e)\\    
    \includegraphics[width=0.2\textwidth]{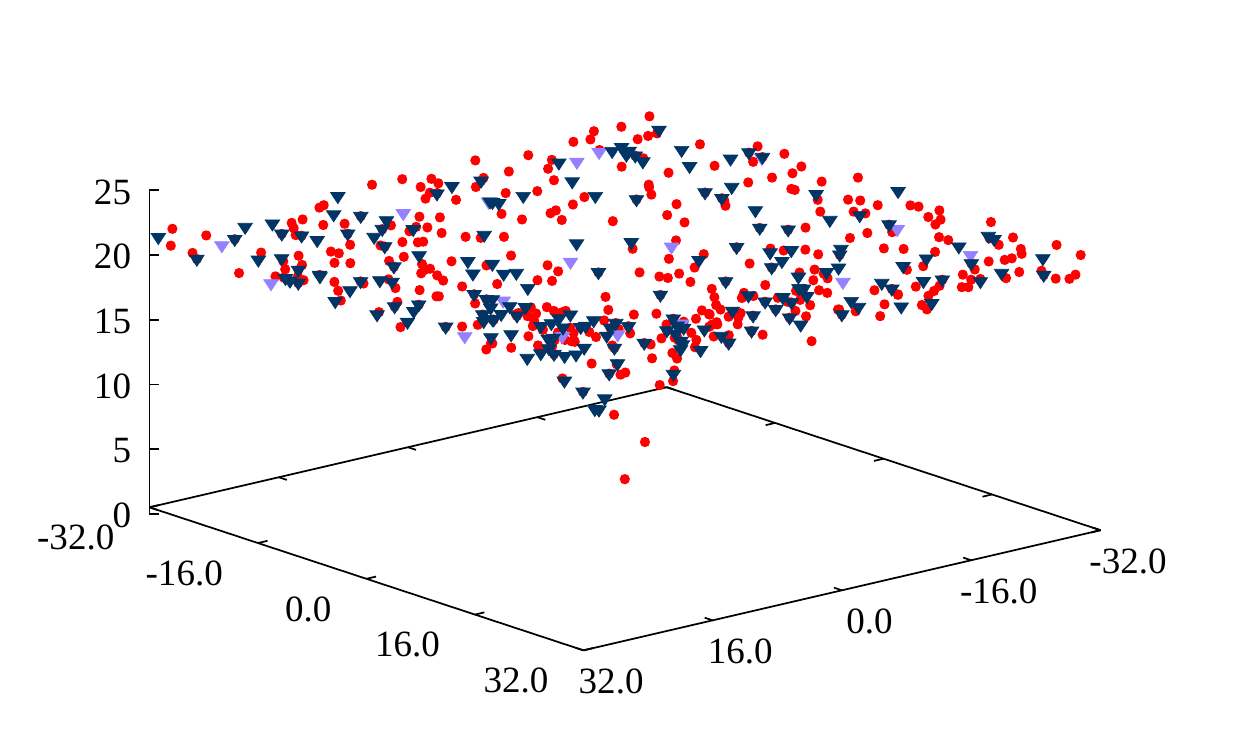} &
    \includegraphics[width=0.2\textwidth]{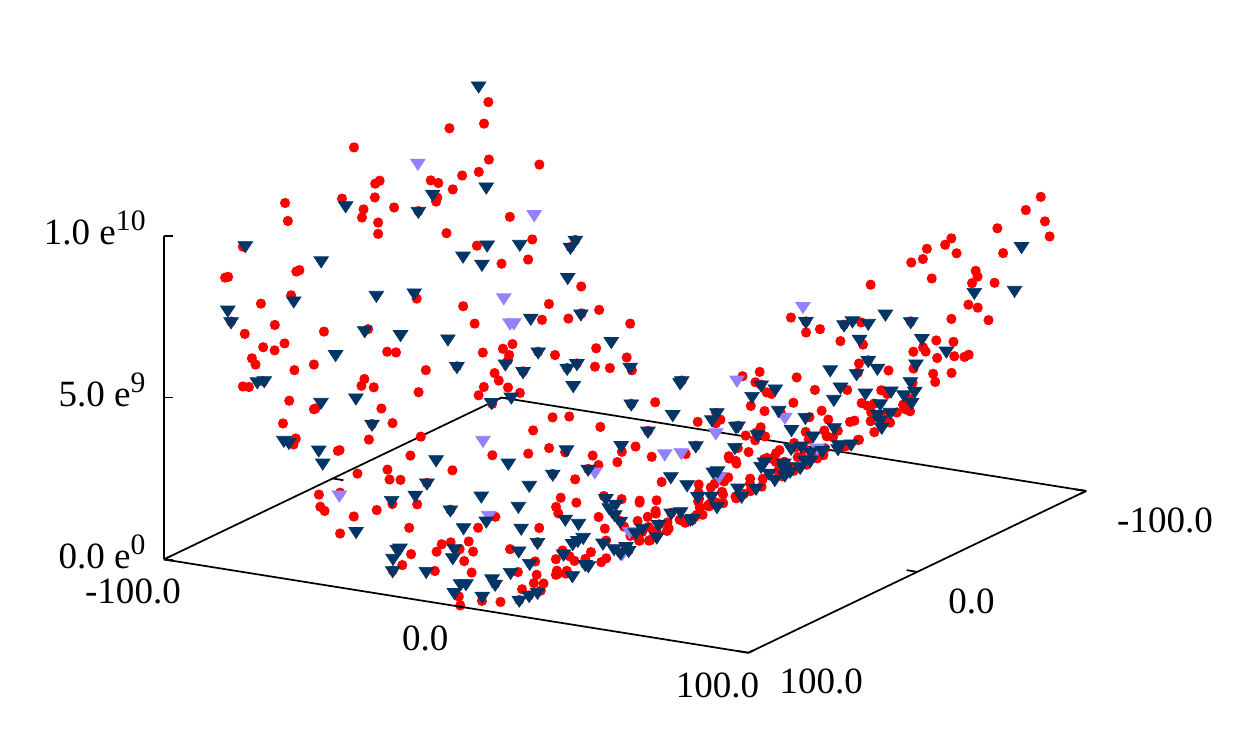} &
    \includegraphics[width=0.2\textwidth]{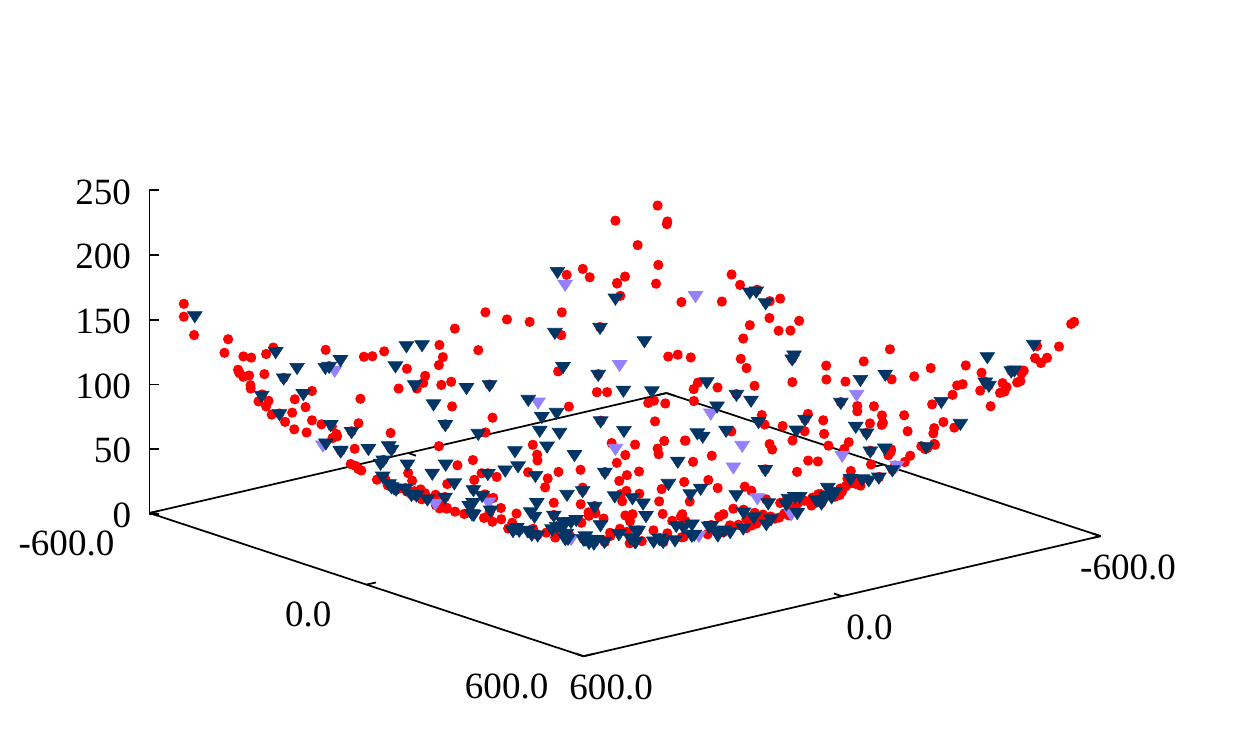} &
    \includegraphics[width=0.2\textwidth]{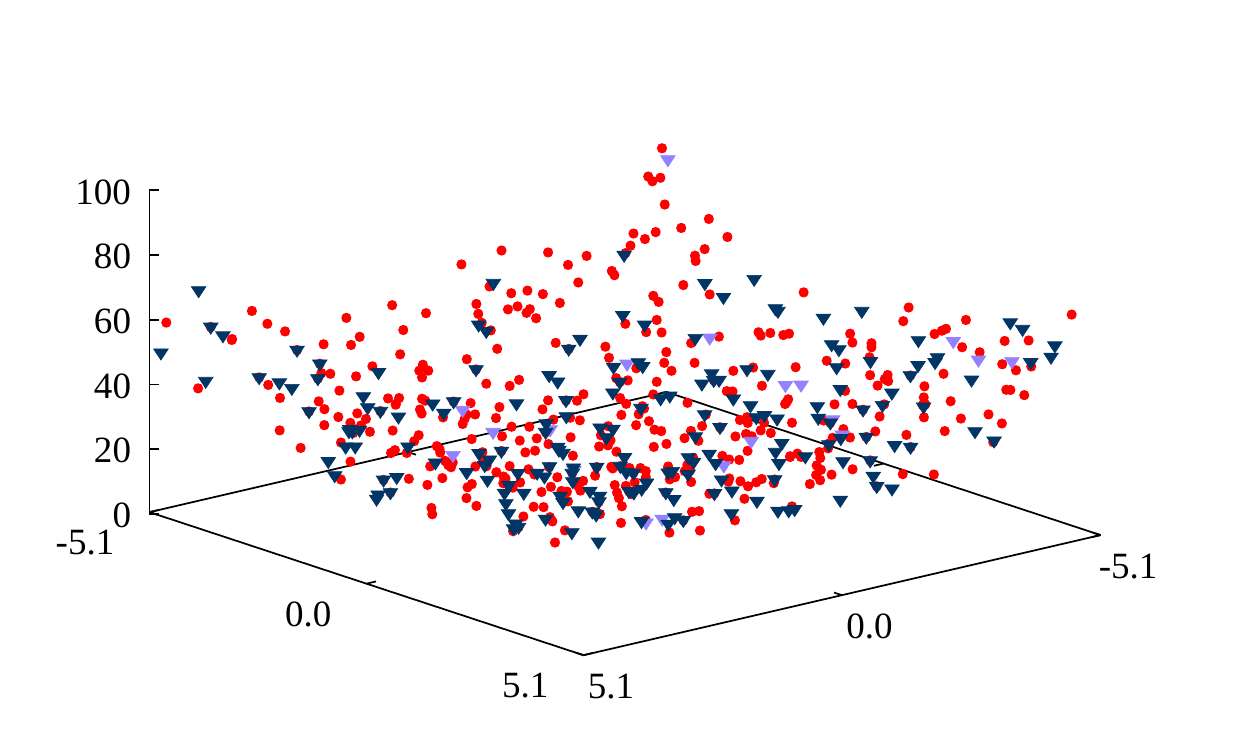} &
    \includegraphics[width=0.2\textwidth]{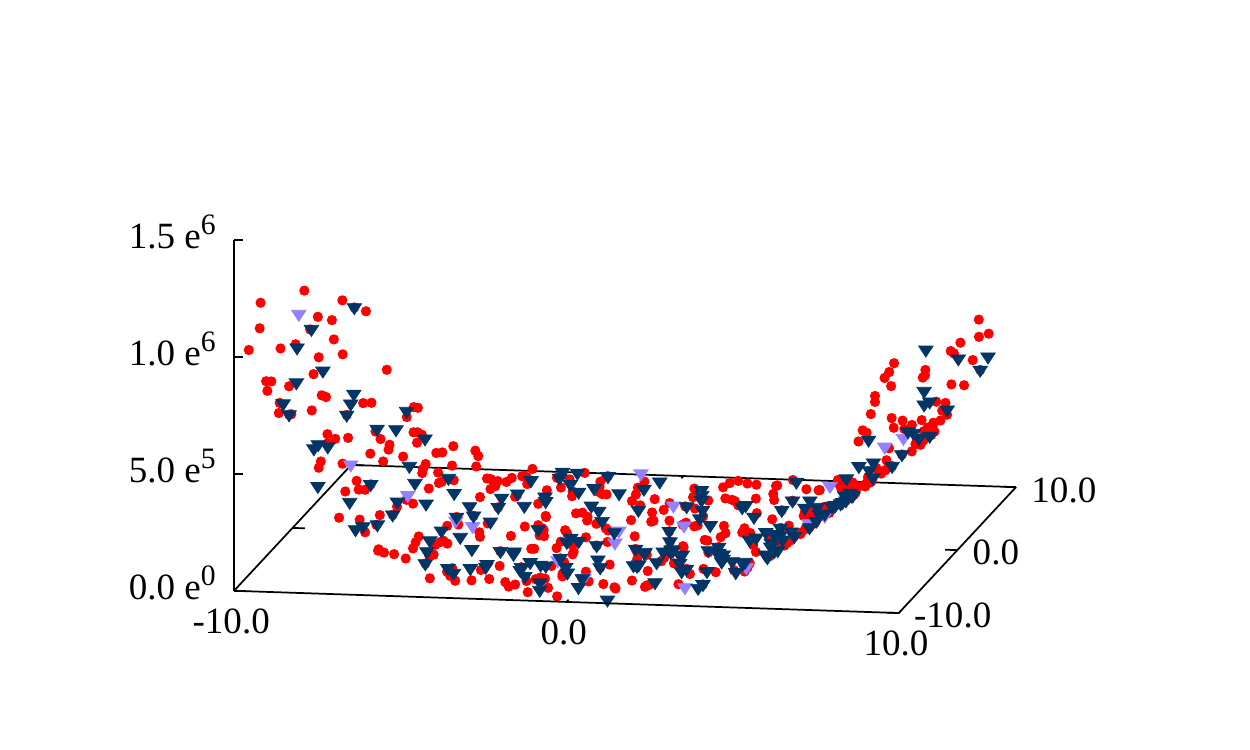}\\
    (f) & (g) & (h)  & (i) & (j)\\
    \includegraphics[width=0.2\textwidth]{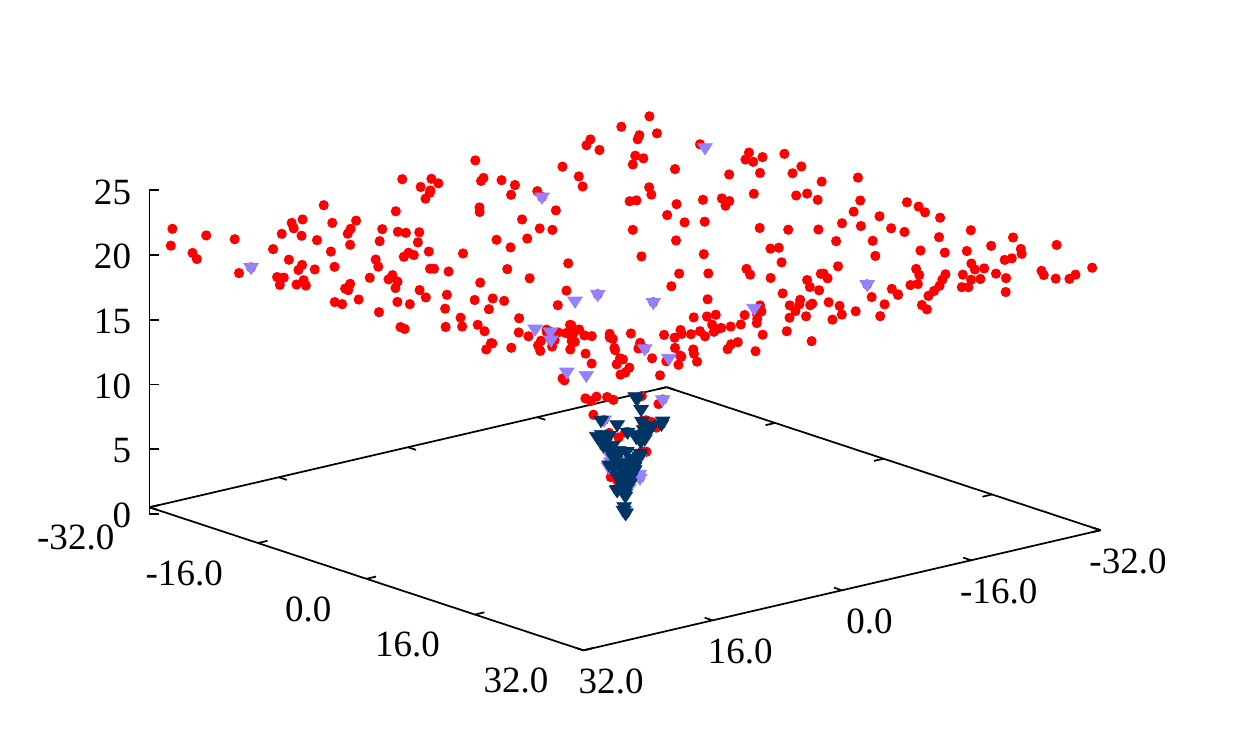} &
    \includegraphics[width=0.2\textwidth]{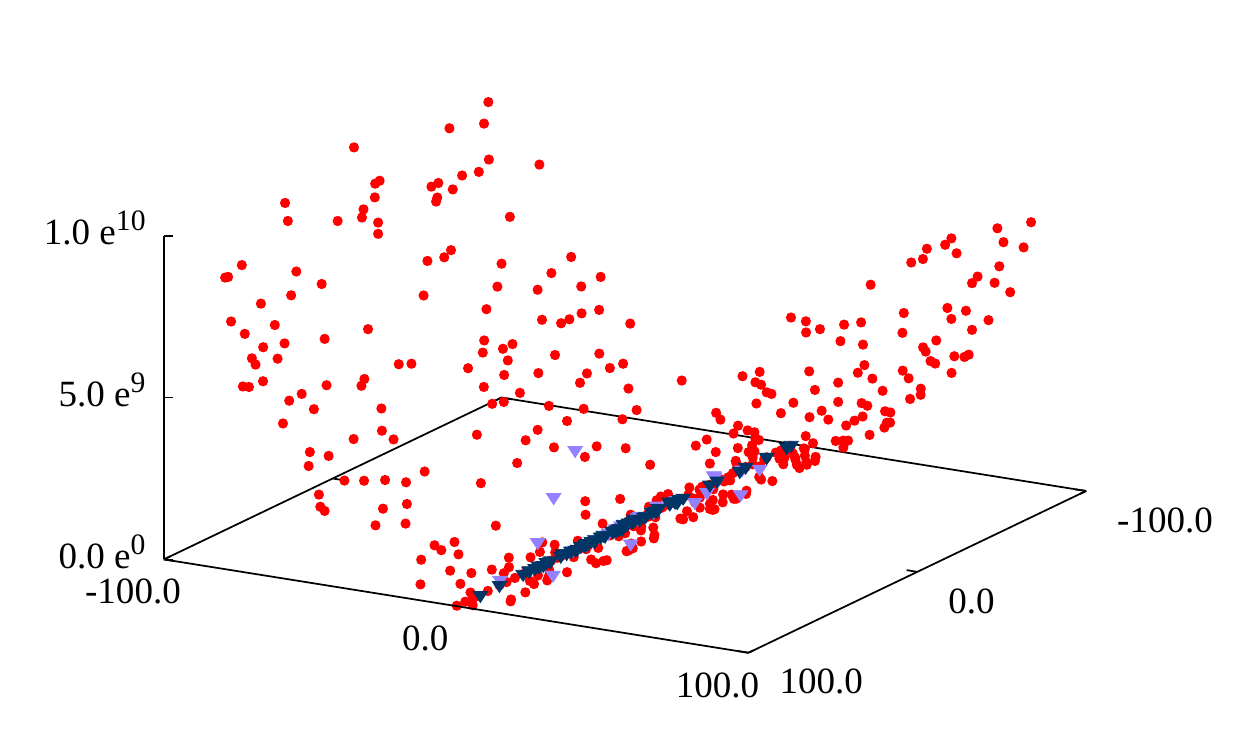} &
    \includegraphics[width=0.2\textwidth]{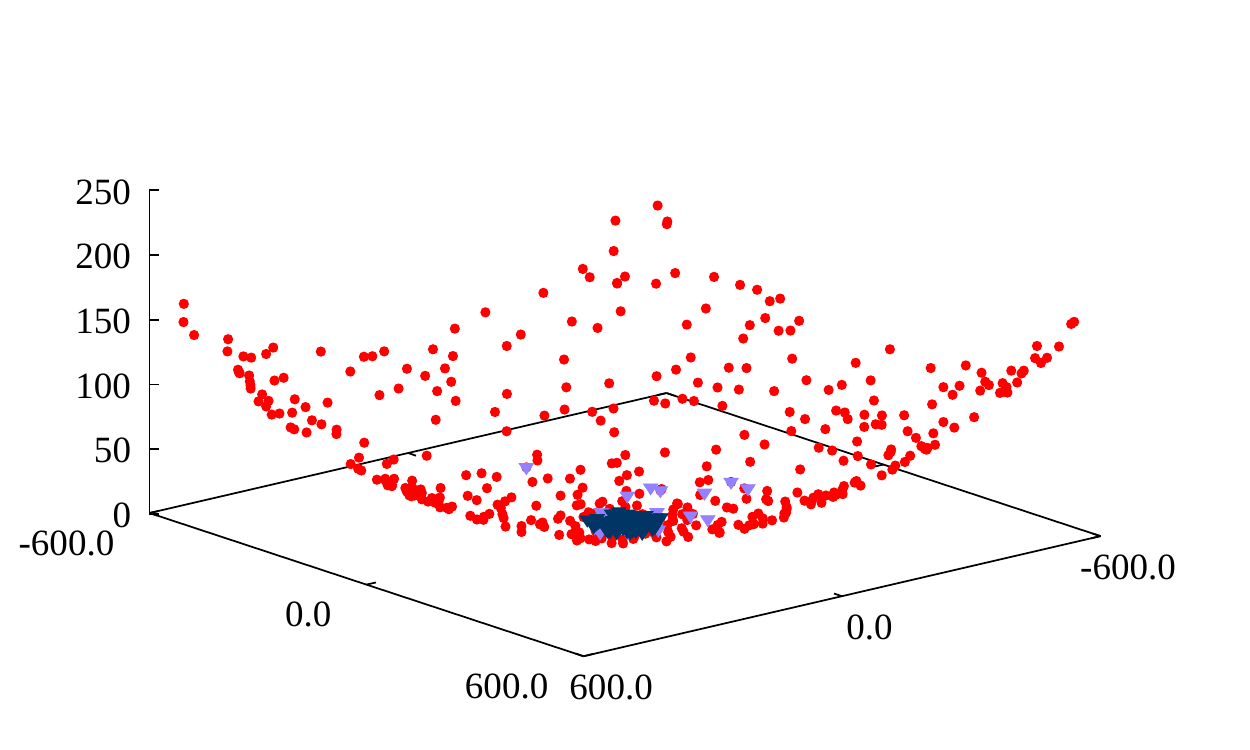} &
    \includegraphics[width=0.2\textwidth]{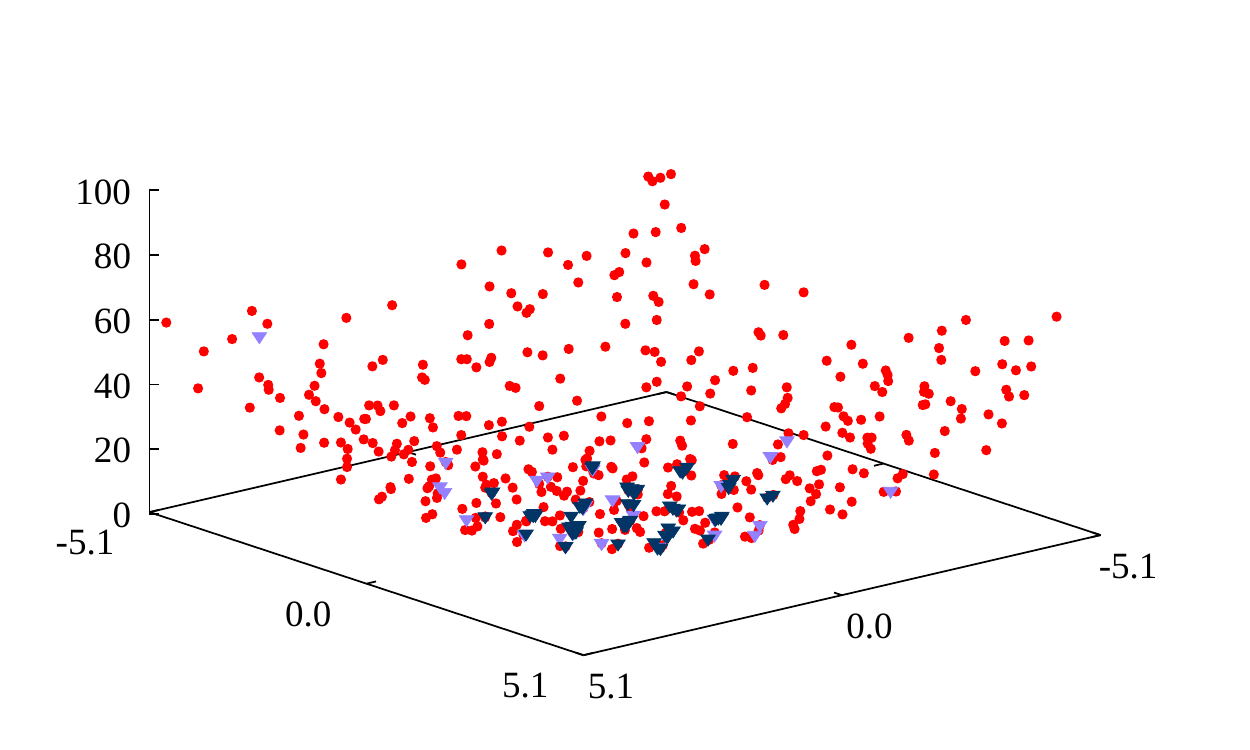} &
    \includegraphics[width=0.2\textwidth]{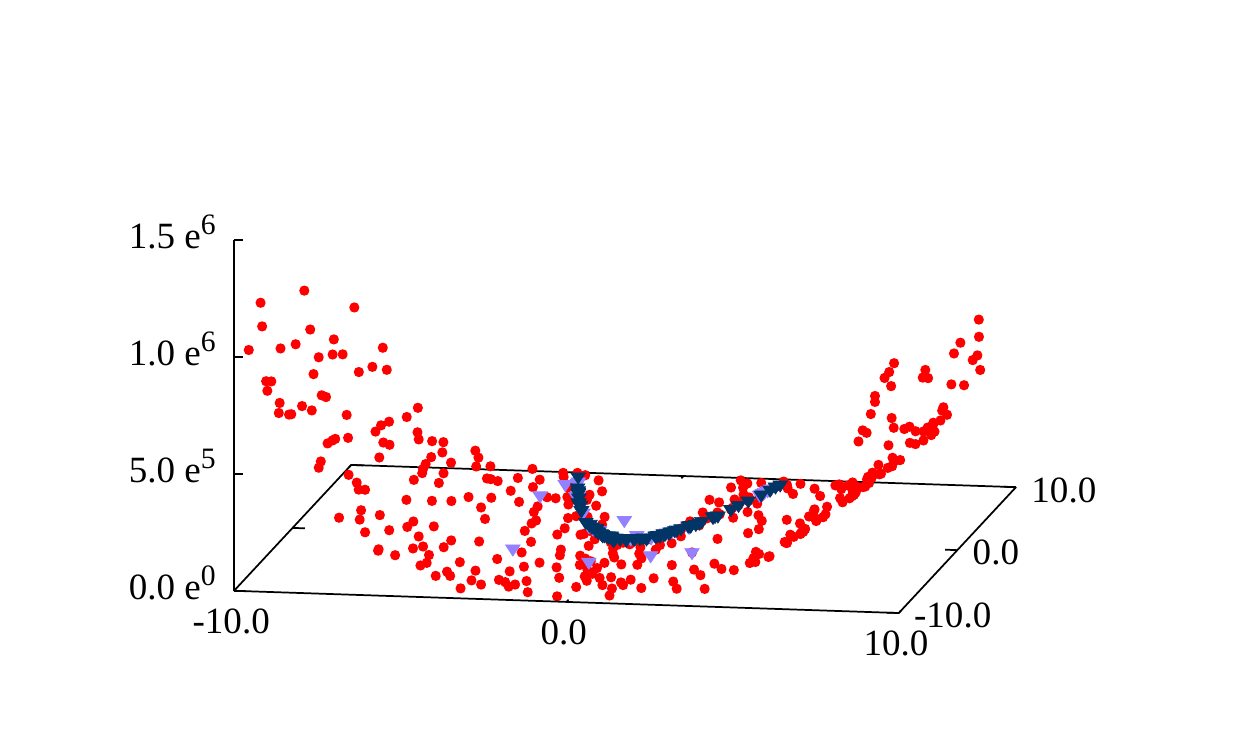} \\
    (k) & (l) & (m) & (n) & (o)\\
    \includegraphics[width=0.2\textwidth]{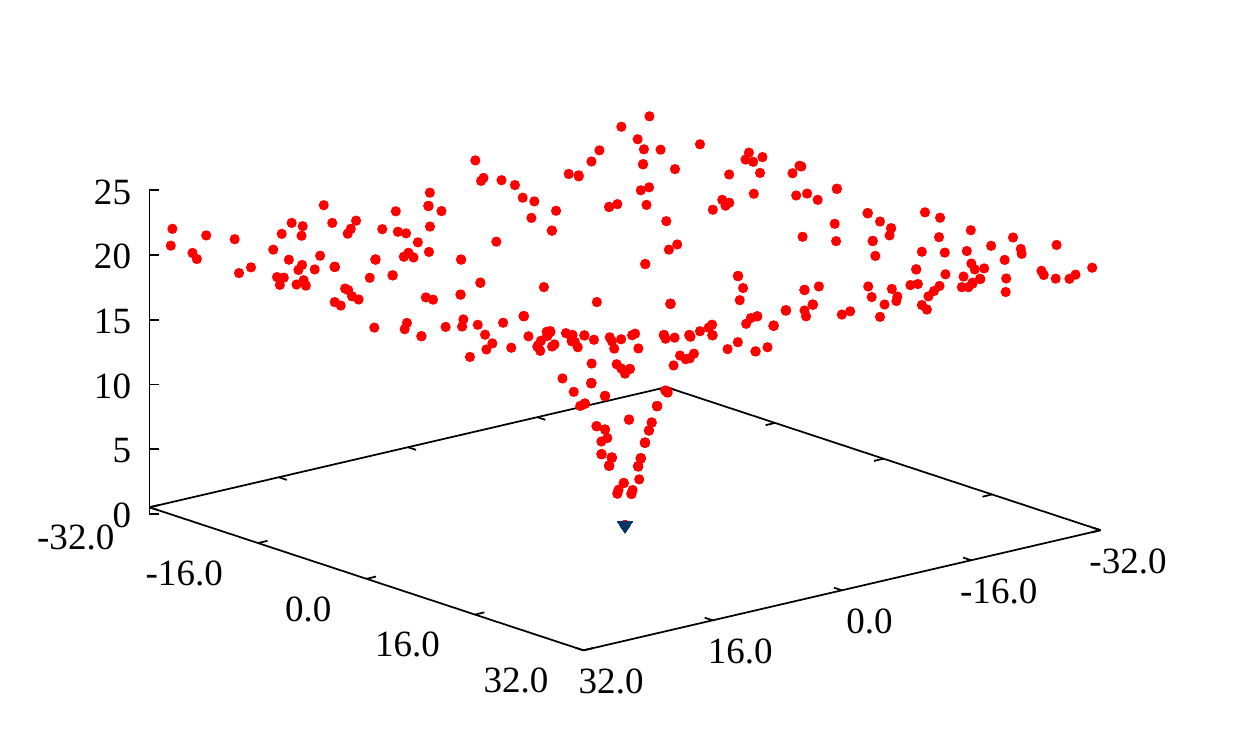} &
    \includegraphics[width=0.2\textwidth]{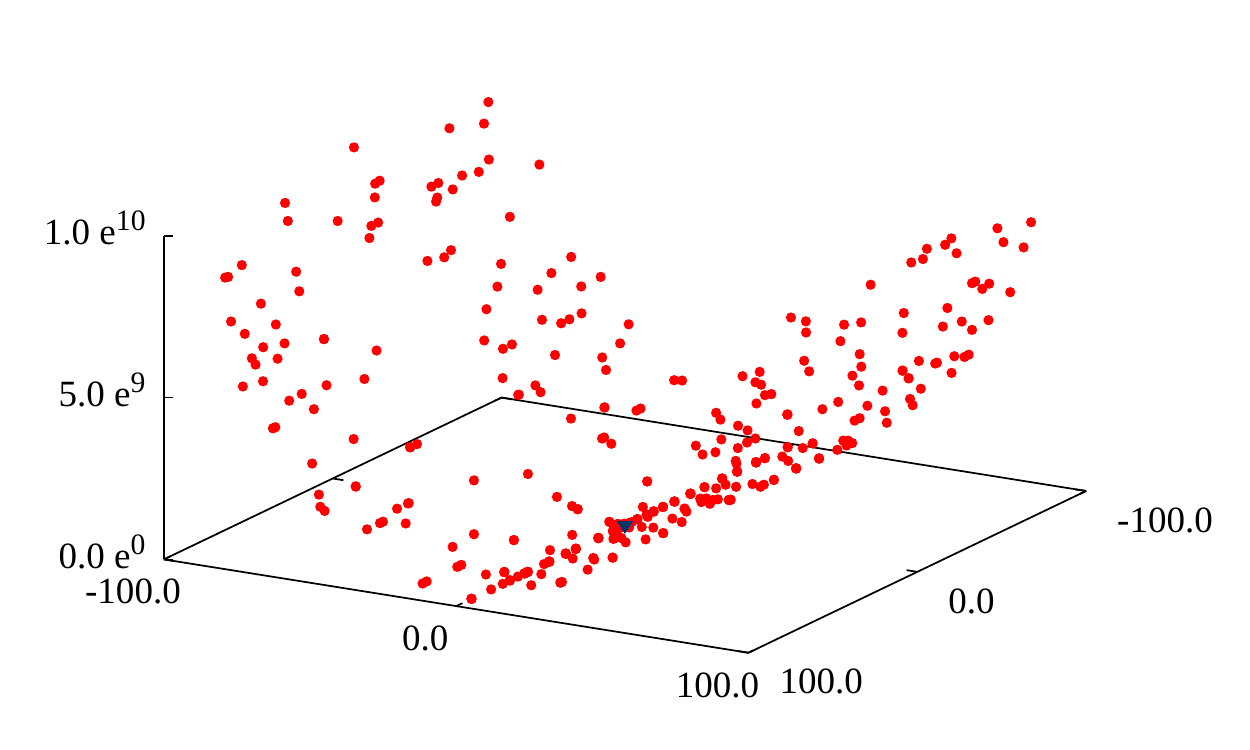} &
    \includegraphics[width=0.2\textwidth]{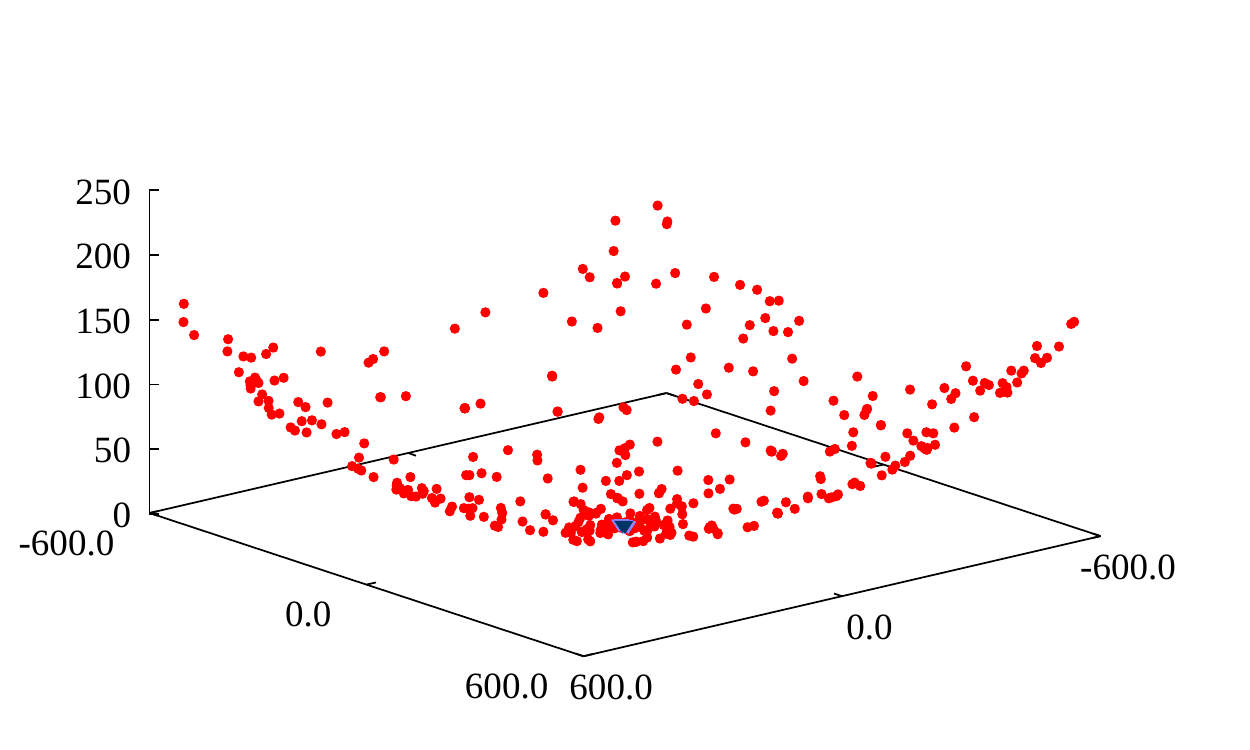} &
    \includegraphics[width=0.2\textwidth]{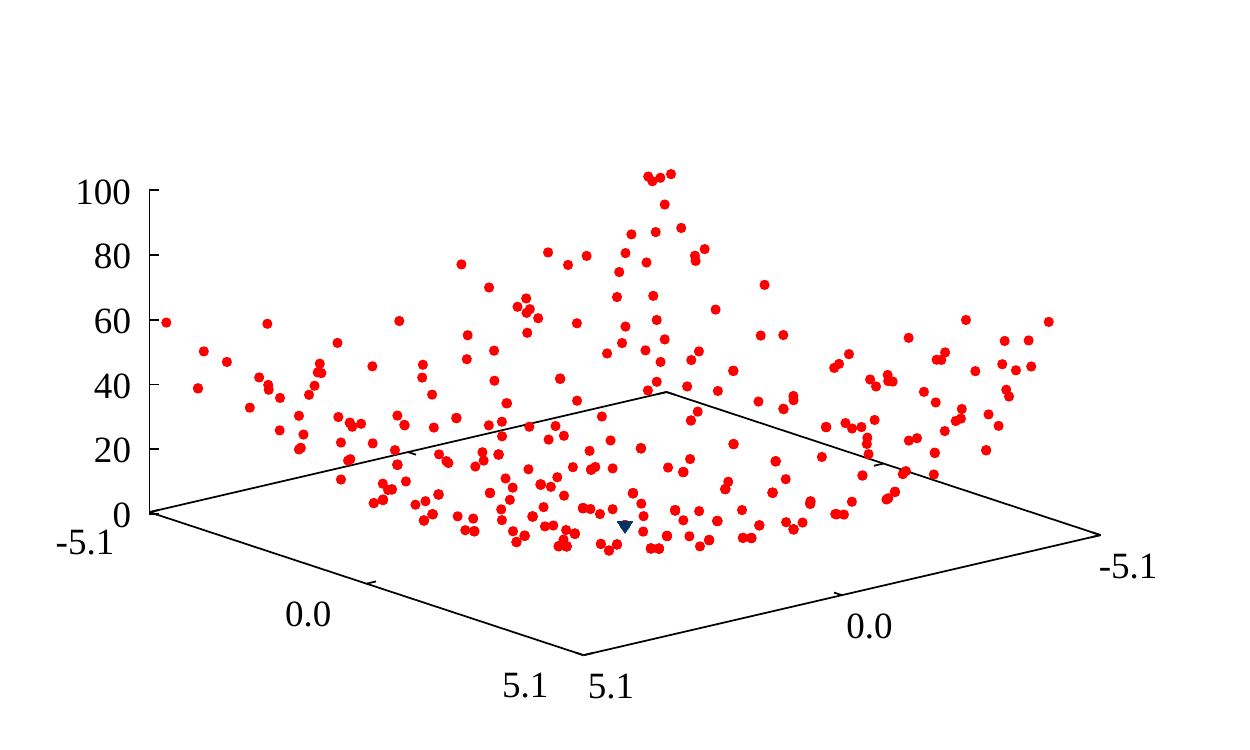} &
    \includegraphics[width=0.2\textwidth]{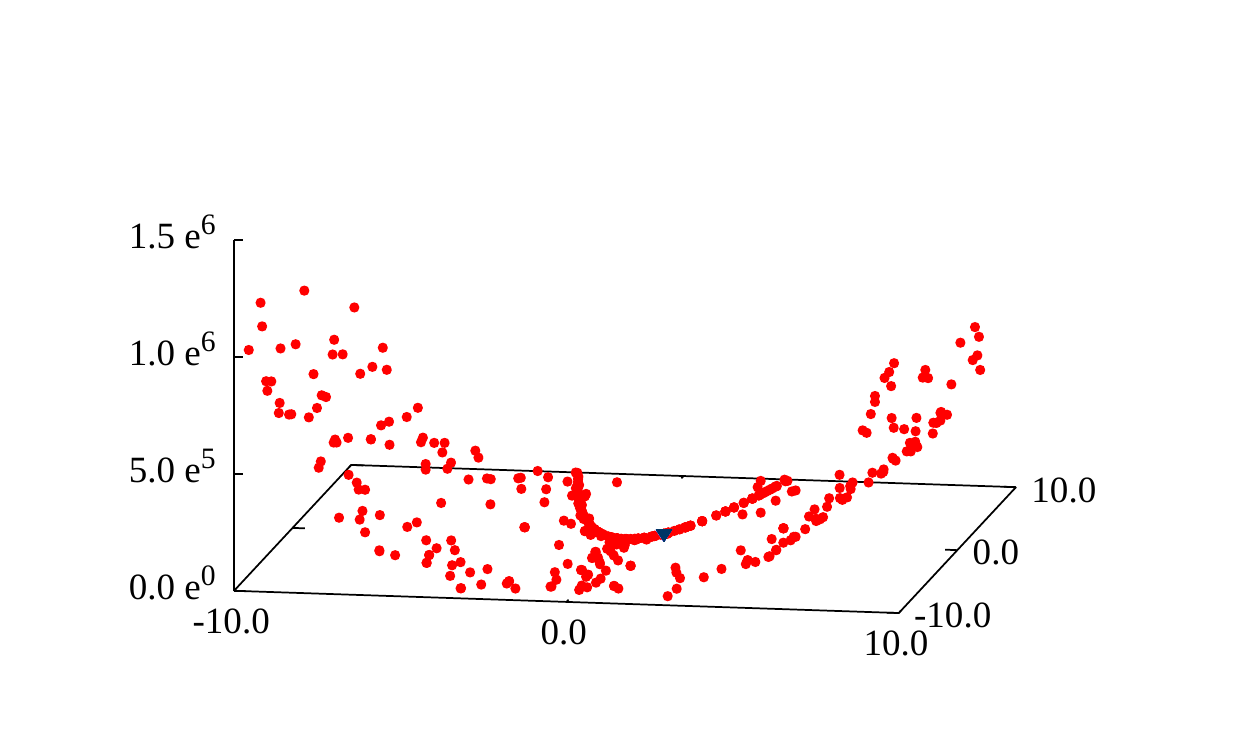} \\
    (p) & (q) & (r) & (s) & (t)\\
  \end{tabular}
  \caption{Simple examples that apply {\xsnos} to the Ackley, Bent
    Cigar, Griewank, Rastrigin, and Rosenbrock functions to explain
    the deformation of space net during the convergence process.
    (a)--(e) The scattering situation of elastic points on the space
    net after 400 evaluations.  (f)--(j) The scattering situation of
    elastic points on the space net and current solutions of $s$ and
    $x$ after 800 evaluations.  (k)--(o) The scattering situation of
    elastic points on the space net and current solutions of $s$ and
    $x$ after 4,000 evaluations.  (p)--(t) The scattering situation of
    elastic points on the space net and current solutions of $s$ and
    $x$ after 20,000 evaluations.  Note that the elastic points, the
    current solution $s$, and the current solution $x$ are marked as
    the red, dark blue, and light blue points in (a)--(t).}
    \label{fig:overall-3}     
  \end{figure*}
\subsection{Statistical  Analysis}

To better understand the performance of the proposed algorithm, we
also conducted a statistical analysis for the {\xsnos} and other
search algorithms compared in this study.  \xfig{fig:cec2122_Wilcoxon}
shows the statistical analysis results of comparing {\xsnos} and all
the other search algorithms for CEC2021 and CEC2022 in terms of the
Wilcoxon test. The terms ``Better,'' ``No Difference,'' and ``Worse''
represent that the end result of {\xsnos} is significantly better
than, not significantly better than, and worse than the compared
search algorithm, respectively. The value of each search algorithm
stands for the number of test functions in such a situation. For
example, the results of NL-SHADE-RSP in CEC2021 show that {\xsnos} can
find better results than NL-SHADE-RSP for 20 and 15 test functions
with 10 dimensions and 20 dimensions. Of course, {\xsnos} cannot find
better results than NL-SHADE-RSP in the other 22 and 25 test functions
while there are 8 and 10 test functions the results of {\xsnos} are
worse than NL-SHADE-RSP.  The results also show that {\xsnos} can find
better results than all the other search algorithms compared in this
paper because the value of ``Better'' (the number of test functions
that {\xsnos} outperforms the other search algorithm) is always larger
than the value of ``Worse'' (the number of test functions that the
other search algorithms outperform {\xsnos}) for all the other search
algorithms in \xfig{fig:cec2122_Wilcoxon}.  Based on these comparisons
and analyses, {\xsnos} can be regarded as an effective search
algorithm that can provide a better way for search in general.

\begin{figure*}[tbh]
  \centering
  \newcommand{\xca}[1]{\multicolumn{5}{l}{#1}}
  \footnotesize
  \setlength\tabcolsep{0pt}
  \begin{tabular}{ccccc}
    \xca{~~~~\includegraphics[width=0.5\textwidth]{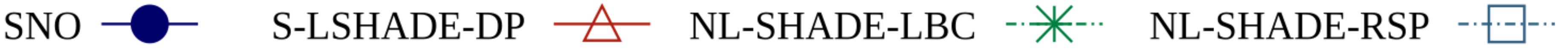}}\\
    \textbf{The convergence analysis.}\\
    \includegraphics[width=0.2\textwidth]{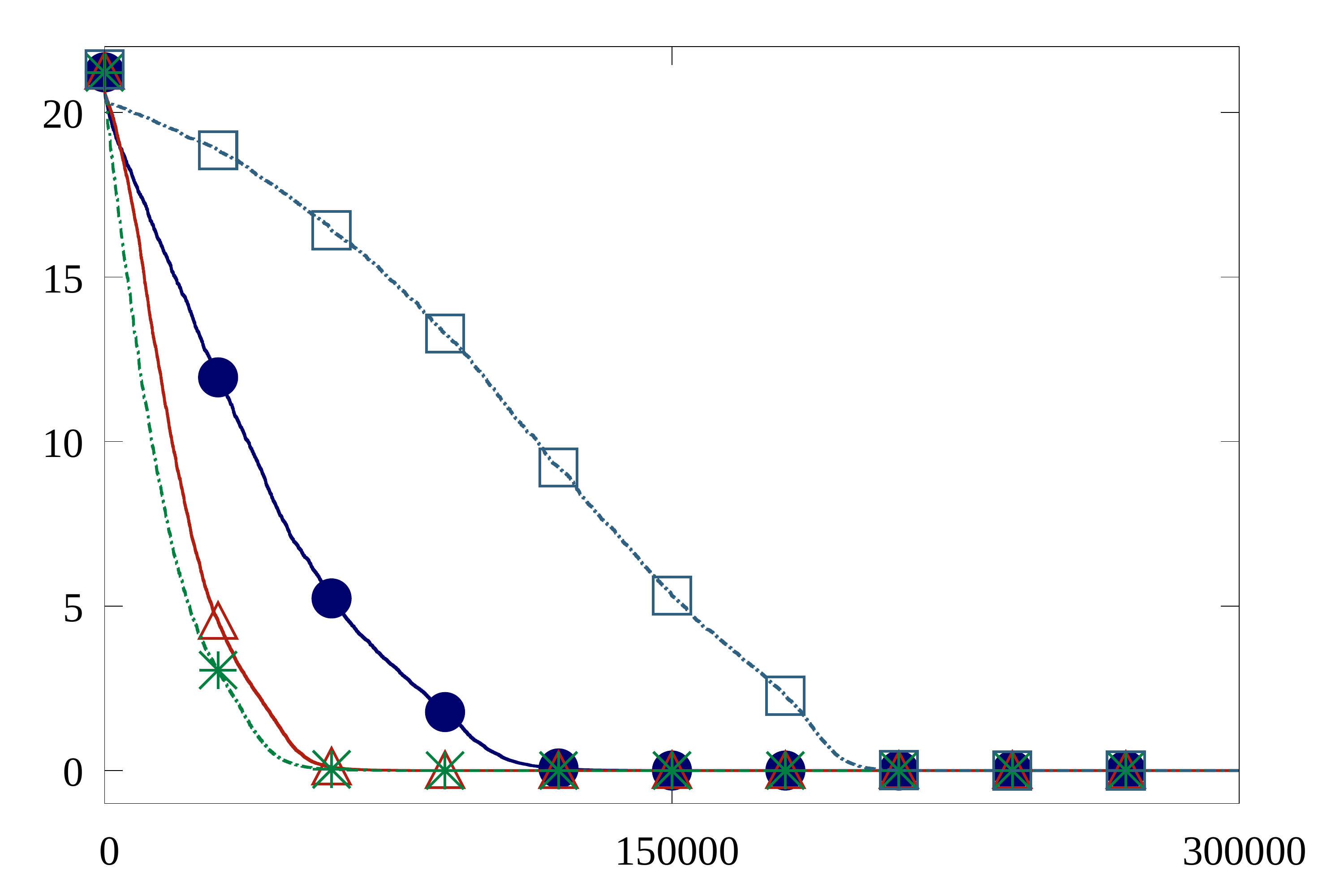} &
    \includegraphics[width=0.2\textwidth]{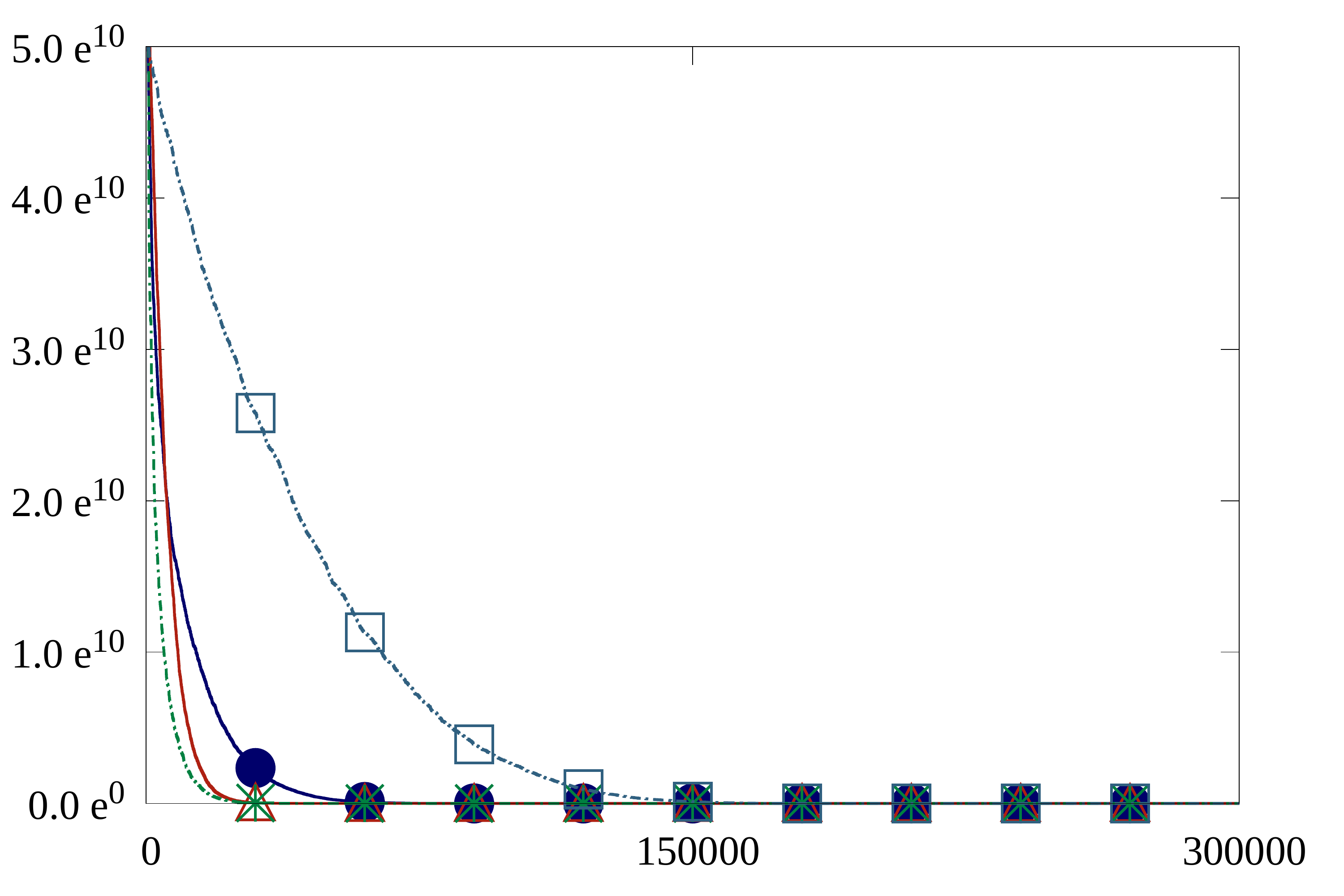} &
    \includegraphics[width=0.2\textwidth]{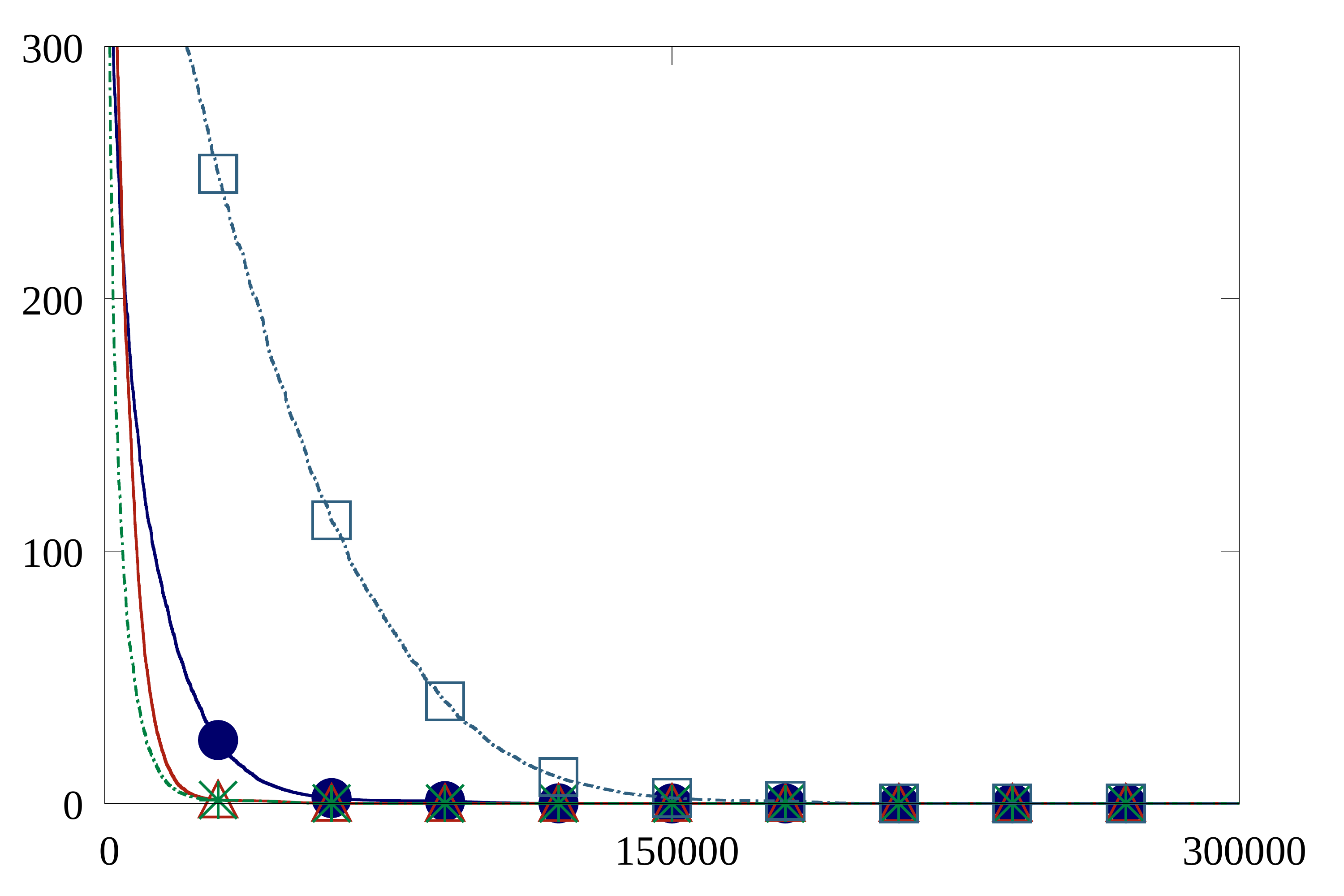} &
    \includegraphics[width=0.2\textwidth]{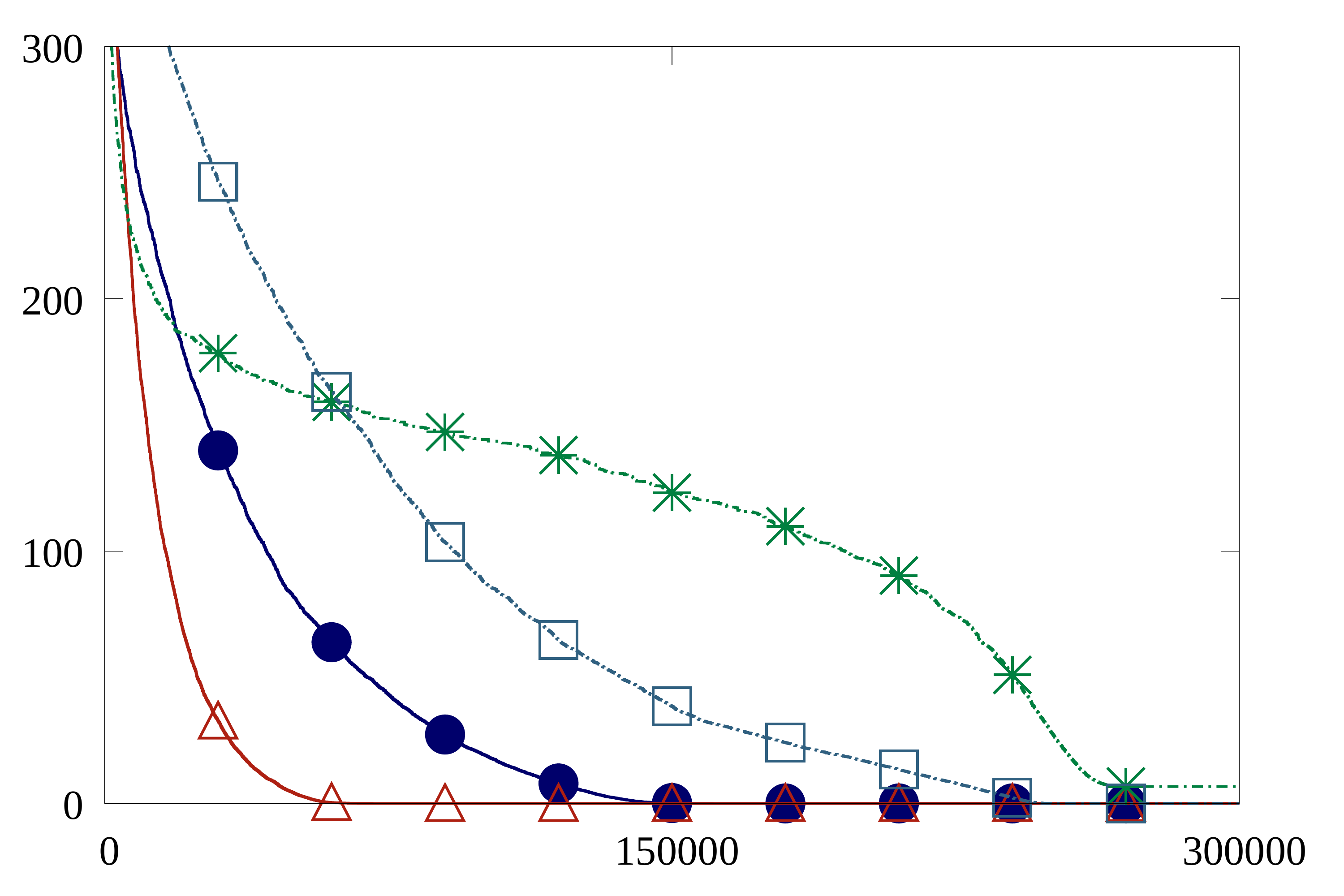} &
    \includegraphics[width=0.2\textwidth]{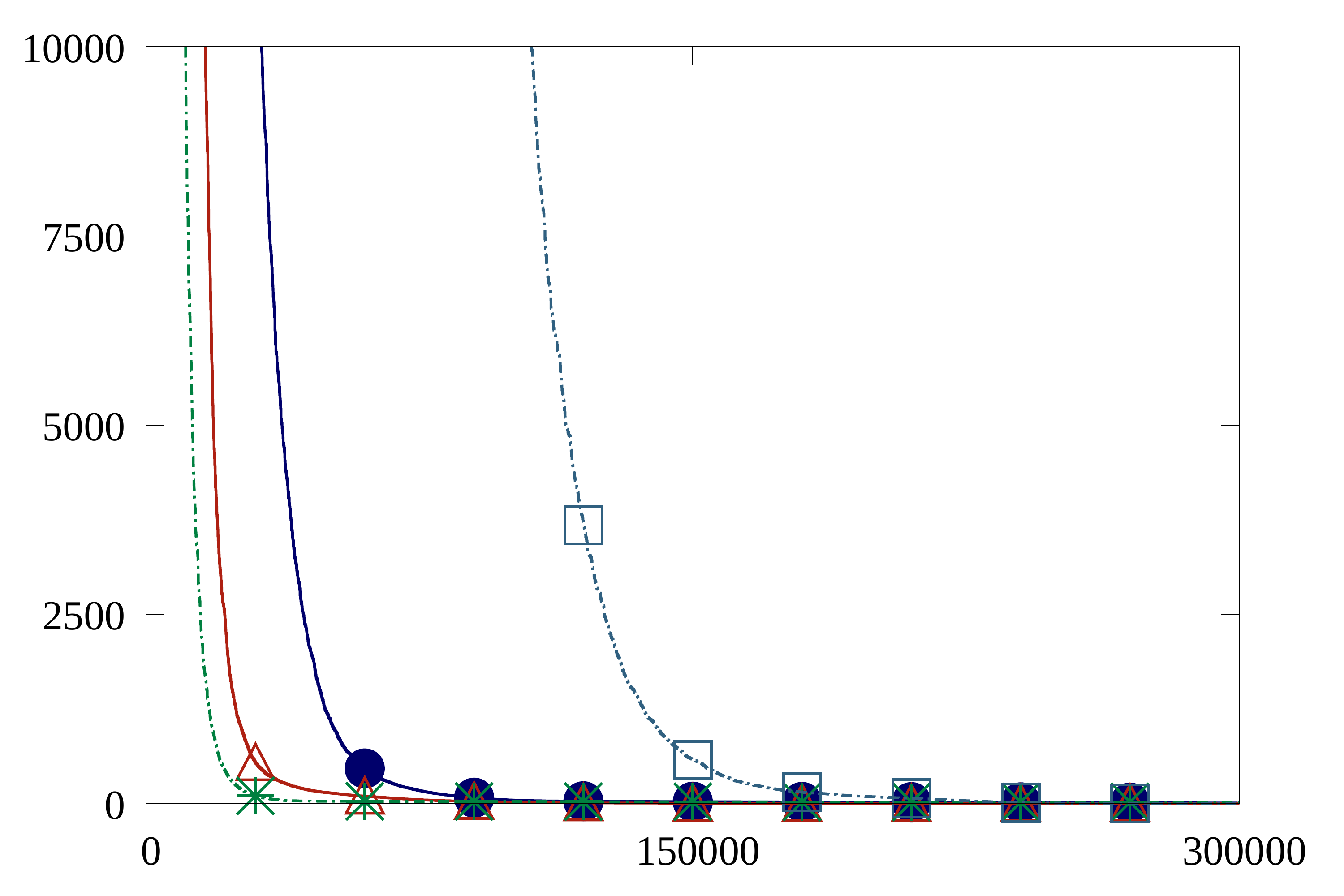} \\
    (a) Ackley & (b) Bent Cigar & (c) Griewank & (d) Rastrigin & (e) Rosenbrock\\\\
    \textbf{The exploration analysis}\\
    \includegraphics[width=0.2\textwidth]{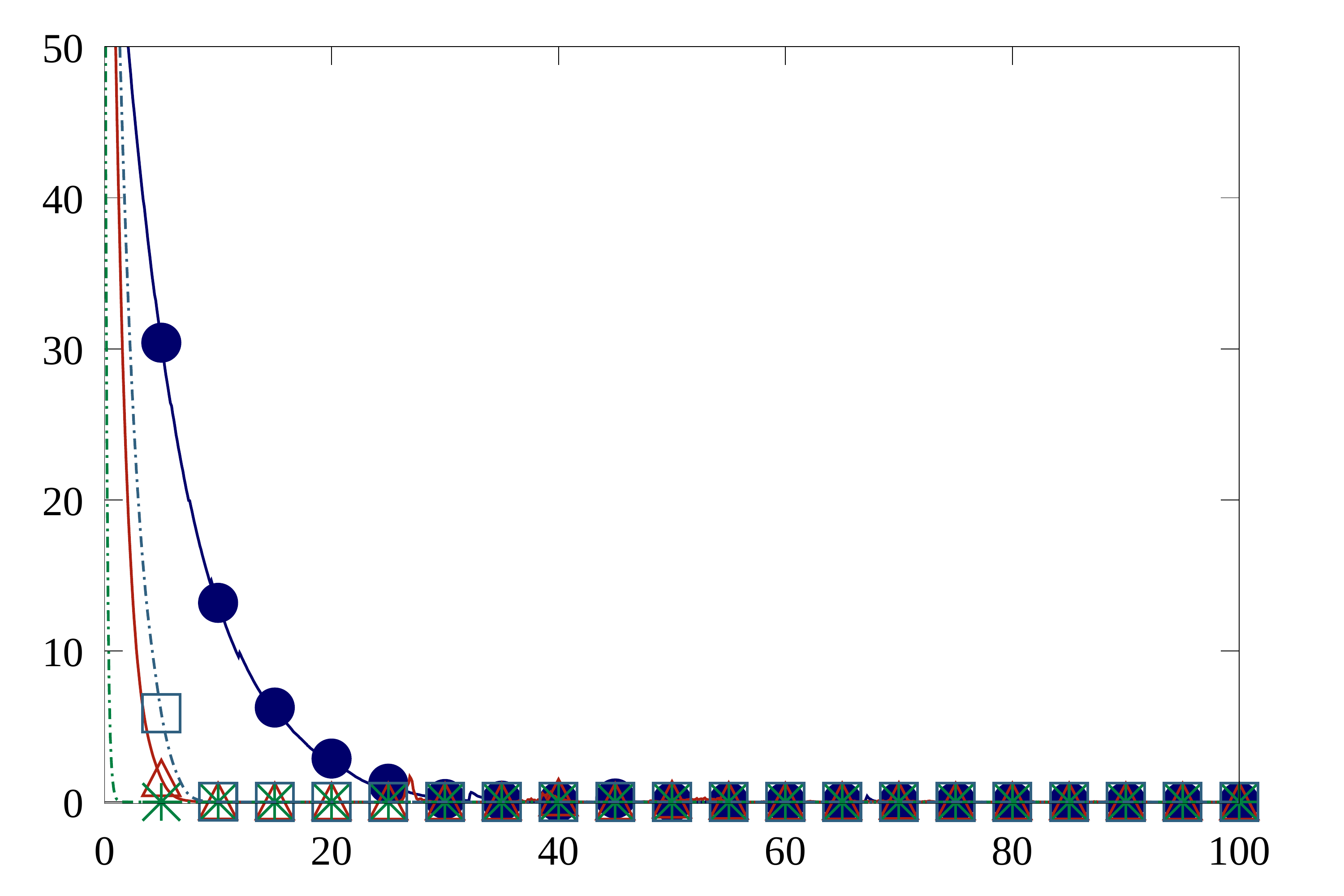} &
    \includegraphics[width=0.2\textwidth]{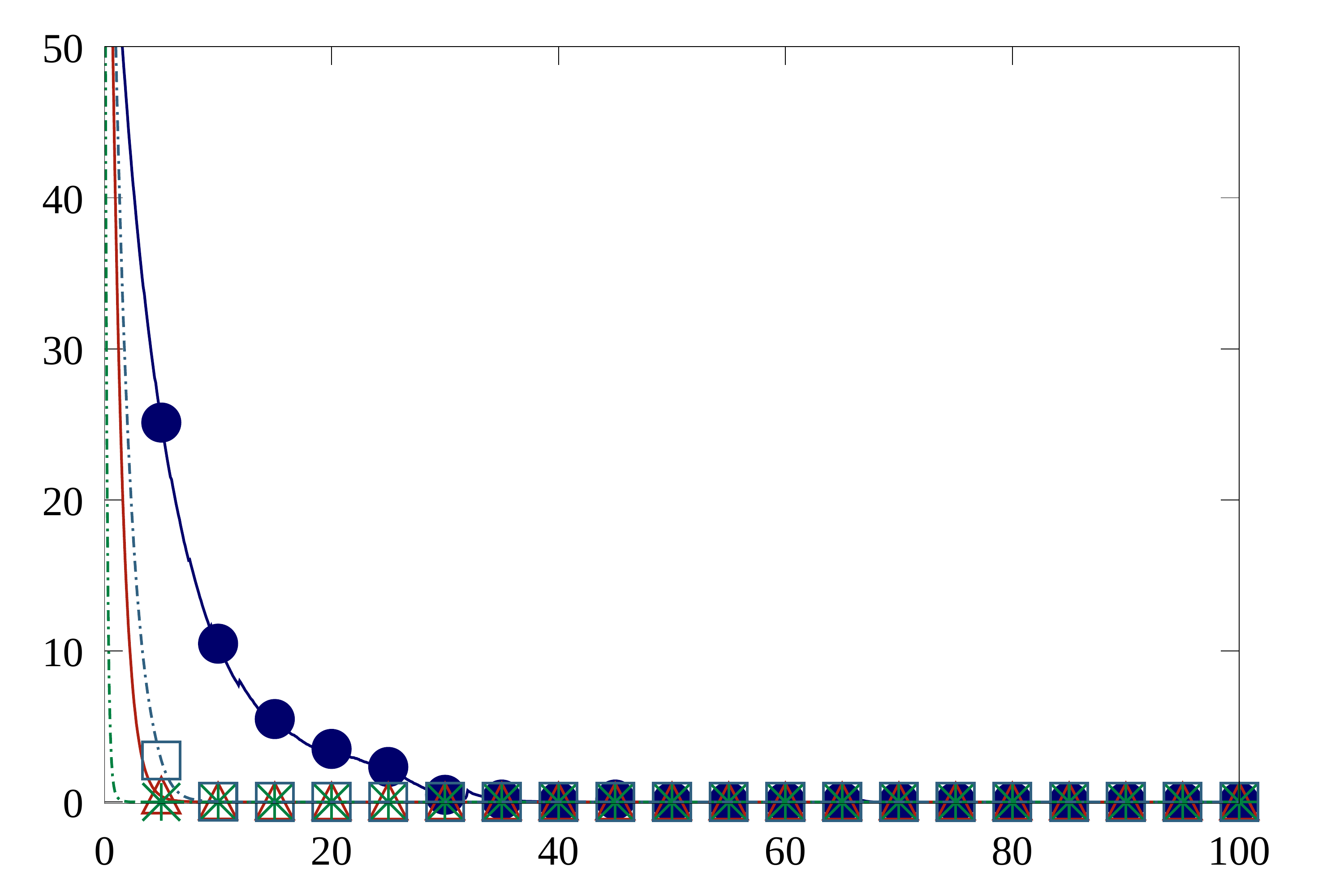} &
    \includegraphics[width=0.2\textwidth]{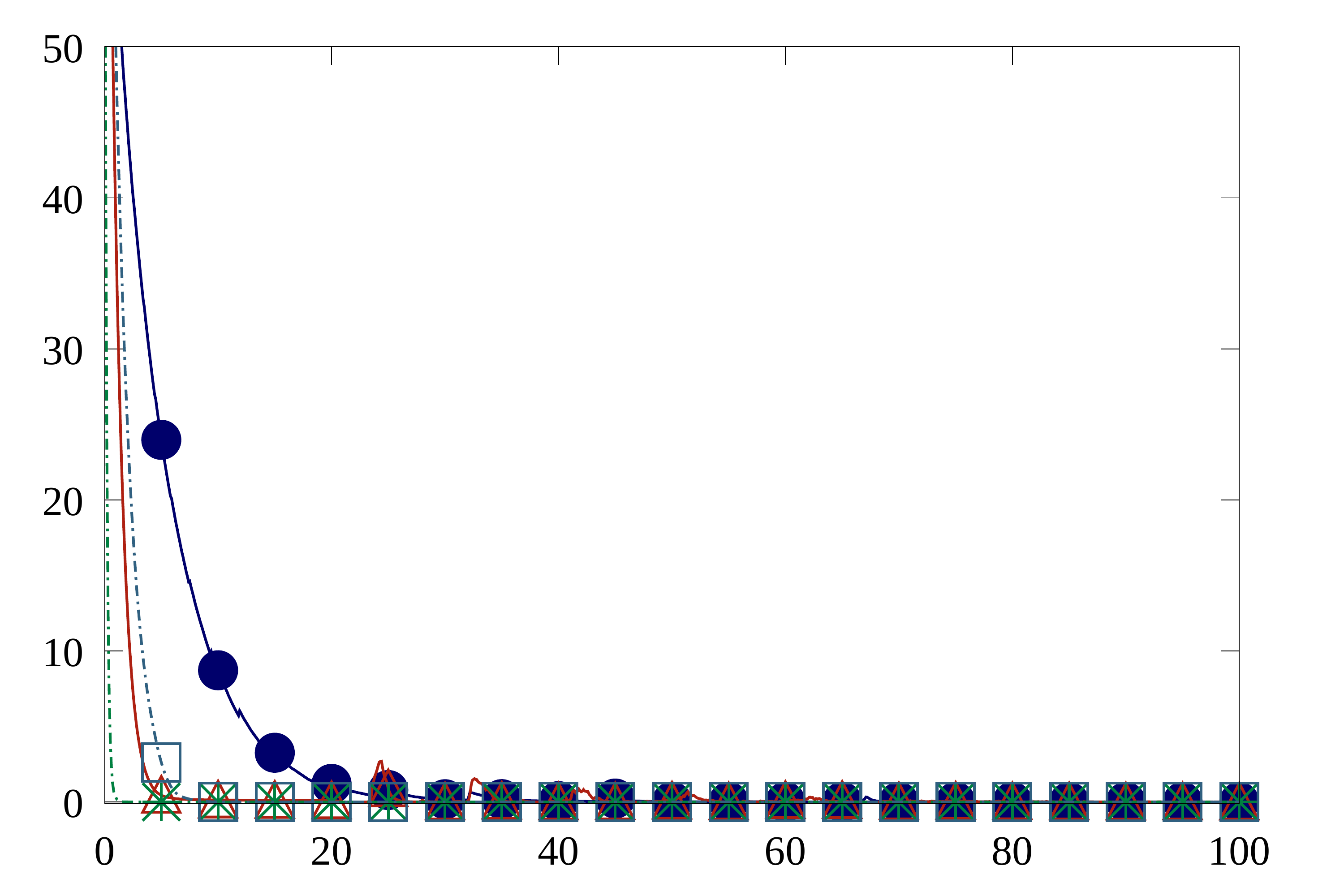} &
    \includegraphics[width=0.2\textwidth]{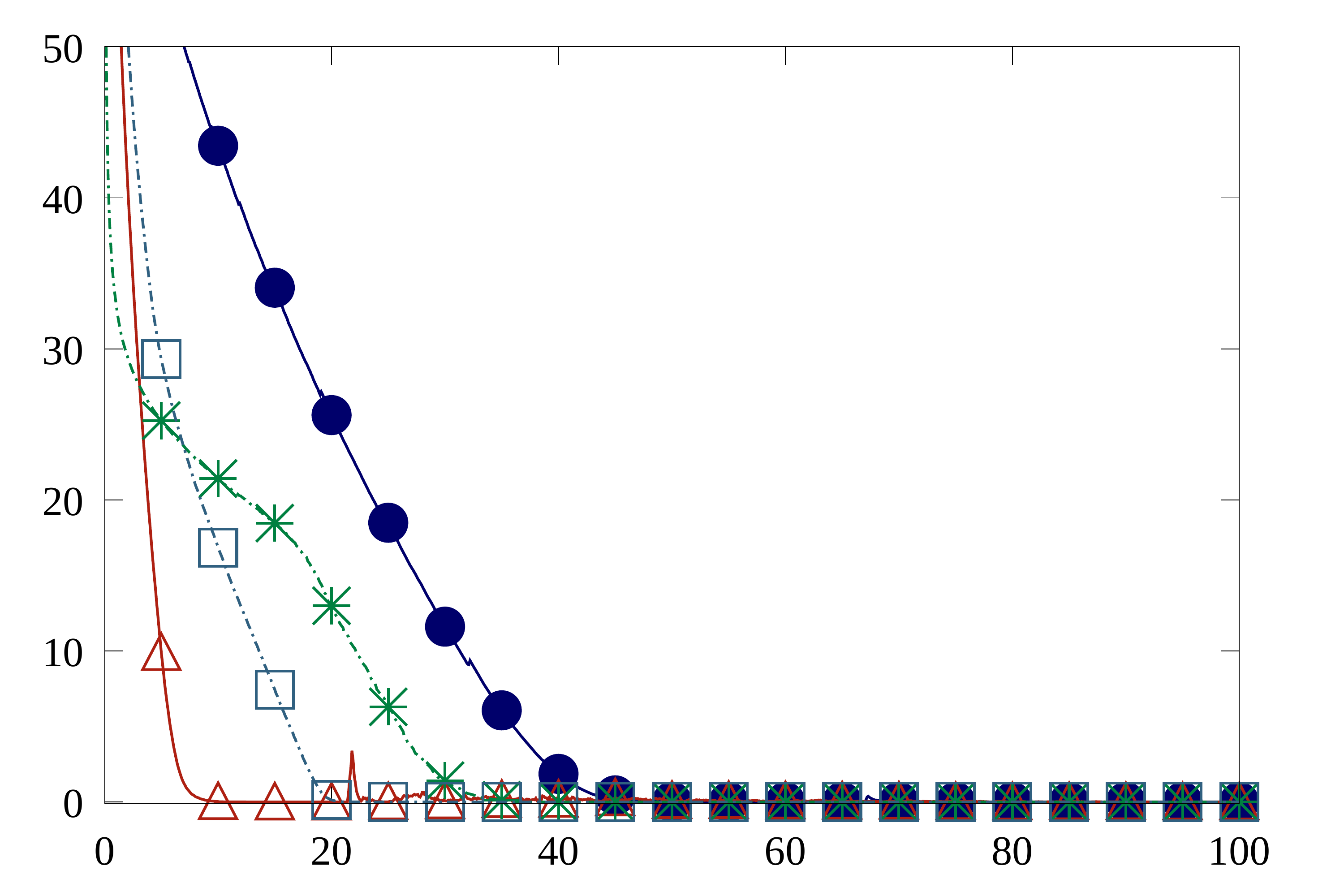} &
    \includegraphics[width=0.2\textwidth]{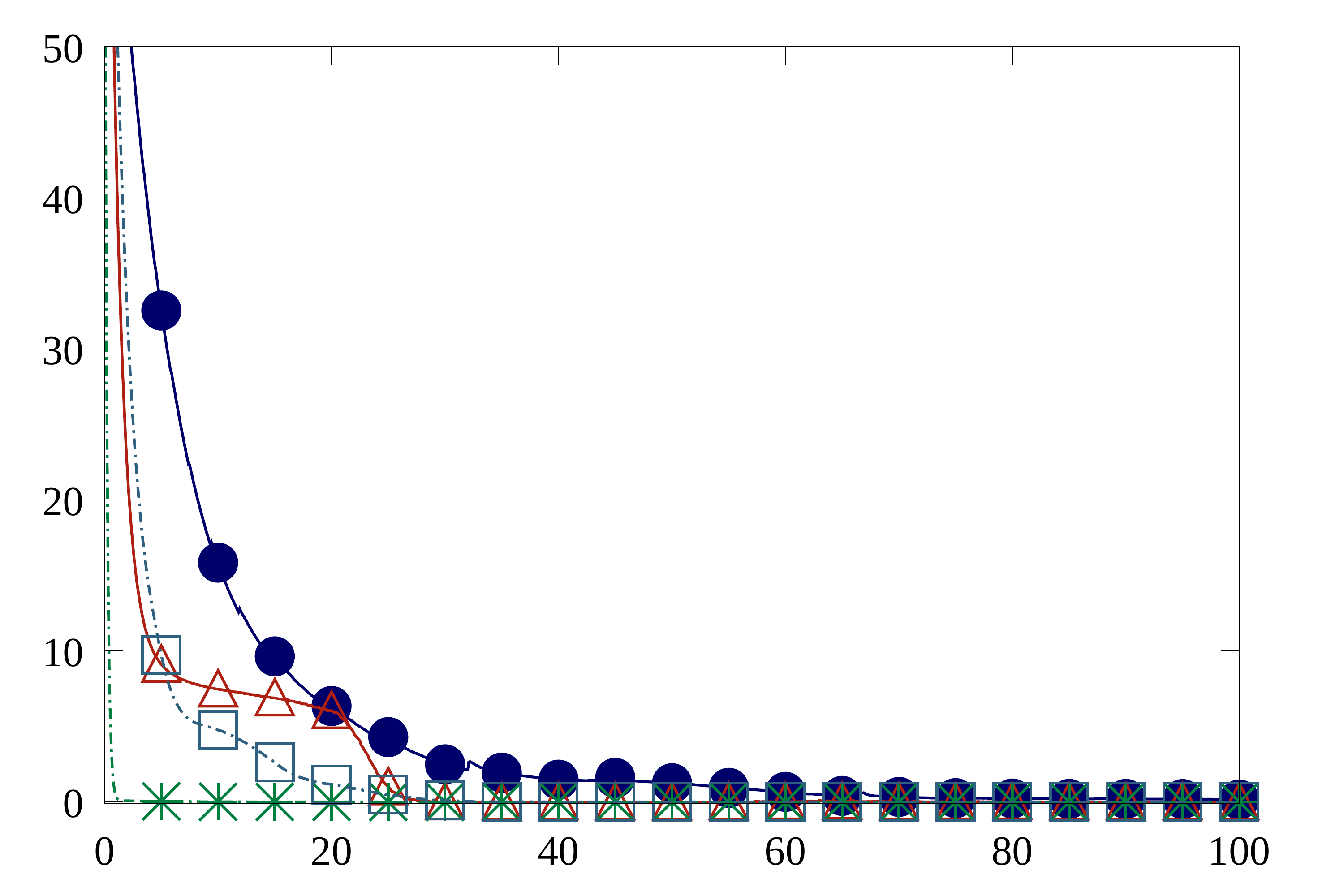} \\
    (f) Ackley & (g) Bent Cigar & (h) Griewank & (i) Rastrigin & (j) Rosenbrock\\\\
    \textbf{The exploitation analysis}\\
    \includegraphics[width=0.2\textwidth]{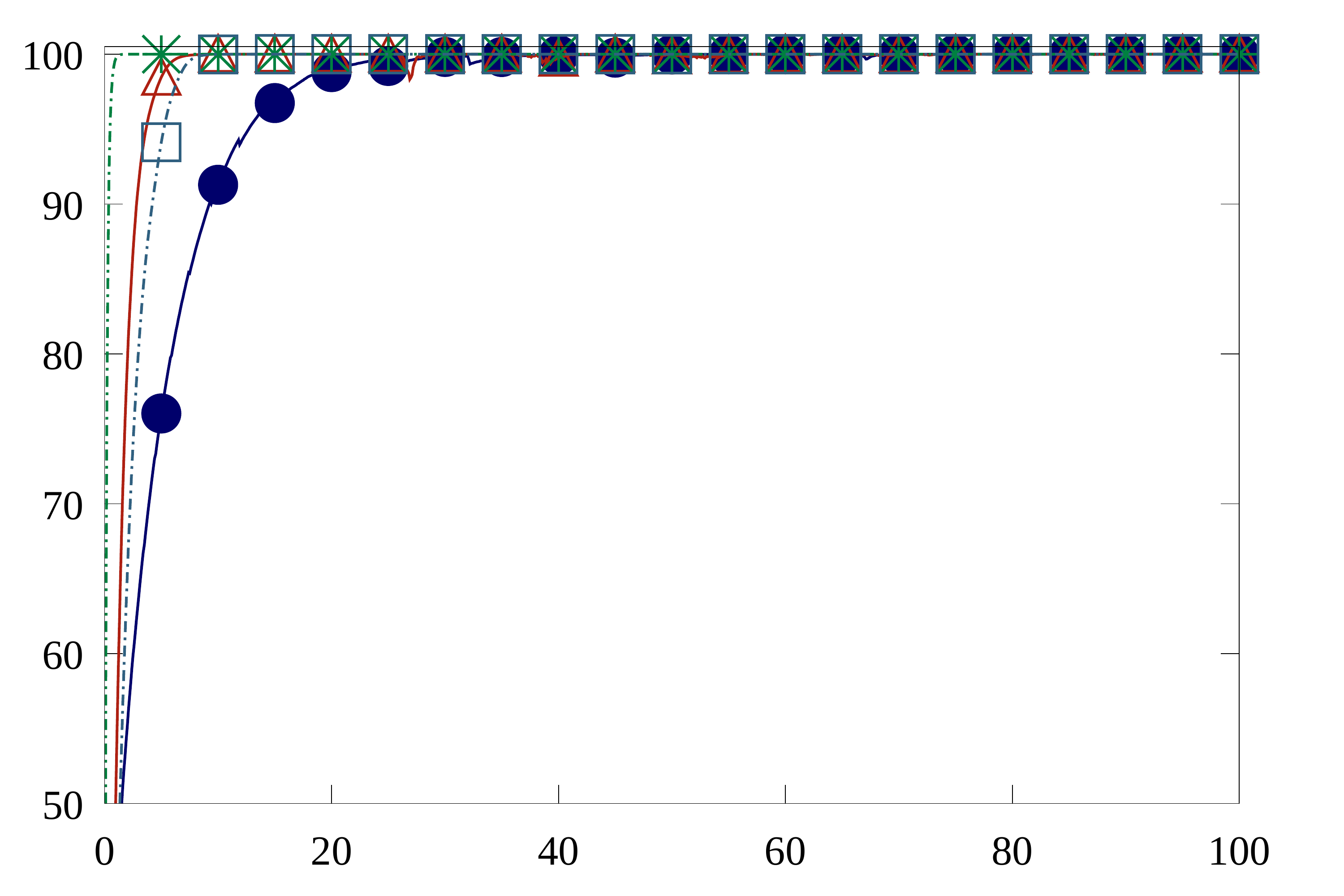} &
    \includegraphics[width=0.2\textwidth]{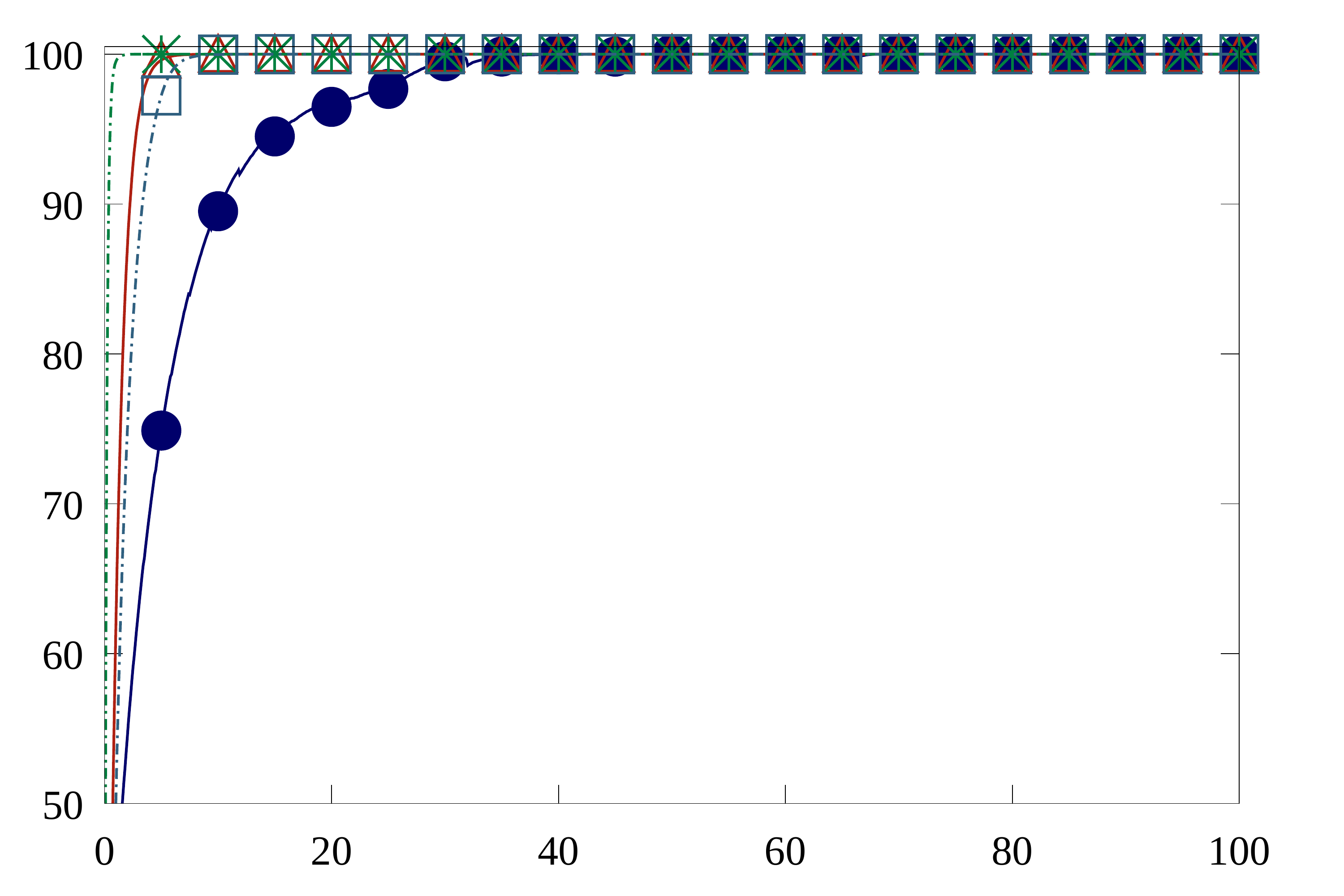} &
    \includegraphics[width=0.2\textwidth]{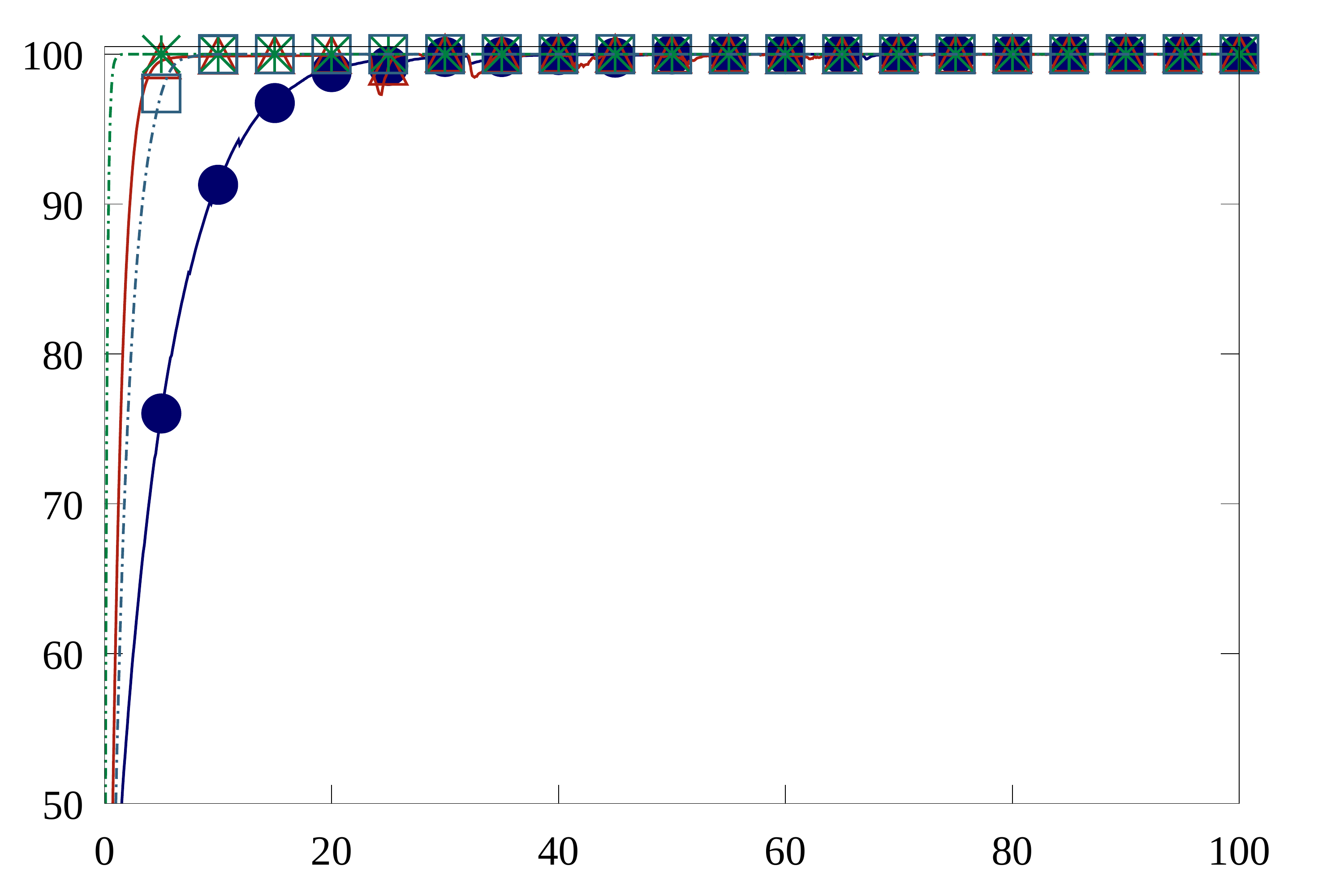} &
    \includegraphics[width=0.2\textwidth]{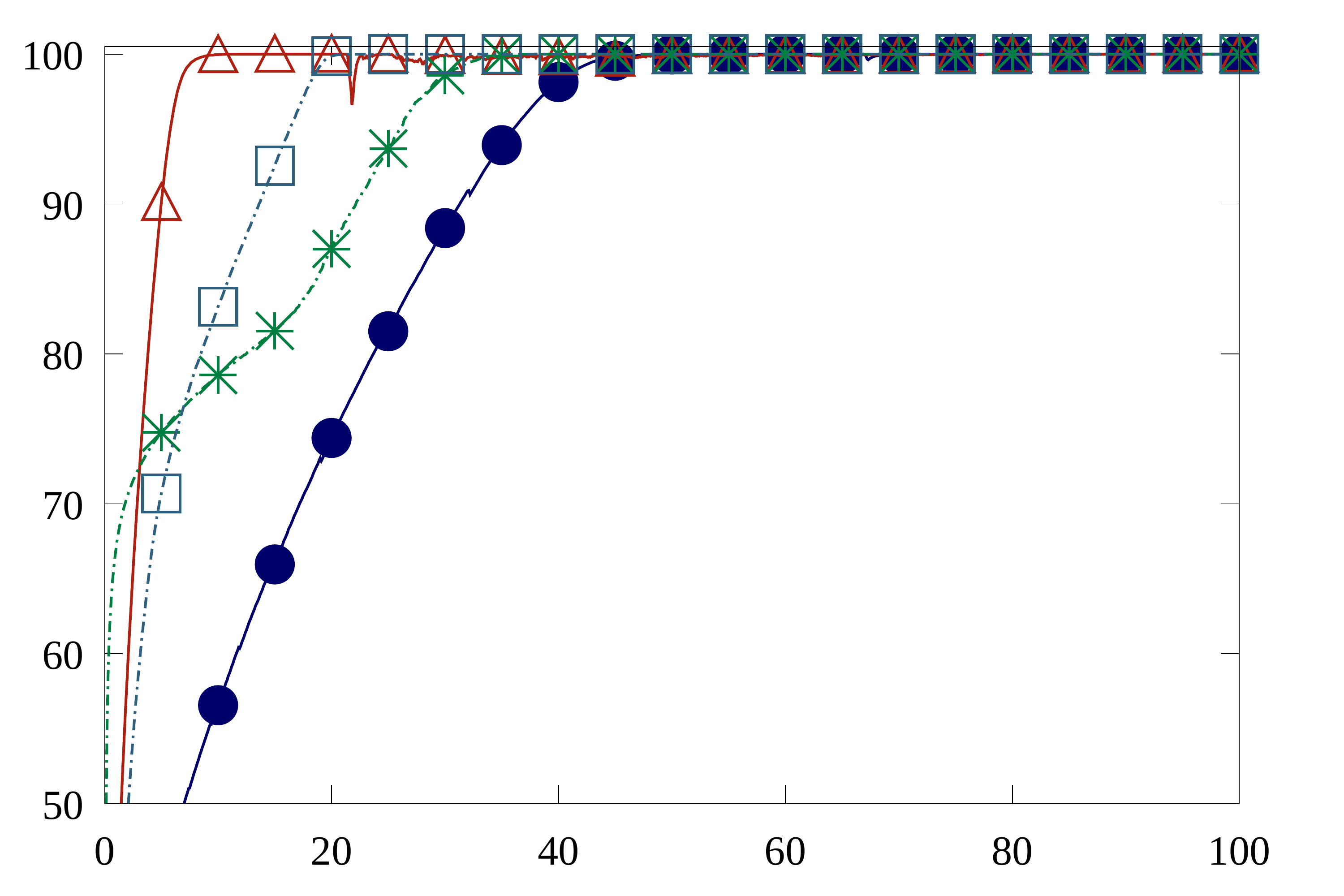} &
    \includegraphics[width=0.2\textwidth]{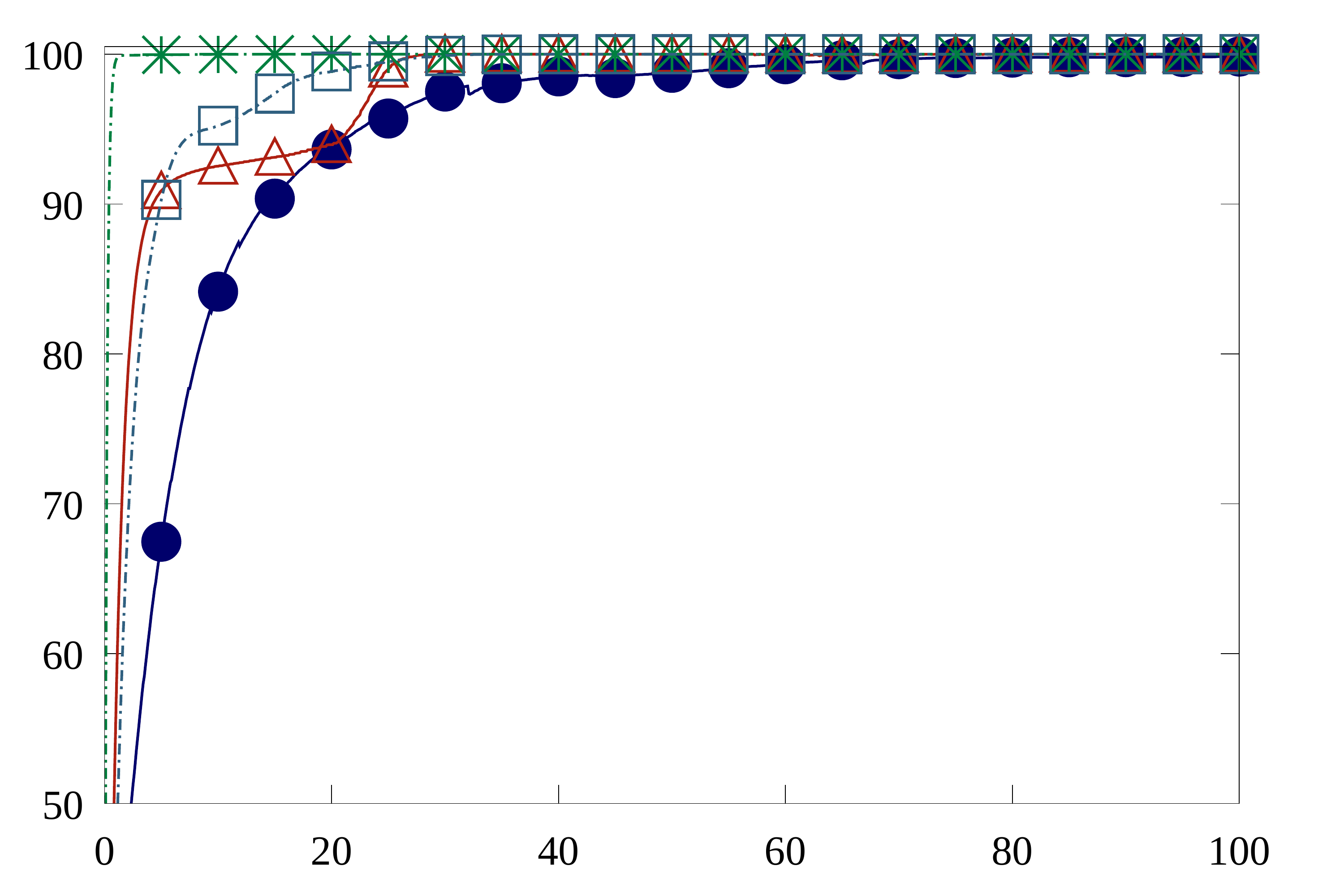} \\
    (k) Ackley & (l) Bent Cigar & (m) Griewank & (n) Rastrigin & (o) Rosenbrock\\\\
    \textbf{The diversity analysis}\\
    \includegraphics[width=0.2\textwidth]{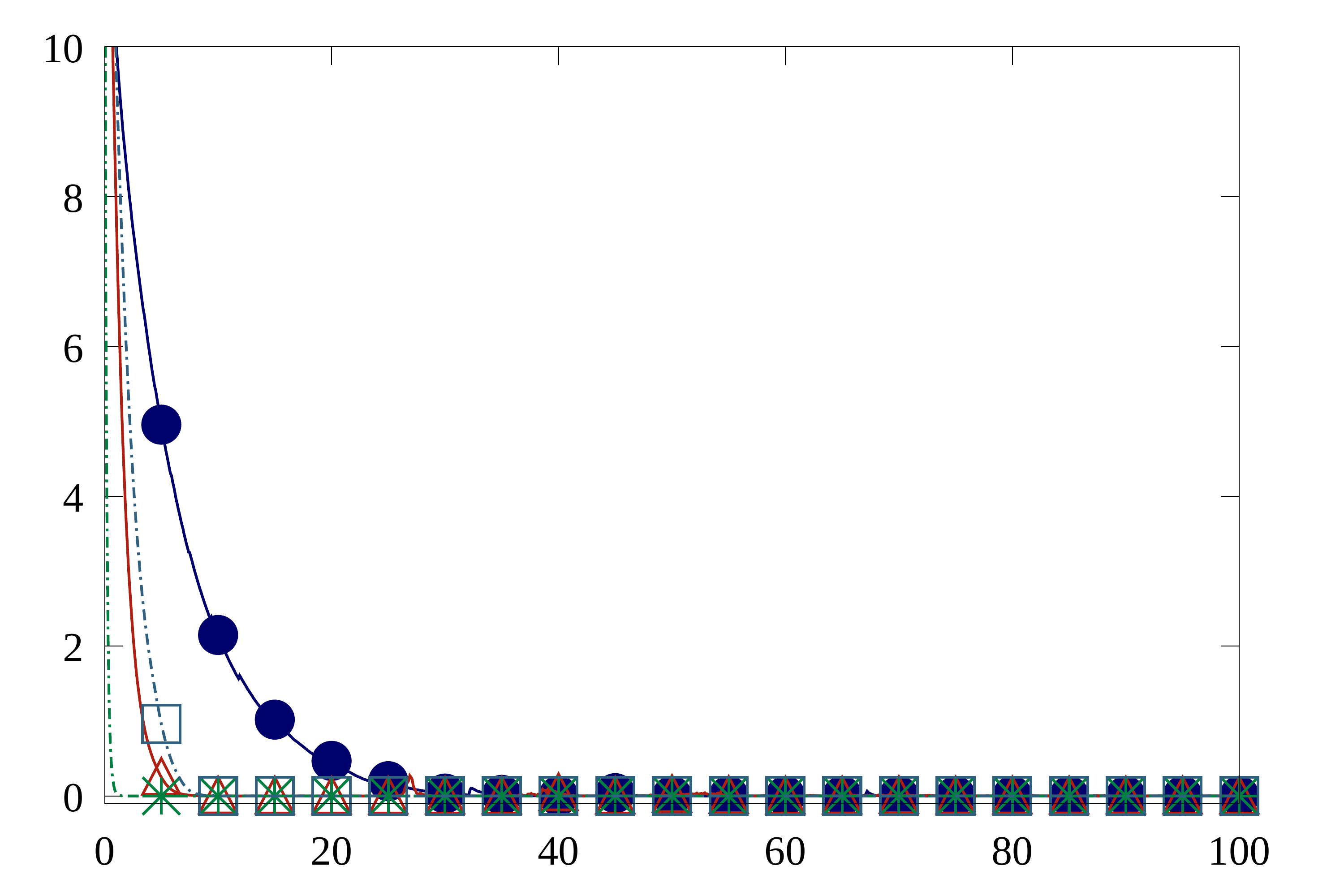} &
    \includegraphics[width=0.2\textwidth]{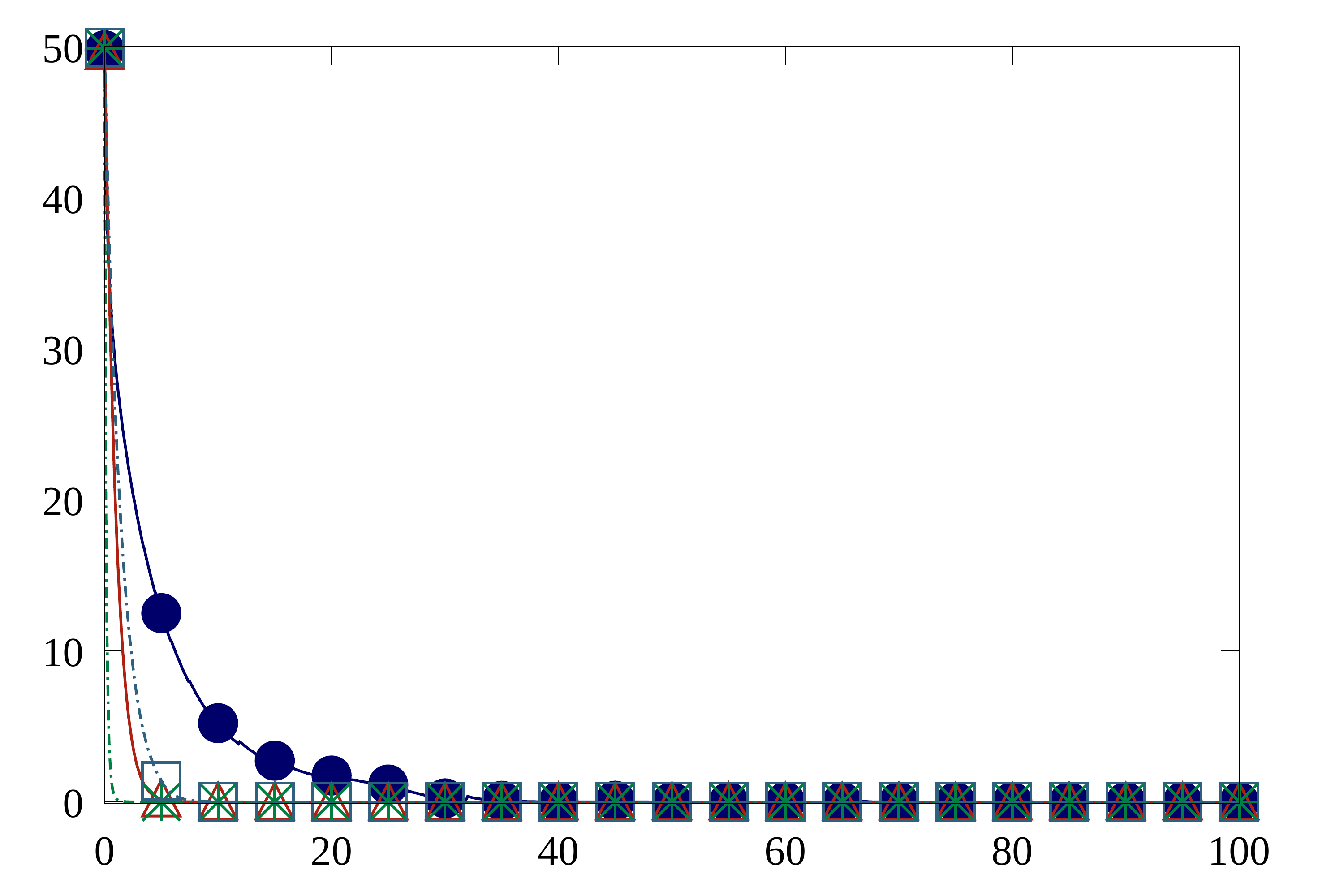} &
    \includegraphics[width=0.2\textwidth]{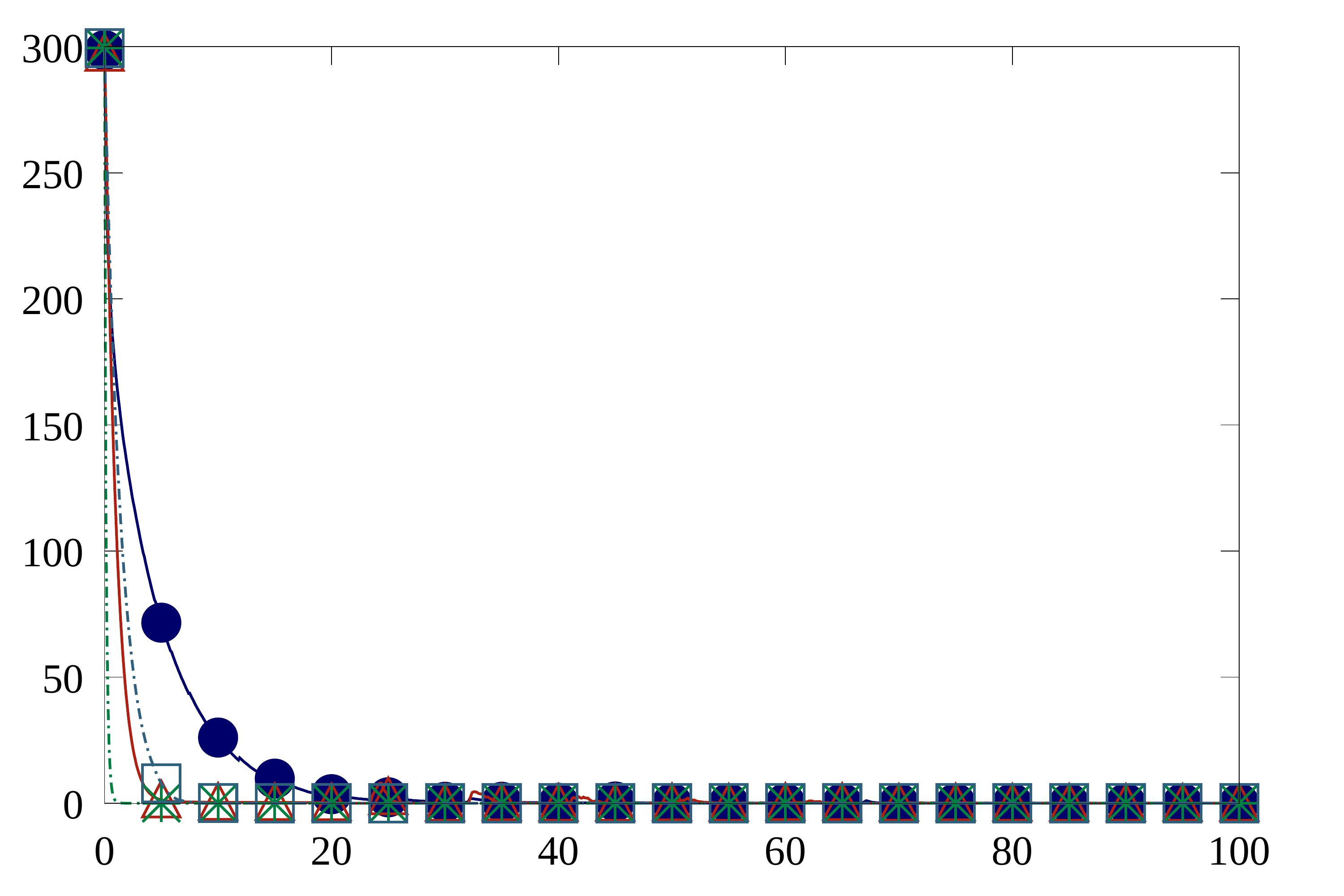} &
    \includegraphics[width=0.2\textwidth]{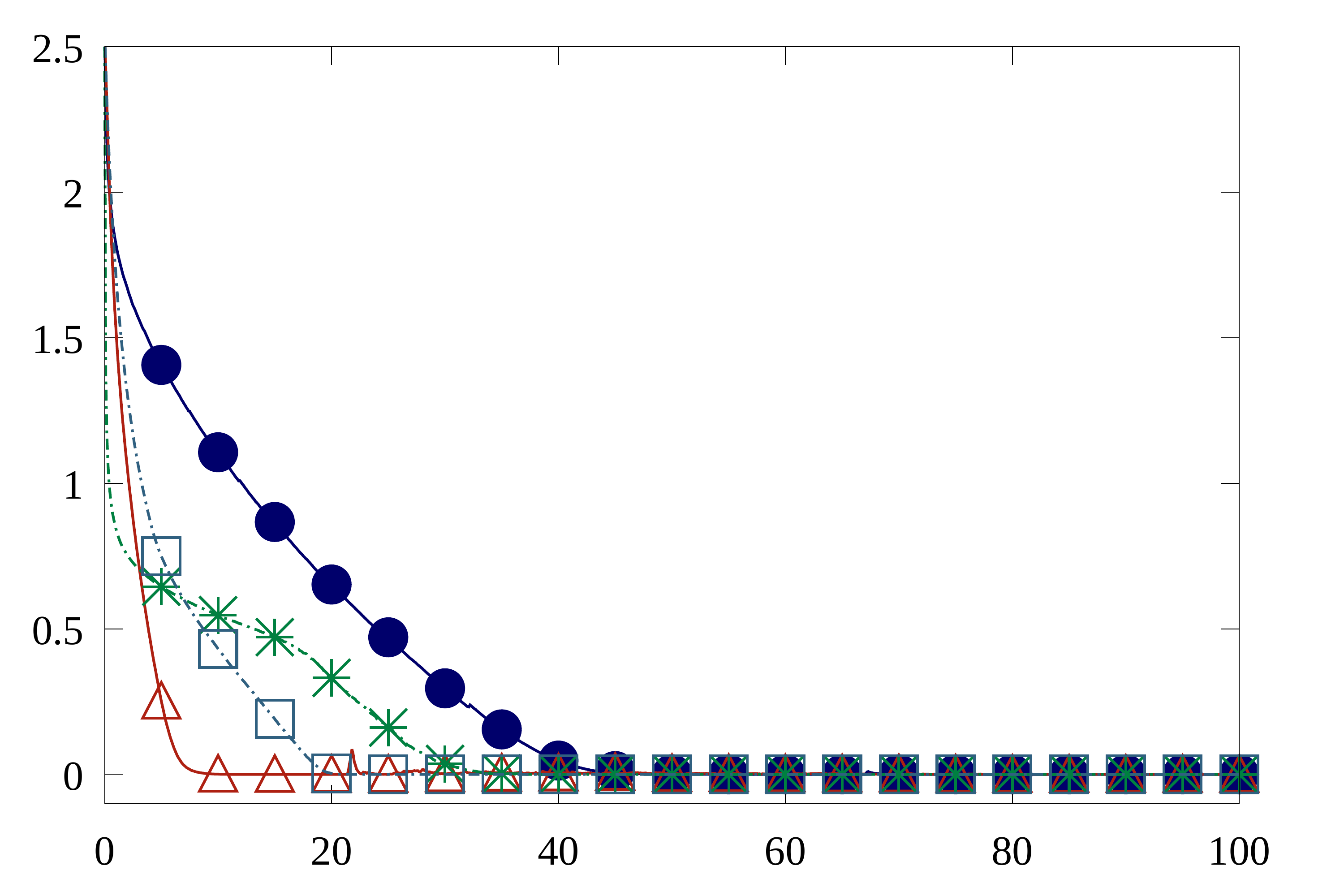} &
    \includegraphics[width=0.2\textwidth]{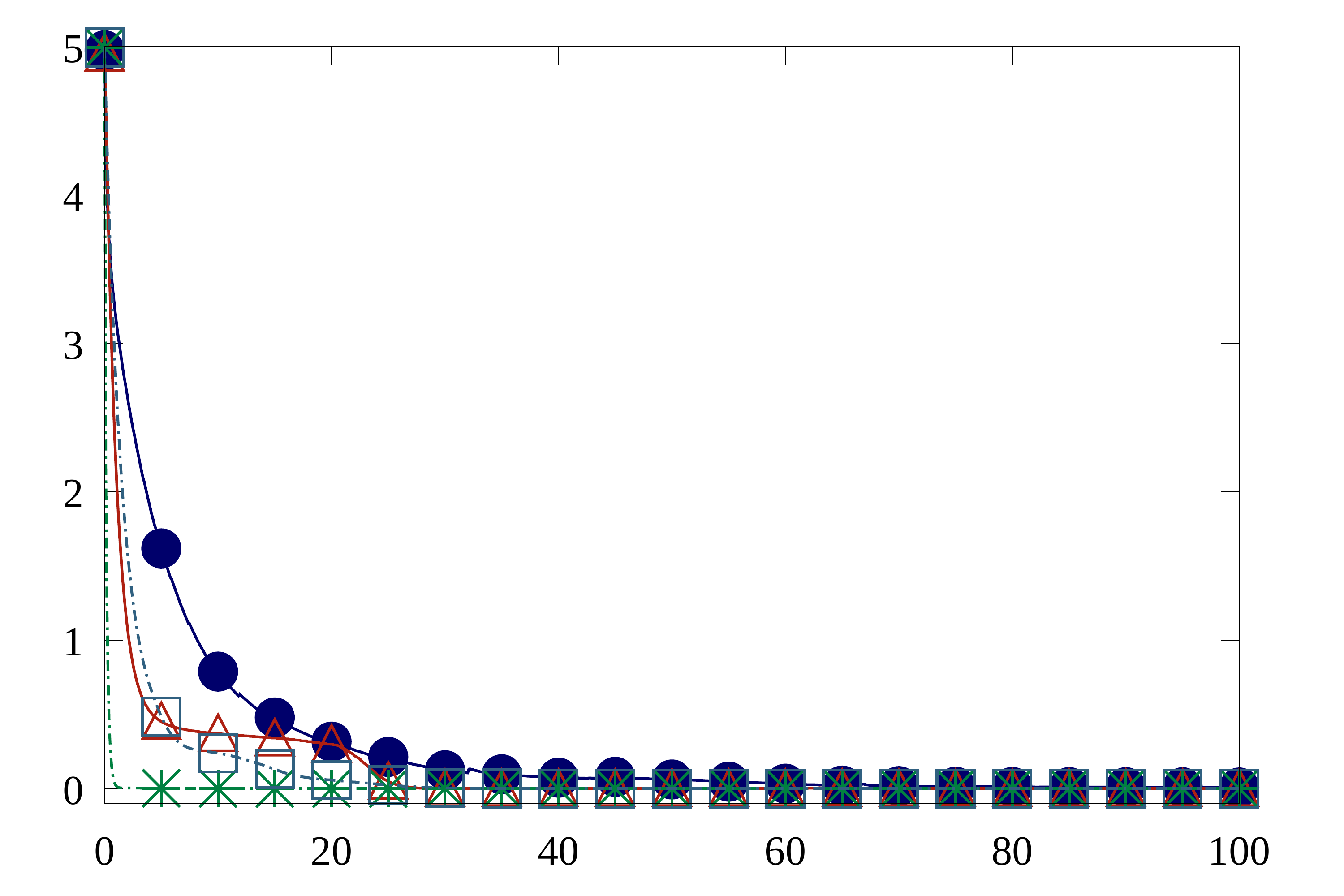} \\
    (p) Ackley & (q) Bent Cigar & (r) Griewank & (s) Rastrigin & (t) Rosenbrock\\
  \end{tabular}
  \caption{The analyses of convergence, exploration, exploitation, and
    diversity of {\xsnos}, S-LSHADE-DP, NL-SHADE-LBC, and
    NL-SHADE-RSP. (a)--(e) The analyses of convergence. (f)--(j) The
    analyses of exploration. (k)--(o) The analyses of
    exploitation. (p)--(t) The analyses of  diversity.}
    \label{convergence}              
\end{figure*}

\subsection{The Deformation of Space Net and Change of Searches}
To further understand the deformation of space net (changes of elastic
points) of {\xsnos} in different situations, it is used to solve the
Ackley, Bent Cigar, Griewank, Rastrigin, and Rosenbrock functions in
different periods during the convergence process, as shown in
\xfig{fig:overall-3}. The landscapes of these functions are first
shown on the top of the figure. In addition to the positions of the
elastic points, the positions of the current solutions $s$ and $x$ are
also given to show the relationship between them.
All the results show that the space net will fit to the landscape
gradually as the number of evaluations increases. The results also
imply that the space net can provide the information of landscape to
help {\xsnos} accurately determine the regions for later
searches. This is why {\xsno} is able to find better results than all
the other {\xmetaha}s for solving different SOPs in most cases.

\subsection{Convergence Analysis}

In this section, we conducted a set of convergence analyses to
understand the convergence circumstance of the proposed algorithm that
consist of analyses of convergence (quality), exploration (global
search), exploitation (local search), and diversity (ratio of global
to local search) \cite{MORALESCASTANEDA-2020}.  As shown in
\xfig{convergence} (a)--(e), the proposed algorithm is compared with
other three state-of-the-art search algorithms (i.e., S-LSHADE-DP,
NL-SHADE-LBC, and NL-SHADE-RSP) for five 30-$d$ basic
functions---Ackley, Ben Cigar, Griewank, Rastrigin, and
Rosenbrock---with 300,000 evaluations in terms of convergence.  Note
that the $x$-axis and $y$-axis here represent the number of
evaluations and the objective value.  It can be easily seen that the
convergence speeds of S-LSHADE-DP and NL-SHADE-LBC are faster than
{\xsnos} and NL-SHADE-RSP for most test functions, i.e., Ackley, Ben
Cigar, Griewank, and Rosenbrock.

\xfig{convergence} (f)--(o) show that the comparisons between these
four search algorithms in terms of the exploration (\xfig{convergence}
(f)--(j)) and exploitation (\xfig{convergence} (k)--(o)) analyses can
be regarded as the analyses of the ``global search ability'' and
``local search ability'' of each search algorithm during the
convergence process, respectively.  Because of the population size
settings of S-LSHADE-DP \cite{Van-2022}, NL-SHADE-LBC
\cite{Stanovov-2022}, and NL-SHADE-RSP \cite{StanovovAS21}, the
numbers of iterations these four search algorithms take are different
to make them the same scale.  The percentage of the convergence
process is used to label the ticks of the $x$-axes of subfigures of
\xfig{convergence} (f)--(o). Moreover, the $y$-axes of these
subfigures stand for the percentage of searches focusing on
exploration or exploitation.  As shown in \xfig{convergence} (f)--(j),
the percentage of exploration searches (global searches) of {\xsnos}
are much higher than the other search algorithms during the period of
the first 20\% to 40\% of the convergence process although exploration
searches for all of them are decreased as the number of iterations
increases.  This means that {\xsnos} will prolong the global searches
during the entire convergence process compared with all the other
search algorithms.  In \xfig{convergence} (k)--(o), we can first
observe that the percentage of exploitation searches (local searches)
of all search algorithms will increase as the number of iterations
increases. The results also show that {\xsnos} will prolong the timing
to increase the exploitation searches compared with the other three
search algorithms.  From the results of \xfig{convergence} (f)--(o),
it can be easily seen that the time and percentage of exploration of
{\xsnos} (global search) are much higher than the other search
algorithms while the timing to enter the exploitation state of
{\xsnos} (local search) is much late compared with the other search
algorithms.

\xfig{convergence} (p)--(t) further provides an integrated perspective
of the search abilities of different search algorithms during the
convergence process that considers the exploration vs.\ exploitation
(global search vs.\ local search) called diversity
\cite{MORALESCASTANEDA-2020}. The $x$-axes and $y$-axes of all
subfigures of \xfig{convergence} (p)--(t) represent the diversity and
the percentage of the convergence process.  The results show that the
search diversities of {\xsnos} are much higher than all the other
search algorithms for these five test functions.  According to our
observations, it is one of the most important reasons that {\xsnos} is
capable of finding better results than all the other search algorithms
compared in this study in most cases.

\section{Discussion}\label{sec3}

In this study, we present a novel {\xmetaha} to solve optimization
problems called {\xsno} that contains a new mechanism (space net) to
extract and accumulate the information of all searched solutions to
depict the landscape of the solution space on the fly during the
convergence process.  Different from most {\xmetaha}s that use only a
few searched solutions to guide the searches, the information of the
solution space constructed by space net will be used to enhance the
search accuracy of {\xsnos}.  Based on the design of the space net, we
then design the ways to generate new candidate solutions of {\xsnos}
to refer to not only the current solutions but also the elastic points
of the space net.  In such a design, current solutions can be regarded
as the current positions we are and elastic points can be regarded as
the map of the space we have.  That is why the metaheuristic algorithm
will now have the vision on the solution space as we pointed out in
the very beginning of this paper.  The simulation results show that
{\xsnos} outperform all the other state-of-the-art search algorithms
for complex SOPs compared in this study.  One of our future works will
be on applying the proposed algorithm to the other kind of
optimization problem, such as combinatorial optimization problem.
Another future work is to develop a more effective way to construct
the space net to depict the solution space to enhance the search
ability of {\xsnos}.

\section{Methods}\label{sec4}
\scriptsize
\subsection{Problem Definition}
The {\xsopa} is a well-known global optimization problem, which can be
found in many real-world scientific and engineering applications in
our daily life.  An {\xsops} can be formulated as:

\begin{xdefinition}
  \label{eq:opt1} 
  Given a function and a set of constraints, this problem aims to find
  the optimum objective value, subject to the constraints, out of all
  the feasible solutions of the function.
  \begin{displaymath}    
    \min\limits_{s\in\mathbf{R}^d}f(s), \xst\ \forall f_j(s)\xoptop b_j, j=1, 2, \dots, m,
  \end{displaymath}
  where
  \begin{itemize}
    \scriptsize
      \item $s=(s_1, s_2, \dots, s_d)$ is a feasible solution, $d$ is
    the number of dimensions, and $s_i$ is in the range $[L_i, U_i]$,
    where $L_i$ and $U_i$ are the lower and upper bounds of the $i$-th
    dimension;
    \item $f(s):\mathbf{R}^d \to \mathbf{R}$ is the objective function to be optimized, 
    \item $f_j(s):\mathbf{R}^d \to \mathbf{R}\xoptop b_j$ is the
  $j$-th constraint where $\xoptop$ is either $<$, $>$, $=$, $\leq$,
  or $\geq$, $m$ is the number of constraints, and
    \item $\mathbf{R}^d$ and $\mathbf{R}$ are the domain and codomain.
  \end{itemize}
\end{xdefinition}

As for a search algorithm, the goal of this problem is to find out a
``good'' solution that has the best objective value from the whole
solution space (i.e., the set of all feasible solutions).  With
\xdf{eq:opt1} at hand, we can now use it to describe an optimization
problems that we would like to test.  The Ackley function
\cite{Ackley-1987, Back-1996} can be used as an example to use this
definition to describe {\xsops}.
\begin{xdefinition}[The Ackley optimization problem]
  \small
  \label{eq:ackley}
  \begin{multline*}
  \min\limits_{s\in\mathbf{R}^n}f(s)=  -20 \exp\left(-0.2 \sqrt{\frac{1}{n} \sum_{i=1}^n s_i^2}\right)\\
  \hfill{}-\exp\left(\frac{1}{n} \sum_{i=1}^n \cos(2\pi s_i)\right) + 20 + e,\\
  \xst\ -30 \leq s_i \leq 30, i=\mathop{\rm 1, 2}, \dots, d.
\end{multline*}    
\end{xdefinition}
It can be easily seen from \xdf{eq:ackley} that a feasible solution of
the Ackley function is within the hypercube $s_i\in[\,-30, 30\,]$,
$\forall i=1, 2, \dots, d$.  The global optimum (minimum) of the
Ackley function is $f(s_b)= 0$ located at $s_b=(0,0,\dots,0)$.  As
shown in \xfig{fig:overall-2}(a), an Ackley function (i.e., $d=2$) has
many local optima that make it difficult for a search algorithm to
find the global optimum from the starting point(s) randomly chosen
from the entire solution space.  That is why it was widely used as the
touchstone to understand the performance of a search algorithm.

The experimental settings of the maximum number of function
evaluations (MaxFES) for the test functions (CEC2021 and CEC2022) are
with different number of dimensions.  In this study, MaxFES is set
equal to 200,000 for the number of dimensions $d=10$ and equal to
10,00,000 for the number of dimensions $d=20$.  Note that a search
algorithm will be terminated when MaxFES is reached or when the error
value is smaller than $10^{–8}$ for each test in the search range
$[\,-100, 100\,]^d$.

\subsection{Details of the Proposed Algorithm}
\label{ssec:algorithm}
\scriptsize
To simplify the discussion of the proposed algorithm, the following
notation will be used throughout this section.

\begingroup
\newcommand{\entrylabel}[1]{#1} \newenvironment{ventry}%
{\begin{list}{}%
    {%
        \renewcommand{\makelabel}{\entrylabel}%
        \settowidth{\labelwidth}{$t$, $t_{\max}$}%
        \setlength{\leftmargin}{1.2\labelwidth}%
        \setlength{\parsep}{0pt}%
        \setlength{\itemsep}{\jot}%
    }} 
{\end{list}}
\begin{ventry} 

    \item [{$s$}] a set of explorers, i.e.,
  $s=\{s_1, s_2, \dots, s_{n_s}\}$, where $s_i$ is the $i$-th
  explorer, and $n_s$ is the population size of $s$. Moreover,
  $s_{i,j}$ is the $j$-th dimension of the $i$-th explorer, and $u$ is
  a set of temporary explorers generated from $s$.

    \item [{$x$}] a set of miners, i.e.,
  $x=\{x_1, x_2, \dots, x_{n_x}\}$, where $x_i$ is the $i$-th miner,
  and $n_x$ is the population size of $x$. Moreover, $x_{i,j}$ is the
  $j$-th dimension of the $i$-th miner, and $v$ is a set of temporary
  miners generated from $v$.
  
    \item [$p$] the space net, i.e., $p=\{p_1, p_2, \dots, p_{n_p}\}$,
  where $p_{i}$ is the $i$-th elastic point of the space net, and $n_p$ is the
  population size of $p$ (i.e., the number of elastic points in the space
  net). Moreover, $p_{i,j}$ denotes the $j$-th dimension of the $i$-th elastic
  point, $q$ a set of elastic points generated
  from $p$, and $n_a$ how many elastic points will
  be attracted and moved toward to a new candidate solution of an
  explorer or a miner.

    \item [$r$] a set of regions used to allocate the elastic points
  in the space net, i.e., $r=\{r_1, r_2, \dots, r_h\}$, where $r_i$ is
  the $i$-th region; $h$ the number of regions, which is set equal to
  $h=(\sqrt{n_p}-1)^2$; $r_{i,k}$ the $k$-th elastic point at the
  $i$-the region; and $r_{i,b}$ the best elastic point at the $i$-th
  region.

    \item [{$e$}] a set of expected value
  $e=\{e_1, e_2, \dots, e_h\}$, each of which consists of
  the visited ratio, improvement value, and best value of the $i$-th region,
  respectively.

     \item [$\alpha, \beta$] sets of control parameter, i.e.,
   $\alpha=\{\alpha_1, \alpha_2, \dots, \alpha_h\}$ and
   $\beta=\{\beta_1, \beta_2, \dots, \beta_h\}$, where
   $\alpha_i$ and $\beta_i$ represent the crossover rate and
   scaling factor associated with the $i$-th region.

     \item [$c_s$, $c_x$] two control factors to control the search of RS
   and PS.
  
     \item [$\delta$] the ratio of the number of function evaluations
   performed so far $\text{FES}$ to the predefined
   maximum number of function evaluations $\text{FES}_{\max}$, i.e.,
   $\delta = \frac{\text{FES}}{\text{FES}_{\max}}$.

     \item [$\lambda()_{a}^{b}$] an adjustment function to get the
   adaptive value of the given parameter $v$ that will increase or
   decrease from its initial value $a$ to the end value $b$ via
   $a+\delta(b-a)$, i.e., $\lambda(\delta)_{1}^{2}$ stands for the
   adaptive value of this function will be computed as
   $1 + \delta(2 - 1)$ where $a=1$ and $b=2$.

     \item [$\phi$] a random number uniformly distributed in the range
   $[0,1]$.

    \item [$t$, $t_{\max}$] $t$ is the current iteration number and
  $t_{\max}$ is the maximum number of iterations.

\end{ventry}
\endgroup

As we mentioned before, the parameter settings of {\xsnos} are based
on the search results of TPE \cite{bergstra2011, akiba2019optuna}. The
detailed parameter settings of {\xsnos} are as follows:
$n_{\psi}^{\text{init}}$ represent the initial value of $n_{\psi}$
where ${\psi} \in \{s, x, \alpha, \beta\}$; and $n_{x}^{\text{end}}$
the end value (maximum value) of $n_{x}$.  Based on the search results
of TPE, we have the following settings: ${n_s}^\text{init} = 190$,
${n_x}^\text{init} = 0.1\times {n_s}^\text{init}$,
${n_x}^\text{end} = 0.2\times {n_s}^\text{init}$, and ${n_p} = 9^2$.
We also have the following settings: ${\rho}_{\max} = 0.7$, 
${\alpha}^{\text{init}} = 0.5$, and ${\beta}^{\text{init}} = 0.1$,
$c_s = 2.0$, and $c_x = 2.5$ for {\xsnos}.

\xalg{algo1} is an outline of the proposed algorithm. As shown in
\xalg{algo1}, the inputs are the objective function and the
parameter settings of the proposed algorithm while the output is the
best found solution $s_b$ (i.e., best so far solution).

\begin{algorithm}
\floatname{algorithm}{\footnotesize{Algorithm}}
\renewcommand{\algorithmicrequire}{\textbf{Input:}}
\renewcommand{\algorithmicensure}{\textbf{Output:}}  
\footnotesize
\caption{\footnotesize{Space Net Optimization}}
\label{algo1}
\begin{algorithmic}[1]
\Require The objective function $f()$  and parameter settings of {\xsnos}.
\Ensure Best found solution $s_b$.
\State 
       Initialization()
\While{$t < t_{\max}$}
\State $e$ = ExpectedValue($r$)
\State $u$ = RegionSearch($s$, $p$, $e$, $r$) 
\State $q$ = SpaceNetAdjustment($u$, $p$, $e$, $r$) 
\State $v$ = PointSearch($x$, $q$, $e$, $r$) 
\State $q$ = SpaceNetAdjustment($v$, $p$, $e$, $r$)
\State $u$, $v$ = PopulationAdjustment($u$, $v$, $q$, $e$, $r$) 
\State $s_b$ = UpdateBestSoFar($s$, $x$, $u$, $v$)
\State $s$ = $u$, $x$ = $v$, $p$ = $q$
\State $t=t+1$
\EndWhile
\State \Return The best found solution $s_b$
\end{algorithmic}
\end{algorithm}

The initialization operator will be invoked first to initialize
{\xsnos} that includes the populations $s$ and $x$, elastic points on
the space net $p$, regions $r$, expected values $e$, best found
solution $s_b$, and two control parameter sets $\alpha$ and $\beta$
based on the inputs and random procedure.  {\xsnos} will then perform
the operators ExpectedValue(), RegionSearch(), PointSearch(),
SpaceNetAdjustment(), PopulationAdjustment(), and BestSoFar(),
iteratively.  Among them, the BestSoFar() operator is the most simple
operator compared with the other operators, which will be used to
update the best found solution, that is, the best so far solution,
during the convergence process.  The other operators of {\xsnos} will
be described in detail below.

\subsubsection{Expected Value}
The expected value of {\xsnos} is inspired by \cite{Tsai_2016} where
it stated that the objective/fitness value of ``a solution'' can be
replaced by the expected value of ``a region'' of the solution space
for a novel {\xmetaha} (search economics; SE) to avoid the searches
trending to particular solutions.  The expected value of a region in
SE is composed of (1) the ratio of the number of time the region has
been searched to the number of times the region region has not been
searched, (2) the average objective value of new candidate solutions
generated in the region, and (3) the objective value of the
best-so-far solution in the region.  Different from the expected value
of most SEs, the expected value of {\xsnos} will use the improvement
value of elastic points belonging to a region to replace the average
objective value of the new candidate solutions generated in the $j$-th
region.  The expected value of a region, say, the $i$-th region $e_i$,
in {\xsnos} is thus defined as
\begin{equation}
  \begin{split}
  e_{i}  = & \mathcal{N}\left(\frac{I^b_i}{I^a_i}\right) + \mathcal{N}\left(\sum_{j = 1}^{4} \left(f(r_{i, j}^{t-1}) - f(r_{i, j}^{t})\right)\right) \\
           & \quad{}+\lambda(\delta)_{2}^{1} \cdot \left(1.0 - \mathcal{N}(f(r_{i, b}^t))\right),
  \end{split}          
    \label{eq:expected_value}
\end{equation}  
where the first term of \xeq{eq:expected_value} represents the impact
of the visited ratio, where $I^a_i$ and $I^b_i$ represent the number
of times the region has been visited and has not been visited.  The
second term represents the impact of the sum of the improvement value
of the elastic points that belong to the $i$-th region at the $t$-th
iteration.  The third term represents the impact of the objective
value of the best elastic point belonging to the $i$-th region at the
$t$-th iteration.  Besides, $\mathcal{N}()$ and $f()$ represent the
normalization and objective functions, respectively.

\subsubsection{Region Search}
The region search operator plays the role of generating new candidate
solutions $u$ from the population $s$ for ``exploration.''
In this operator, a number of regions (i.e.,
$\lambda(\delta)_{0.1}^{1.0} \times h $) will first be selected from
all regions $r$ based on their expected values in $e$.  The number of
selected top regions $\lambda(\delta)_{0.1}^{1.0} \times h $ will be
decreased gradually from $h$ at the very beginning down to
$0.1\times h$ during the convergence process.
These selected regions will then be used in the roulette wheel
selection to determine a region $r^{\text{sel}}_i$ based on the
expected values in $e$. A reference point $p_{i}^{\text{sel}}$ will
then be picked from the selected region $r^{\text{sel}}_i$ to generate
the new candidate solution of population $s$. The reference point
generation is defined as
\begin{equation}
    p_{i}^{\text{sel}} = \begin{cases}
        \text{Tourn}(r^{\text{sel}}_i), & \text{if}~\phi < \lambda(\delta)_{0.1}^{1.0}, \\
        r_{i,b}^{\text{sel}}, & \text{otherwise},
    \end{cases}
    \label{eq:net_point_region}
\end{equation}
where $\text{Tourn}(r^{\text{sel}}_i)$ represents the tournament
selection for the region $r^{\text{sel}}_i$ that will select an
elastic point while $r_{i,b}^{\text{sel}}$ represents the best elastic
point in the selected region $r^{\text{sel}}_i$.

Now, the region search operator will use the reference point to
generate a new candidate solution $u_i$ of population $s$.  A random
number $j_{\text{rand}}$ in the range $[1, d\,]$ will be first
generated for $u_i$; another random number $\phi_c$ in the range
$[0, 1]$ will then be generated for $u_{i,j}$ to determine whether to
generate a new candidate value or use the value in $s_{i,j}$ as the
value of the $j$-th dimension of the new candidate solution $u_{i,j}$,
which is similar to the crossover operator of DE \cite{Storn1997-DE}.
In case $\phi_c \geq \alpha_i$ and $j \neq j_{\text{rand}}$,
\begin{equation}
  u_{i,j} = s_{i,j},
  \label{eq:crossover}    
\end{equation}
where $\alpha_i$ is a crossover value associated with
$r^{\text{sel}}_i$. In case $\phi_c < \alpha_i$ or
$j = j_{\text{rand}}$, the $j$-th dimension of $u_i$ will be
calculated as follows:
\begin{equation}
  u_{i,j}
  = \begin{cases}
        p_{i,j}^{\text{sel}} + \beta_i \times \left(s_{r_1,j} - s_{r_2,j}\right),&\text{if}~\phi < {\delta^{c_s}}, \\
        s_{i,j} + \beta_i \times \left(p_{i,j}^{\text{sel}} - s_{r_1,j}\right),&\text{otherwise},
    \end{cases}
    \label{eq:strategy_region}
\end{equation}
where $\beta_i$ is the scaling factor associated with
$r^{\text{sel}}_i$.  At the end of the region search, if $f(u_{i})$ is
better than $f(s_{i})$, it will be used to replace the $s_i$ at next
iteration.

\subsubsection{Point Search}
The point search operator is used to generate the new candidate
solutions $v$ from population $s$ that is similar to the region
search, but for ``exploitation.'' Like the
\xeqs{eq:crossover}{eq:strategy_region} of region search, it will
first generate a random number $\phi_c$ for $v_i$ and a random number
$j_{\text{rand}}$ for $v_{i,j}$.  If $\phi_c \geq \alpha_i$ and
$j \neq j_{\text{rand}}$, $v_{i,j}$ will be set equal to $x_{i,j}$;
otherwise, the value of the $j$-th dimension of $x_i$ can be
calculated as follows:
\begin{equation}
  v_{\phi_x,j}
  = \begin{cases}
        p_{\phi_x,j}^{\text{top-}\rho} + \beta_i \times \left( x_{r_1,j} - x_{r_2,j} \right), & \text{if}~\phi < {\delta^{c_x}}, \\
        x_{\phi_x,j} + \beta_i \times \left( x_{r_1,j} - x_{r_2,j} \right), & \text{otherwise},
    \end{cases}
    \label{eq:strategy_point}
\end{equation}
where $\phi_x$ is a random number in the range $[1, n_x]$.  However,
there are some differences between region search and point search
operators.  The first difference is that the $i$-th update of region
search is for the $i$-th solution $s_i$ at each iteration, but it will
be a random one $x_{\phi_x}$ for point search.  The second difference
is that the $i$-th new candidate solution $p^{\text{sel}}_i$ in region
search is to select one elastic point by roulette wheel then
tournament selections from the elastic points belonging to the
specific region, but the $i$-th new candidate solution
$p_{\phi_x}^{\text{top}-\rho}$ in point search is to randomly select
one elastic point from top-$\rho$ elastic points of all elastic points
in the space net.  The value of $\rho$ is set equal to
$\lambda(\phi)_{0.1}^{\rho_{\max}}$ where $\rho_{\max}$ is the maximum
value of $\rho$.  Like the region search, if $f(v_{\phi_x})$ is better
than $f(x_{\phi_x})$, it will be used to replace $x_{\phi_x}$ at next
iteration.

\subsubsection{Space Net Adjustment}
The role of space net in {\xsnos} is to depict the landscape based on
the searched solutions of $s$ and $x$.

This operator will first update how many elastic points will be
attracted by a newly candidate solution $u_i$ from region search (or
new candidate solution $v_i$ generated from point search) that is
defined as follows:
\begin{equation}
    n_a = \lceil n_{a,\max} \times \delta \rceil,
    \label{eq:adjust_num_net}
\end{equation}
where $n_{a,\max}$ represents the predefined maximum number of elastic
points that will be attracted by a newly candidate solution.
In this study, ${n_{a, \max}}$ is set equal to 5.
To update the location of elastic points, in addition to the solutions
at $t$ iteration (i.e., $s$ and $x$), solutions at $t+1$ iteration are
also used to ``replace'' the elastic points of space net. Two
additional ways that can also be used to generate the reference point
to move the elastic points are defined as follows:
\begin{equation}
    p_{i}^{a} = \nu + \beta_i \cdot (\nu_{r_1} - \nu_{r_2}),
    \label{eq:strategy_1_net}
\end{equation}
\begin{equation}
    p_{i}^{b} = p_{i} + \beta_i \cdot (\nu - p_{i}) + \beta_i \cdot ( \nu_{r_1} - \nu_{r_2}),
    \label{eq:strategy_2_net}
\end{equation}
where $\nu \in \{u_{\phi_u}, v_{\phi_v}\}$ and
$\nu_{r_1}, \nu_{r_2} \in \{s_{\phi_s}, x_{\phi_x}\}$, with
$\phi_u, \phi_v, \phi_s, \phi_x$ representing a random number in the
ranges $[1, n_u]$, $[1, n_v]$, $[1, n_s]$, and $[1, n_x]$,
respectively.  Similar to the new candidate solution generation of
region search (e.g., \xeqs{eq:crossover}{eq:strategy_region}), the
$j$-th dimension of new candidate solution $q^a_i$ and $q^b_i$ will
also be checked to see if $\phi_c < \alpha_i$ or $j$ is equal to a
random number $j_{\text{rand}}$. If yes, $q^a_{i,j}$ will set equal to
$p^a_{i,j}$; otherwise, $q^a_{i,j}$ will be set equal to
$p_{i,j}$. Also, the same checks and updates will also be used for
$q^b_{i,j}$.
After that, the new candidate solution $q_i$ will be generated as follows:
\begin{equation}
    q_{i} = \begin{cases}
        \nu, & \text{if}~i = 1, \\
        \text{Dist}_{\min} (q^a_i, q^b_i, \nu), & \text{if}~i \neq 1~\text{and}~\phi < \delta, \\
        \text{Dist}_{\min} (q^a_i, q^b_i,  p_i), & \text{otherwise},
    \end{cases}
    \label{eq:closest_net}
\end{equation}
where $\text{Dist}_{\min}(g_1, g_2, g_3)$ is a function to
find out the most closest solution (i.e., $g_1$ or $g_2$) of solution
$g_3$.
Last, if $f(q_i)$ is better than $f(p_i)$, it will be used to replace
$p_i$ at next iteration.

\subsubsection{Population Adjustment}
One major task of this operator is to adjust the size of populations
$s$ and $x$, as follows:
\begin{equation}
  n=\lambda(\delta^{(1 - \sqrt{\delta})})_{n_{\text{init}}}^{n_{\text{end}}},
  \label{eq:ps_n1n2}
\end{equation}
where $n$ can be either the value of $n_s$ or $n_x$.
By using this equation to adjust the population size of $s$, $n_s$
will be set equal to $n$, $n_{\text{init}}$ a predefined population
size of $s$ at the very beginning, and $n_{\text{end}}$ a predefined
population size of $s$ at the end during the convergence process of
{\xsnos}.  Of course, this equation will also be used to adjust the
population size of $x$ during the convergence process.
The main difference between $s$ and $x$ for the population resize
using \xeq{eq:ps_n1n2} is that the population size of $s$ will be
reduced gradually while the population size of $x$ will be increased
gradually. For this reason, the values of $n_{\text{init}}$ is larger
than $n_{\text{end}}$ for $s$ while $n_{\text{init}}$ is smaller than
$n_{\text{end}}$ for $x$.

In addition to the population size adjustment, this operator will also
generate the new candidate solution $v_p$ that is the third way to
generate a new candidate solution of {\xsnos}. Note that the other two
are by using the RS and PS operators.  The $j$-th dimension of $v_p$
will be generated as follows:
\begin{equation}
    v_{p, j} = \begin{cases}
        \delta^2 \cdot p_{\phi_i,j}^{\text{top-}\rho} + (1 - \delta^2) \cdot \phi^{U_i}_{L_i}, & \text{if}~\phi < 0.5, \\
        p_{\phi_i,j}^{\text{top-}\rho}, & \text{otherwise},
    \end{cases}
    \label{eq:create_pop}
\end{equation}
where $p_{\phi_i}^{\text{top-}\rho}$ is a randomly selected elastic
point from the top-$\rho$ elastic points in terms of the objective
value, which is the same as \xeq{eq:strategy_point}, and
$\phi^{U_i}_{L_i}$ is a random number uniformly distributed in the
range $[L_i, U_i]$.

\subsubsection{Summary}

A detailed discussion of the proposed algorithm was given in
\xsec{ssec:algorithm} that is used for solving SOPs in \xsec{sec2}.
We will now turn our discussion to the roles and relationships of
three populations and how to generate new candidate solutions by three
different operators, to explain the distinguishing features of
{\xsnos}.  One of the distinguishing features of {\xsnos} is that it
will generate \emph{three populations} for the search. Among them,
explorers $s$ and miners $x$ can be regarded as the searched solutions
at each iteration while elastic points $p$ that belong to space net
can be regarded as the accumulated information of all searched
solutions from the very beginning until the current iteration. The
population size of $s$, $x$, and $p$ are $n_s$, $n_x$, and $n_p$,
respectively, and $n_a$ is used to determine how many elastic points
will be affected by new solutions $u$ and $v$.  The roles of explorers
and miners are for the global and local search during the convergence
process, which imply that explorers will search for larger scope than
miners in the solution space. That is why the population size of $s$
will be decreased while the population size of $x$ will be increased
during the convergence process.  The role of $p$ is to keep and
provide information of the landscape of the solution space, which will
be used in the calculation of the expected value so that {\xsnos} will
make accurate searches. The number of elastic points in $p$ will not
be changed during the convergence process.  Another distinguishing
feature of {\xsnos} is that it will generate new candidate solutions
by region search (i.e., a certain number of new explorers $u$), point
search (i.e., a certain number of new miners $v$), and population
adjustment (i.e., a certain number of new miners $v_p$).  In addition
to point search, {\xsnos} uses population adjustment to generate
another new miner $v_p$ because (1) it needs to increase the number of
miners when {\xsnos} determines to decrease the population size of $s$
and increase the population size of $x$, and (2) it uses another
strategy to generate the new candidate solution to increase the search
diversity.

\bibliographystyle{apalike}

\end{document}

%% file: lstset.tex
\usepackage{listings}
\usepackage{color}

\definecolor{olivegreen}{rgb}{0,0.5,0.3}

%% file: macros.tex
\newcommand{\xeq}[1]{Eq.~(\ref{#1})}
\newcommand{\xeqs}[2]{Eqs.~(\ref{#1}) and~(\ref{#2})}

\newcommand{\xfig}[1]{Fig.~\ref{#1}}
\newcommand{\xfiga}[1]{Figs.~\ref{#1}}

\newcommand{\xsec}[1]{Section~\ref{#1}}

\newcommand{\xalg}[1]{Algorithm~\ref{#1}}

\newcommand{\xsno}{space net optimization}
\newcommand{\xsnos}{SNO}
\newcommand{\xsnoa}{\xsno\ (\xsnos)}

\newcommand{\xmetaha}{metaheuristic algorithm}
\newcommand{\xMetaha}{Metaheuristic algorithm}

\newcommand{\xsop}{single objective bound constrained problem}
\newcommand{\xsops}{SOP}
\newcommand{\xsopa}{\xsop\ (\xsops)}

\newcommand{\xalgsize}{\footnotesize}

\definecolor{darkred1}{rgb}{0.8, 0.0, 0.0}
\definecolor{darkred2}{rgb}{0.8, 0.25, 0.33}
\definecolor{darkred3}{rgb}{0.87, 0.19, 0.39}
\definecolor{darkred4}{rgb}{0.7, 0.11, 0.11}
\definecolor{darkblue}{rgb}{0.0, 0.0, 0.8}

\iffinal

\else
\usepackage[normalem]{ulem}

\newcommand{\xe}[1]{\textcolor{darkblue}{#1}} 
\fi

\newtheorem{xdefinition}{Definition}

%% file: fig/overall-3.pdf_t
\begin{picture}(0,0)%
\includegraphics{fig/overall-3.pdf}%
\end{picture}%
\setlength{\unitlength}{3947sp}%
\begingroup\makeatletter\ifx\SetFigFont\undefined%
\gdef\SetFigFont#1#2#3#4#5{%
  \reset@font\fontsize{#1}{#2pt}%
  \fontfamily{#3}\fontseries{#4}\fontshape{#5}%
  \selectfont}%
\fi\endgroup%
\begin{picture}(12971,10527)(-3658,-10576)
\put(-2999,-4186){\makebox(0,0)[lb]{\smash{{\SetFigFont{14}{16.8}{\familydefault}{\mddefault}{\updefault}{\color[rgb]{0,0,0}0.8}%
}}}}
\put(-2399,-4186){\makebox(0,0)[lb]{\smash{{\SetFigFont{14}{16.8}{\familydefault}{\mddefault}{\updefault}{\color[rgb]{0,0,0}0.7}%
}}}}
\put(-1799,-4186){\makebox(0,0)[lb]{\smash{{\SetFigFont{14}{16.8}{\familydefault}{\mddefault}{\updefault}{\color[rgb]{0,0,0}0.9}%
}}}}
\put(3609,-8544){\makebox(0,0)[lb]{\smash{{\SetFigFont{10}{12.0}{\familydefault}{\mddefault}{\updefault}{\color[rgb]{0,0,0}Add}%
}}}}
\put(-2249,-586){\makebox(0,0)[b]{\smash{{\SetFigFont{12}{14.4}{\familydefault}{\mddefault}{\updefault}{\color[rgb]{0,0,0}Elastic Points $p$}%
}}}}
\put(-663,-1036){\makebox(0,0)[b]{\smash{{\SetFigFont{11}{13.2}{\familydefault}{\mddefault}{\updefault}{\color[rgb]{0,0,0}$r_1$}%
}}}}
\put(-149,-1036){\makebox(0,0)[b]{\smash{{\SetFigFont{11}{13.2}{\familydefault}{\mddefault}{\updefault}{\color[rgb]{0,0,0}$r_2$}%
}}}}
\put(365,-1036){\makebox(0,0)[b]{\smash{{\SetFigFont{11}{13.2}{\familydefault}{\mddefault}{\updefault}{\color[rgb]{0,0,0}$r_3$}%
}}}}
\put(-663,-1550){\makebox(0,0)[b]{\smash{{\SetFigFont{11}{13.2}{\familydefault}{\mddefault}{\updefault}{\color[rgb]{0,0,0}$r_4$}%
}}}}
\put(-149,-1550){\makebox(0,0)[b]{\smash{{\SetFigFont{11}{13.2}{\familydefault}{\mddefault}{\updefault}{\color[rgb]{0,0,0}$r_5$}%
}}}}
\put(365,-1550){\makebox(0,0)[b]{\smash{{\SetFigFont{11}{13.2}{\familydefault}{\mddefault}{\updefault}{\color[rgb]{0,0,0}$r_6$}%
}}}}
\put(-663,-2065){\makebox(0,0)[b]{\smash{{\SetFigFont{11}{13.2}{\familydefault}{\mddefault}{\updefault}{\color[rgb]{0,0,0}$r_7$}%
}}}}
\put(-149,-2065){\makebox(0,0)[b]{\smash{{\SetFigFont{11}{13.2}{\familydefault}{\mddefault}{\updefault}{\color[rgb]{0,0,0}$r_8$}%
}}}}
\put(365,-2065){\makebox(0,0)[b]{\smash{{\SetFigFont{11}{13.2}{\familydefault}{\mddefault}{\updefault}{\color[rgb]{0,0,0}$r_9$}%
}}}}
\put(-149,-586){\makebox(0,0)[b]{\smash{{\SetFigFont{12}{14.4}{\familydefault}{\mddefault}{\updefault}{\color[rgb]{0,0,0}Regions $r$}%
}}}}
\put(1951,-586){\makebox(0,0)[b]{\smash{{\SetFigFont{12}{14.4}{\familydefault}{\mddefault}{\updefault}{\color[rgb]{0,0,0}Expected Value $e$}%
}}}}
\put(1426,-1036){\makebox(0,0)[b]{\smash{{\SetFigFont{11}{13.2}{\familydefault}{\mddefault}{\updefault}{\color[rgb]{0,0,0}$e_1$}%
}}}}
\put(1951,-1036){\makebox(0,0)[b]{\smash{{\SetFigFont{11}{13.2}{\familydefault}{\mddefault}{\updefault}{\color[rgb]{0,0,0}$e_2$}%
}}}}
\put(2476,-1036){\makebox(0,0)[b]{\smash{{\SetFigFont{11}{13.2}{\familydefault}{\mddefault}{\updefault}{\color[rgb]{0,0,0}$e_3$}%
}}}}
\put(1426,-1561){\makebox(0,0)[b]{\smash{{\SetFigFont{11}{13.2}{\familydefault}{\mddefault}{\updefault}{\color[rgb]{0,0,0}$e_4$}%
}}}}
\put(1951,-1561){\makebox(0,0)[b]{\smash{{\SetFigFont{11}{13.2}{\familydefault}{\mddefault}{\updefault}{\color[rgb]{0,0,0}$e_5$}%
}}}}
\put(2476,-1561){\makebox(0,0)[b]{\smash{{\SetFigFont{11}{13.2}{\familydefault}{\mddefault}{\updefault}{\color[rgb]{0,0,0}$e_6$}%
}}}}
\put(1426,-2086){\makebox(0,0)[b]{\smash{{\SetFigFont{11}{13.2}{\familydefault}{\mddefault}{\updefault}{\color[rgb]{0,0,0}$e_7$}%
}}}}
\put(1951,-2086){\makebox(0,0)[b]{\smash{{\SetFigFont{11}{13.2}{\familydefault}{\mddefault}{\updefault}{\color[rgb]{0,0,0}$e_8$}%
}}}}
\put(2476,-2086){\makebox(0,0)[b]{\smash{{\SetFigFont{11}{13.2}{\familydefault}{\mddefault}{\updefault}{\color[rgb]{0,0,0}$e_9$}%
}}}}
\put(3976,-586){\makebox(0,0)[b]{\smash{{\SetFigFont{12}{14.4}{\familydefault}{\mddefault}{\updefault}{\color[rgb]{0,0,0}Population $s$}%
}}}}
\put(-2631,-9076){\makebox(0,0)[lb]{\smash{{\SetFigFont{10}{12.0}{\familydefault}{\mddefault}{\updefault}{\color[rgb]{0,0,0}Delete}%
}}}}
\put(7576,-1171){\makebox(0,0)[lb]{\smash{{\SetFigFont{10}{12.0}{\familydefault}{\mddefault}{\updefault}{\color[rgb]{0,0,0}A Solution in $s$}%
}}}}
\put(7576,-1471){\makebox(0,0)[lb]{\smash{{\SetFigFont{10}{12.0}{\familydefault}{\mddefault}{\updefault}{\color[rgb]{0,0,0}A Solution in $x$}%
}}}}
\put(7576,-2071){\makebox(0,0)[lb]{\smash{{\SetFigFont{10}{12.0}{\familydefault}{\mddefault}{\updefault}{\color[rgb]{0,0,0}A Possible Solution of $x$}%
}}}}
\put(7576,-1771){\makebox(0,0)[lb]{\smash{{\SetFigFont{10}{12.0}{\familydefault}{\mddefault}{\updefault}{\color[rgb]{0,0,0}A Possible Solution of $s$}%
}}}}
\put(7576,-2371){\makebox(0,0)[lb]{\smash{{\SetFigFont{10}{12.0}{\familydefault}{\mddefault}{\updefault}{\color[rgb]{0,0,0}A New Solution of $s$ or $x$}%
}}}}
\put(7576,-871){\makebox(0,0)[lb]{\smash{{\SetFigFont{10}{12.0}{\familydefault}{\mddefault}{\updefault}{\color[rgb]{0,0,0}An Elastic Point}%
}}}}
\put(3076,-2986){\makebox(0,0)[lb]{\smash{{\SetFigFont{12}{14.4}{\familydefault}{\mddefault}{\updefault}{\color[rgb]{0,0,0}Space Net Adjustment (SNA)}%
}}}}
\put(-3374,-2461){\rotatebox{90.0}{\makebox(0,0)[lb]{\smash{{\SetFigFont{12}{14.4}{\familydefault}{\mddefault}{\updefault}{\color[rgb]{0,0,0}$t_0$}%
}}}}}
\put(-3524,-10111){\rotatebox{90.0}{\makebox(0,0)[lb]{\smash{{\SetFigFont{12}{14.4}{\familydefault}{\mddefault}{\updefault}{\color[rgb]{0,0,0}$t$++}%
}}}}}
\put(-3374,-10561){\rotatebox{90.0}{\makebox(0,0)[lb]{\smash{{\SetFigFont{12}{14.4}{\familydefault}{\mddefault}{\updefault}{\color[rgb]{0,0,0}$t_{\max}$}%
}}}}}
\put(-3224,-7786){\makebox(0,0)[lb]{\smash{{\SetFigFont{12}{14.4}{\familydefault}{\mddefault}{\updefault}{\color[rgb]{0,0,0}Populations Adjustment (PA)}%
}}}}
\put(-3224,-5386){\makebox(0,0)[lb]{\smash{{\SetFigFont{12}{14.4}{\familydefault}{\mddefault}{\updefault}{\color[rgb]{0,0,0}Point Search (PS)}%
}}}}
\put(-3224,-10486){\makebox(0,0)[lb]{\smash{{\SetFigFont{12}{14.4}{\familydefault}{\mddefault}{\updefault}{\color[rgb]{0,0,0}Output Result}%
}}}}
\put(-3224,-286){\makebox(0,0)[lb]{\smash{{\SetFigFont{12}{14.4}{\familydefault}{\mddefault}{\updefault}{\color[rgb]{0,0,0}Initialiation}%
}}}}
\put(6151,-586){\makebox(0,0)[b]{\smash{{\SetFigFont{12}{14.4}{\familydefault}{\mddefault}{\updefault}{\color[rgb]{0,0,0}Population $x$}%
}}}}
\put(6151,-9886){\makebox(0,0)[b]{\smash{{\SetFigFont{12}{14.4}{\familydefault}{\mddefault}{\updefault}{\color[rgb]{0,0,0}$x=x+v_p$}%
}}}}
\put(8251,-9886){\makebox(0,0)[b]{\smash{{\SetFigFont{12}{14.4}{\familydefault}{\mddefault}{\updefault}{\color[rgb]{0,0,0}$s$ and $x$}%
}}}}
\put(-149,-9886){\makebox(0,0)[b]{\smash{{\SetFigFont{12}{14.4}{\familydefault}{\mddefault}{\updefault}{\color[rgb]{0,0,0}$s=s\setminus s_w$}%
}}}}
\put(4051,-8911){\makebox(0,0)[b]{\smash{{\SetFigFont{12}{14.4}{\familydefault}{\mddefault}{\updefault}{\color[rgb]{0,0,0}$v_p$}%
}}}}
\put(1951,-9886){\makebox(0,0)[b]{\smash{{\SetFigFont{12}{14.4}{\familydefault}{\mddefault}{\updefault}{\color[rgb]{0,0,0}$x$}%
}}}}
\put(-2474,-9436){\makebox(0,0)[b]{\smash{{\SetFigFont{12}{14.4}{\familydefault}{\mddefault}{\updefault}{\color[rgb]{0,0,0}$s_w$}%
}}}}
\put(-3224,-2986){\makebox(0,0)[lb]{\smash{{\SetFigFont{12}{14.4}{\familydefault}{\mddefault}{\updefault}{\color[rgb]{0,0,0}Region Search (RS)}%
}}}}
\put(5551,-2986){\makebox(0,0)[lb]{\smash{{\SetFigFont{12}{14.4}{\familydefault}{\mddefault}{\updefault}{\color[rgb]{0,0,0}$p \rightarrow q \rightarrow p$}%
}}}}
\put(-1574,-2986){\makebox(0,0)[lb]{\smash{{\SetFigFont{12}{14.4}{\familydefault}{\mddefault}{\updefault}{\color[rgb]{0,0,0}$s \rightarrow u \rightarrow s$}%
}}}}
\put(-1724,-5386){\makebox(0,0)[lb]{\smash{{\SetFigFont{12}{14.4}{\familydefault}{\mddefault}{\updefault}{\color[rgb]{0,0,0}$x \rightarrow v \rightarrow x$}%
}}}}
\put(-749,-7786){\makebox(0,0)[lb]{\smash{{\SetFigFont{12}{14.4}{\familydefault}{\mddefault}{\updefault}{\color[rgb]{0,0,0}$s=s\setminus s_w$, $x=x+v_p$}%
}}}}
\put(5551,-5386){\makebox(0,0)[lb]{\smash{{\SetFigFont{12}{14.4}{\familydefault}{\mddefault}{\updefault}{\color[rgb]{0,0,0}$p \rightarrow q \rightarrow p$}%
}}}}
\put(3076,-5386){\makebox(0,0)[lb]{\smash{{\SetFigFont{12}{14.4}{\familydefault}{\mddefault}{\updefault}{\color[rgb]{0,0,0}Space Net Adjustment (SNA)}%
}}}}
\put(7576,-571){\makebox(0,0)[lb]{\smash{{\SetFigFont{10}{12.0}{\familydefault}{\mddefault}{\updefault}{\color[rgb]{0,0,0}Space Net}%
}}}}
\put(-1349,-3886){\makebox(0,0)[b]{\smash{{\SetFigFont{10}{12.0}{\familydefault}{\mddefault}{\updefault}{\color[rgb]{0.000,0.000,0.000}8}%
}}}}
\put(-1949,-3886){\makebox(0,0)[b]{\smash{{\SetFigFont{10}{12.0}{\familydefault}{\mddefault}{\updefault}{\color[rgb]{0.000,0.000,0.000}7}%
}}}}
\put(-2549,-3886){\makebox(0,0)[b]{\smash{{\SetFigFont{10}{12.0}{\familydefault}{\mddefault}{\updefault}{\color[rgb]{0.000,0.000,0.000}6}%
}}}}
\put(-3149,-3886){\makebox(0,0)[b]{\smash{{\SetFigFont{10}{12.0}{\familydefault}{\mddefault}{\updefault}{\color[rgb]{0.000,0.000,0.000}5}%
}}}}
\put(-1349,-4486){\makebox(0,0)[b]{\smash{{\SetFigFont{10}{12.0}{\familydefault}{\mddefault}{\updefault}{\color[rgb]{0.000,0.000,0.000}12}%
}}}}
\put(-1949,-4486){\makebox(0,0)[b]{\smash{{\SetFigFont{10}{12.0}{\familydefault}{\mddefault}{\updefault}{\color[rgb]{0.000,0.000,0.000}11}%
}}}}
\put(-2549,-4486){\makebox(0,0)[b]{\smash{{\SetFigFont{10}{12.0}{\familydefault}{\mddefault}{\updefault}{\color[rgb]{0.000,0.000,0.000}10}%
}}}}
\put(-3149,-4486){\makebox(0,0)[b]{\smash{{\SetFigFont{10}{12.0}{\familydefault}{\mddefault}{\updefault}{\color[rgb]{0.000,0.000,0.000}9}%
}}}}
\put(-1349,-5686){\makebox(0,0)[b]{\smash{{\SetFigFont{10}{12.0}{\familydefault}{\mddefault}{\updefault}{\color[rgb]{0.000,0.000,0.000}4}%
}}}}
\put(-1441,-6513){\makebox(0,0)[b]{\smash{{\SetFigFont{10}{12.0}{\familydefault}{\mddefault}{\updefault}{\color[rgb]{0.000,0.000,0.000}8}%
}}}}
\put(-1411,-6776){\makebox(0,0)[b]{\smash{{\SetFigFont{10}{12.0}{\familydefault}{\mddefault}{\updefault}{\color[rgb]{0.000,0.000,0.000}12}%
}}}}
\put(-1701,-6417){\makebox(0,0)[b]{\smash{{\SetFigFont{10}{12.0}{\familydefault}{\mddefault}{\updefault}{\color[rgb]{0.000,0.000,0.000}7}%
}}}}
\put(-2838,-5832){\makebox(0,0)[b]{\smash{{\SetFigFont{10}{12.0}{\familydefault}{\mddefault}{\updefault}{\color[rgb]{0.000,0.000,0.000}1}%
}}}}
\put(-2296,-5828){\makebox(0,0)[b]{\smash{{\SetFigFont{10}{12.0}{\familydefault}{\mddefault}{\updefault}{\color[rgb]{0.000,0.000,0.000}2}%
}}}}
\put(-1704,-5913){\makebox(0,0)[b]{\smash{{\SetFigFont{10}{12.0}{\familydefault}{\mddefault}{\updefault}{\color[rgb]{0.000,0.000,0.000}3}%
}}}}
\put(-2340,-6242){\makebox(0,0)[b]{\smash{{\SetFigFont{10}{12.0}{\familydefault}{\mddefault}{\updefault}{\color[rgb]{0.000,0.000,0.000}6}%
}}}}
\put(-2865,-6272){\makebox(0,0)[b]{\smash{{\SetFigFont{10}{12.0}{\familydefault}{\mddefault}{\updefault}{\color[rgb]{0.000,0.000,0.000}5}%
}}}}
\put(-2843,-6863){\makebox(0,0)[b]{\smash{{\SetFigFont{10}{12.0}{\familydefault}{\mddefault}{\updefault}{\color[rgb]{0.000,0.000,0.000}9}%
}}}}
\put(-2763,-7225){\makebox(0,0)[b]{\smash{{\SetFigFont{10}{12.0}{\familydefault}{\mddefault}{\updefault}{\color[rgb]{0.000,0.000,0.000}13}%
}}}}
\put(-2261,-7198){\makebox(0,0)[b]{\smash{{\SetFigFont{10}{12.0}{\familydefault}{\mddefault}{\updefault}{\color[rgb]{0.000,0.000,0.000}14}%
}}}}
\put(-1780,-7188){\makebox(0,0)[b]{\smash{{\SetFigFont{10}{12.0}{\familydefault}{\mddefault}{\updefault}{\color[rgb]{0.000,0.000,0.000}15}%
}}}}
\put(-1402,-7136){\makebox(0,0)[b]{\smash{{\SetFigFont{10}{12.0}{\familydefault}{\mddefault}{\updefault}{\color[rgb]{0.000,0.000,0.000}16}%
}}}}
\put(-2309,-6711){\makebox(0,0)[b]{\smash{{\SetFigFont{10}{12.0}{\familydefault}{\mddefault}{\updefault}{\color[rgb]{0.000,0.000,0.000}10}%
}}}}
\put(-1842,-6614){\makebox(0,0)[b]{\smash{{\SetFigFont{10}{12.0}{\familydefault}{\mddefault}{\updefault}{\color[rgb]{0.000,0.000,0.000}11}%
}}}}
\put(9059,-6513){\makebox(0,0)[b]{\smash{{\SetFigFont{10}{12.0}{\familydefault}{\mddefault}{\updefault}{\color[rgb]{0.000,0.000,0.000}8}%
}}}}
\put(7737,-7225){\makebox(0,0)[b]{\smash{{\SetFigFont{10}{12.0}{\familydefault}{\mddefault}{\updefault}{\color[rgb]{0.000,0.000,0.000}13}%
}}}}
\put(8863,-5893){\makebox(0,0)[b]{\smash{{\SetFigFont{10}{12.0}{\familydefault}{\mddefault}{\updefault}{\color[rgb]{0.000,0.000,0.000}3}%
}}}}
\put(9114,-5709){\makebox(0,0)[b]{\smash{{\SetFigFont{10}{12.0}{\familydefault}{\mddefault}{\updefault}{\color[rgb]{0.000,0.000,0.000}4}%
}}}}
\put(7854,-5977){\makebox(0,0)[b]{\smash{{\SetFigFont{10}{12.0}{\familydefault}{\mddefault}{\updefault}{\color[rgb]{0.000,0.000,0.000}1}%
}}}}
\put(8305,-5952){\makebox(0,0)[b]{\smash{{\SetFigFont{10}{12.0}{\familydefault}{\mddefault}{\updefault}{\color[rgb]{0.000,0.000,0.000}2}%
}}}}
\put(8292,-6702){\makebox(0,0)[b]{\smash{{\SetFigFont{10}{12.0}{\familydefault}{\mddefault}{\updefault}{\color[rgb]{0.000,0.000,0.000}10}%
}}}}
\put(7849,-6837){\makebox(0,0)[b]{\smash{{\SetFigFont{10}{12.0}{\familydefault}{\mddefault}{\updefault}{\color[rgb]{0.000,0.000,0.000}9}%
}}}}
\put(8248,-7026){\makebox(0,0)[b]{\smash{{\SetFigFont{10}{12.0}{\familydefault}{\mddefault}{\updefault}{\color[rgb]{0.000,0.000,0.000}14}%
}}}}
\put(8703,-7043){\makebox(0,0)[b]{\smash{{\SetFigFont{10}{12.0}{\familydefault}{\mddefault}{\updefault}{\color[rgb]{0.000,0.000,0.000}15}%
}}}}
\put(8687,-6532){\makebox(0,0)[b]{\smash{{\SetFigFont{10}{12.0}{\familydefault}{\mddefault}{\updefault}{\color[rgb]{0.000,0.000,0.000}11}%
}}}}
\put(8853,-6261){\makebox(0,0)[b]{\smash{{\SetFigFont{10}{12.0}{\familydefault}{\mddefault}{\updefault}{\color[rgb]{0.000,0.000,0.000}7}%
}}}}
\put(8962,-6770){\makebox(0,0)[b]{\smash{{\SetFigFont{10}{12.0}{\familydefault}{\mddefault}{\updefault}{\color[rgb]{0.000,0.000,0.000}12}%
}}}}
\put(9104,-7128){\makebox(0,0)[b]{\smash{{\SetFigFont{10}{12.0}{\familydefault}{\mddefault}{\updefault}{\color[rgb]{0.000,0.000,0.000}16}%
}}}}
\put(7702,-6244){\makebox(0,0)[b]{\smash{{\SetFigFont{10}{12.0}{\familydefault}{\mddefault}{\updefault}{\color[rgb]{0.000,0.000,0.000}5}%
}}}}
\put(8160,-6178){\makebox(0,0)[b]{\smash{{\SetFigFont{10}{12.0}{\familydefault}{\mddefault}{\updefault}{\color[rgb]{0.000,0.000,0.000}6}%
}}}}
\put(6763,-5893){\makebox(0,0)[b]{\smash{{\SetFigFont{10}{12.0}{\familydefault}{\mddefault}{\updefault}{\color[rgb]{0.000,0.000,0.000}3}%
}}}}
\put(7014,-5709){\makebox(0,0)[b]{\smash{{\SetFigFont{10}{12.0}{\familydefault}{\mddefault}{\updefault}{\color[rgb]{0.000,0.000,0.000}4}%
}}}}
\put(6803,-6269){\makebox(0,0)[b]{\smash{{\SetFigFont{10}{12.0}{\familydefault}{\mddefault}{\updefault}{\color[rgb]{0.000,0.000,0.000}7}%
}}}}
\put(9151,-3286){\makebox(0,0)[b]{\smash{{\SetFigFont{10}{12.0}{\familydefault}{\mddefault}{\updefault}{\color[rgb]{0.000,0.000,0.000}4}%
}}}}
\put(9059,-4113){\makebox(0,0)[b]{\smash{{\SetFigFont{10}{12.0}{\familydefault}{\mddefault}{\updefault}{\color[rgb]{0.000,0.000,0.000}8}%
}}}}
\put(9089,-4376){\makebox(0,0)[b]{\smash{{\SetFigFont{10}{12.0}{\familydefault}{\mddefault}{\updefault}{\color[rgb]{0.000,0.000,0.000}12}%
}}}}
\put(8799,-4017){\makebox(0,0)[b]{\smash{{\SetFigFont{10}{12.0}{\familydefault}{\mddefault}{\updefault}{\color[rgb]{0.000,0.000,0.000}7}%
}}}}
\put(7662,-3432){\makebox(0,0)[b]{\smash{{\SetFigFont{10}{12.0}{\familydefault}{\mddefault}{\updefault}{\color[rgb]{0.000,0.000,0.000}1}%
}}}}
\put(8204,-3428){\makebox(0,0)[b]{\smash{{\SetFigFont{10}{12.0}{\familydefault}{\mddefault}{\updefault}{\color[rgb]{0.000,0.000,0.000}2}%
}}}}
\put(8796,-3513){\makebox(0,0)[b]{\smash{{\SetFigFont{10}{12.0}{\familydefault}{\mddefault}{\updefault}{\color[rgb]{0.000,0.000,0.000}3}%
}}}}
\put(8160,-3842){\makebox(0,0)[b]{\smash{{\SetFigFont{10}{12.0}{\familydefault}{\mddefault}{\updefault}{\color[rgb]{0.000,0.000,0.000}6}%
}}}}
\put(7635,-3872){\makebox(0,0)[b]{\smash{{\SetFigFont{10}{12.0}{\familydefault}{\mddefault}{\updefault}{\color[rgb]{0.000,0.000,0.000}5}%
}}}}
\put(7657,-4463){\makebox(0,0)[b]{\smash{{\SetFigFont{10}{12.0}{\familydefault}{\mddefault}{\updefault}{\color[rgb]{0.000,0.000,0.000}9}%
}}}}
\put(7737,-4825){\makebox(0,0)[b]{\smash{{\SetFigFont{10}{12.0}{\familydefault}{\mddefault}{\updefault}{\color[rgb]{0.000,0.000,0.000}13}%
}}}}
\put(8239,-4798){\makebox(0,0)[b]{\smash{{\SetFigFont{10}{12.0}{\familydefault}{\mddefault}{\updefault}{\color[rgb]{0.000,0.000,0.000}14}%
}}}}
\put(8720,-4788){\makebox(0,0)[b]{\smash{{\SetFigFont{10}{12.0}{\familydefault}{\mddefault}{\updefault}{\color[rgb]{0.000,0.000,0.000}15}%
}}}}
\put(9098,-4736){\makebox(0,0)[b]{\smash{{\SetFigFont{10}{12.0}{\familydefault}{\mddefault}{\updefault}{\color[rgb]{0.000,0.000,0.000}16}%
}}}}
\put(8191,-4311){\makebox(0,0)[b]{\smash{{\SetFigFont{10}{12.0}{\familydefault}{\mddefault}{\updefault}{\color[rgb]{0.000,0.000,0.000}10}%
}}}}
\put(8658,-4214){\makebox(0,0)[b]{\smash{{\SetFigFont{10}{12.0}{\familydefault}{\mddefault}{\updefault}{\color[rgb]{0.000,0.000,0.000}11}%
}}}}
\put(6959,-4113){\makebox(0,0)[b]{\smash{{\SetFigFont{10}{12.0}{\familydefault}{\mddefault}{\updefault}{\color[rgb]{0.000,0.000,0.000}8}%
}}}}
\put(6989,-4376){\makebox(0,0)[b]{\smash{{\SetFigFont{10}{12.0}{\familydefault}{\mddefault}{\updefault}{\color[rgb]{0.000,0.000,0.000}12}%
}}}}
\put(6699,-4017){\makebox(0,0)[b]{\smash{{\SetFigFont{10}{12.0}{\familydefault}{\mddefault}{\updefault}{\color[rgb]{0.000,0.000,0.000}7}%
}}}}
\put(751,-5686){\makebox(0,0)[b]{\smash{{\SetFigFont{10}{12.0}{\familydefault}{\mddefault}{\updefault}{\color[rgb]{0.000,0.000,0.000}4}%
}}}}
\put(7426,-871){\makebox(0,0)[b]{\smash{{\SetFigFont{10}{12.0}{\familydefault}{\mddefault}{\updefault}{\color[rgb]{0.000,0.000,0.000}1}%
}}}}
\put(622,-2322){\makebox(0,0)[b]{\smash{{\SetFigFont{9}{10.8}{\familydefault}{\mddefault}{\updefault}{\color[rgb]{0.000,0.000,0.000}16}%
}}}}
\put(622,-1807){\makebox(0,0)[b]{\smash{{\SetFigFont{9}{10.8}{\familydefault}{\mddefault}{\updefault}{\color[rgb]{0.000,0.000,0.000}12}%
}}}}
\put(-1478,-2322){\makebox(0,0)[b]{\smash{{\SetFigFont{9}{10.8}{\familydefault}{\mddefault}{\updefault}{\color[rgb]{0.000,0.000,0.000}16}%
}}}}
\put(-1478,-1807){\makebox(0,0)[b]{\smash{{\SetFigFont{9}{10.8}{\familydefault}{\mddefault}{\updefault}{\color[rgb]{0.000,0.000,0.000}12}%
}}}}
\put(-1992,-2322){\makebox(0,0)[b]{\smash{{\SetFigFont{9}{10.8}{\familydefault}{\mddefault}{\updefault}{\color[rgb]{0.000,0.000,0.000}15}%
}}}}
\put(-2506,-2322){\makebox(0,0)[b]{\smash{{\SetFigFont{9}{10.8}{\familydefault}{\mddefault}{\updefault}{\color[rgb]{0.000,0.000,0.000}14}%
}}}}
\put(-3020,-2322){\makebox(0,0)[b]{\smash{{\SetFigFont{9}{10.8}{\familydefault}{\mddefault}{\updefault}{\color[rgb]{0.000,0.000,0.000}13}%
}}}}
\put(-1992,-1807){\makebox(0,0)[b]{\smash{{\SetFigFont{9}{10.8}{\familydefault}{\mddefault}{\updefault}{\color[rgb]{0.000,0.000,0.000}11}%
}}}}
\put(-2506,-1807){\makebox(0,0)[b]{\smash{{\SetFigFont{9}{10.8}{\familydefault}{\mddefault}{\updefault}{\color[rgb]{0.000,0.000,0.000}10}%
}}}}
\put(-3020,-1807){\makebox(0,0)[b]{\smash{{\SetFigFont{9}{10.8}{\familydefault}{\mddefault}{\updefault}{\color[rgb]{0.000,0.000,0.000}9}%
}}}}
\put(-1478,-1293){\makebox(0,0)[b]{\smash{{\SetFigFont{9}{10.8}{\familydefault}{\mddefault}{\updefault}{\color[rgb]{0.000,0.000,0.000}8}%
}}}}
\put(-1992,-1293){\makebox(0,0)[b]{\smash{{\SetFigFont{9}{10.8}{\familydefault}{\mddefault}{\updefault}{\color[rgb]{0.000,0.000,0.000}7}%
}}}}
\put(-2506,-1293){\makebox(0,0)[b]{\smash{{\SetFigFont{9}{10.8}{\familydefault}{\mddefault}{\updefault}{\color[rgb]{0.000,0.000,0.000}6}%
}}}}
\put(-3020,-1293){\makebox(0,0)[b]{\smash{{\SetFigFont{9}{10.8}{\familydefault}{\mddefault}{\updefault}{\color[rgb]{0.000,0.000,0.000}5}%
}}}}
\put(-1478,-779){\makebox(0,0)[b]{\smash{{\SetFigFont{9}{10.8}{\familydefault}{\mddefault}{\updefault}{\color[rgb]{0.000,0.000,0.000}4}%
}}}}
\put(-1992,-779){\makebox(0,0)[b]{\smash{{\SetFigFont{9}{10.8}{\familydefault}{\mddefault}{\updefault}{\color[rgb]{0.000,0.000,0.000}3}%
}}}}
\put(-2506,-779){\makebox(0,0)[b]{\smash{{\SetFigFont{9}{10.8}{\familydefault}{\mddefault}{\updefault}{\color[rgb]{0.000,0.000,0.000}2}%
}}}}
\put(-3020,-779){\makebox(0,0)[b]{\smash{{\SetFigFont{9}{10.8}{\familydefault}{\mddefault}{\updefault}{\color[rgb]{0.000,0.000,0.000}1}%
}}}}
\put(108,-2322){\makebox(0,0)[b]{\smash{{\SetFigFont{9}{10.8}{\familydefault}{\mddefault}{\updefault}{\color[rgb]{0.000,0.000,0.000}15}%
}}}}
\put(-406,-2322){\makebox(0,0)[b]{\smash{{\SetFigFont{9}{10.8}{\familydefault}{\mddefault}{\updefault}{\color[rgb]{0.000,0.000,0.000}14}%
}}}}
\put(-920,-2322){\makebox(0,0)[b]{\smash{{\SetFigFont{9}{10.8}{\familydefault}{\mddefault}{\updefault}{\color[rgb]{0.000,0.000,0.000}13}%
}}}}
\put(108,-1807){\makebox(0,0)[b]{\smash{{\SetFigFont{9}{10.8}{\familydefault}{\mddefault}{\updefault}{\color[rgb]{0.000,0.000,0.000}11}%
}}}}
\put(-406,-1807){\makebox(0,0)[b]{\smash{{\SetFigFont{9}{10.8}{\familydefault}{\mddefault}{\updefault}{\color[rgb]{0.000,0.000,0.000}10}%
}}}}
\put(-920,-1807){\makebox(0,0)[b]{\smash{{\SetFigFont{9}{10.8}{\familydefault}{\mddefault}{\updefault}{\color[rgb]{0.000,0.000,0.000}9}%
}}}}
\put(622,-1293){\makebox(0,0)[b]{\smash{{\SetFigFont{9}{10.8}{\familydefault}{\mddefault}{\updefault}{\color[rgb]{0.000,0.000,0.000}8}%
}}}}
\put(108,-1293){\makebox(0,0)[b]{\smash{{\SetFigFont{9}{10.8}{\familydefault}{\mddefault}{\updefault}{\color[rgb]{0.000,0.000,0.000}7}%
}}}}
\put(-406,-1293){\makebox(0,0)[b]{\smash{{\SetFigFont{9}{10.8}{\familydefault}{\mddefault}{\updefault}{\color[rgb]{0.000,0.000,0.000}6}%
}}}}
\put(-920,-1293){\makebox(0,0)[b]{\smash{{\SetFigFont{9}{10.8}{\familydefault}{\mddefault}{\updefault}{\color[rgb]{0.000,0.000,0.000}5}%
}}}}
\put(622,-779){\makebox(0,0)[b]{\smash{{\SetFigFont{9}{10.8}{\familydefault}{\mddefault}{\updefault}{\color[rgb]{0.000,0.000,0.000}4}%
}}}}
\put(108,-779){\makebox(0,0)[b]{\smash{{\SetFigFont{9}{10.8}{\familydefault}{\mddefault}{\updefault}{\color[rgb]{0.000,0.000,0.000}3}%
}}}}
\put(-406,-779){\makebox(0,0)[b]{\smash{{\SetFigFont{9}{10.8}{\familydefault}{\mddefault}{\updefault}{\color[rgb]{0.000,0.000,0.000}2}%
}}}}
\put(-920,-779){\makebox(0,0)[b]{\smash{{\SetFigFont{9}{10.8}{\familydefault}{\mddefault}{\updefault}{\color[rgb]{0.000,0.000,0.000}1}%
}}}}
\put(2722,-2321){\makebox(0,0)[b]{\smash{{\SetFigFont{9}{10.8}{\familydefault}{\mddefault}{\updefault}{\color[rgb]{0.000,0.000,0.000}16}%
}}}}
\put(2722,-1808){\makebox(0,0)[b]{\smash{{\SetFigFont{9}{10.8}{\familydefault}{\mddefault}{\updefault}{\color[rgb]{0.000,0.000,0.000}12}%
}}}}
\put(2208,-2321){\makebox(0,0)[b]{\smash{{\SetFigFont{9}{10.8}{\familydefault}{\mddefault}{\updefault}{\color[rgb]{0.000,0.000,0.000}15}%
}}}}
\put(1694,-2321){\makebox(0,0)[b]{\smash{{\SetFigFont{9}{10.8}{\familydefault}{\mddefault}{\updefault}{\color[rgb]{0.000,0.000,0.000}14}%
}}}}
\put(2208,-1808){\makebox(0,0)[b]{\smash{{\SetFigFont{9}{10.8}{\familydefault}{\mddefault}{\updefault}{\color[rgb]{0.000,0.000,0.000}11}%
}}}}
\put(1694,-1808){\makebox(0,0)[b]{\smash{{\SetFigFont{9}{10.8}{\familydefault}{\mddefault}{\updefault}{\color[rgb]{0.000,0.000,0.000}10}%
}}}}
\put(1180,-1808){\makebox(0,0)[b]{\smash{{\SetFigFont{9}{10.8}{\familydefault}{\mddefault}{\updefault}{\color[rgb]{0.000,0.000,0.000}9}%
}}}}
\put(2722,-1293){\makebox(0,0)[b]{\smash{{\SetFigFont{9}{10.8}{\familydefault}{\mddefault}{\updefault}{\color[rgb]{0.000,0.000,0.000}8}%
}}}}
\put(2208,-1293){\makebox(0,0)[b]{\smash{{\SetFigFont{9}{10.8}{\familydefault}{\mddefault}{\updefault}{\color[rgb]{0.000,0.000,0.000}7}%
}}}}
\put(1694,-1293){\makebox(0,0)[b]{\smash{{\SetFigFont{9}{10.8}{\familydefault}{\mddefault}{\updefault}{\color[rgb]{0.000,0.000,0.000}6}%
}}}}
\put(1180,-1293){\makebox(0,0)[b]{\smash{{\SetFigFont{9}{10.8}{\familydefault}{\mddefault}{\updefault}{\color[rgb]{0.000,0.000,0.000}5}%
}}}}
\put(2722,-779){\makebox(0,0)[b]{\smash{{\SetFigFont{9}{10.8}{\familydefault}{\mddefault}{\updefault}{\color[rgb]{0.000,0.000,0.000}4}%
}}}}
\put(2208,-779){\makebox(0,0)[b]{\smash{{\SetFigFont{9}{10.8}{\familydefault}{\mddefault}{\updefault}{\color[rgb]{0.000,0.000,0.000}3}%
}}}}
\put(1694,-779){\makebox(0,0)[b]{\smash{{\SetFigFont{9}{10.8}{\familydefault}{\mddefault}{\updefault}{\color[rgb]{0.000,0.000,0.000}2}%
}}}}
\put(1180,-779){\makebox(0,0)[b]{\smash{{\SetFigFont{9}{10.8}{\familydefault}{\mddefault}{\updefault}{\color[rgb]{0.000,0.000,0.000}1}%
}}}}
\put(1180,-2321){\makebox(0,0)[b]{\smash{{\SetFigFont{9}{10.8}{\familydefault}{\mddefault}{\updefault}{\color[rgb]{0.000,0.000,0.000}13}%
}}}}
\put(4951,-4486){\makebox(0,0)[b]{\smash{{\SetFigFont{10}{12.0}{\familydefault}{\mddefault}{\updefault}{\color[rgb]{0.000,0.000,0.000}12}%
}}}}
\put(4951,-3886){\makebox(0,0)[b]{\smash{{\SetFigFont{10}{12.0}{\familydefault}{\mddefault}{\updefault}{\color[rgb]{0.000,0.000,0.000}8}%
}}}}
\put(4351,-3886){\makebox(0,0)[b]{\smash{{\SetFigFont{10}{12.0}{\familydefault}{\mddefault}{\updefault}{\color[rgb]{0.000,0.000,0.000}7}%
}}}}
\put(751,-3886){\makebox(0,0)[b]{\smash{{\SetFigFont{10}{12.0}{\familydefault}{\mddefault}{\updefault}{\color[rgb]{0.000,0.000,0.000}8}%
}}}}
\put(2851,-3886){\makebox(0,0)[b]{\smash{{\SetFigFont{10}{12.0}{\familydefault}{\mddefault}{\updefault}{\color[rgb]{0.000,0.000,0.000}8}%
}}}}
\put(4951,-5686){\makebox(0,0)[b]{\smash{{\SetFigFont{10}{12.0}{\familydefault}{\mddefault}{\updefault}{\color[rgb]{0.000,0.000,0.000}4}%
}}}}
\put(4599,-6417){\makebox(0,0)[b]{\smash{{\SetFigFont{10}{12.0}{\familydefault}{\mddefault}{\updefault}{\color[rgb]{0.000,0.000,0.000}7}%
}}}}
\put(4596,-5913){\makebox(0,0)[b]{\smash{{\SetFigFont{10}{12.0}{\familydefault}{\mddefault}{\updefault}{\color[rgb]{0.000,0.000,0.000}3}%
}}}}
\put(2851,-5686){\makebox(0,0)[b]{\smash{{\SetFigFont{10}{12.0}{\familydefault}{\mddefault}{\updefault}{\color[rgb]{0.000,0.000,0.000}4}%
}}}}
\end{picture}%